\documentclass[11pt]{article}

\usepackage{amsthm}
\usepackage{amsmath}
\usepackage{amsfonts}
\usepackage{mathrsfs}
\usepackage{dsfont}

\usepackage{amssymb}
\usepackage{enumerate}
\usepackage{graphicx}
\usepackage{mathrsfs}

\usepackage{a4}
\usepackage{url}
\usepackage{algorithm}
\usepackage{algorithmicx}
\usepackage{algcompatible}
\usepackage{algpseudocode}
\usepackage{color}
\usepackage{booktabs}

\usepackage[authoryear]{natbib}

\newcommand{\bea}{\begin{eqnarray*}}
\newcommand{\eea}{\end{eqnarray*}}
\newcommand{\be}{\begin{eqnarray}}
\newcommand{\ee}{\end{eqnarray}}

\def\ra{\rightarrow}
\def\raas{\stackrel{\rm a.s.}{\ra}}
\def\rad{\stackrel{\rm d}{\ra}}

\def\dd{\mathrm{d}}
\def\arg{\mathrm{arg}}
\def\Arg{\mathrm{Arg}}

\def\erf{\mathrm{erf}}

\def\supp{\mathrm{supp}}

\def\mt{\theta}

\DeclareMathOperator{\CR}{\mathsf{CR}}

\def\MMD{\mathsf{MMD}}

\def\KH{\mathsf{KH}}
\def\GM{\mathsf{GM}}

\def\Ex{\mathsf{E}}

\def\var{\mathsf{var}}

\def\hg{\widehat{g}}
\def\tg{\widetilde{g}}

\def\ma{\alpha}
\def\mg{\gamma}

\def\ml{\lambda}
\def\mt{\theta}

\def\ms{\sigma}

\def\Ex{\mathsf{E}}

\def\ra{\rightarrow}
\def\TT{^\top}
\def\e1{\mathsf{e}}

\def\ab{\mathbf{a}}

\def\Ib{\mathbf{I}}

\def\Kb{\mathbf{K}}

\def\mob{\mathbf{\omega}}

\def\Xb{\mathbf{X}}

\def\1b{\mathbf{1}}
\def\0b{\mathbf{0}}

\def\eb{\mathbf{e}}
\def\gb{\mathbf{g}}

\def\kb{\mathbf{k}}
\def\pb{\mathbf{p}}

\def\ub{\mathbf{u}}

\def\wb{\mathbf{w}}
\def\xb{\mathbf{x}}

\def\SE{{\mathscr E}}

\def\SH{\mathcal{H}}
\def\SI{\mathds{I}}

\def\SM{{\mathscr M}}
\def\SN{{\mathscr N}}
\def\SO{\mathcal{O}}
\def\SP{{\mathscr P}}
\def\SS{{\mathscr S}}

\def\SX{{\mathscr X}}

\newcommand{\fin}{\mbox{}~\hfill\mbox{$\lhd$}}
\newcommand{\carre}{\mbox{}~\hfill\rule{2mm}{2mm}}

\newtheorem{rmk}{Remark}

\newtheorem{thm}{Theorem}

\newtheorem{lemma}{Lemma}

\newcommand{\vsp}{\vspace{0.3cm}}

\begin{document}

\title{Performance analysis of greedy algorithms for minimising a Maximum Mean Discrepancy}

\author{Luc Pronzato\\
\mbox{}\\
Universit\'e C\^ote d'Azur, CNRS, Laboratoire I3S\\
B\^at.\ Euclide, Les Algorithmes, 2000 route des lucioles,\\
06900 Sophia Antipolis, France \\
{\tt Luc.Pronzato@cnrs.fr}}

\maketitle

\begin{abstract} We analyse the performance of several iterative algorithms for the quantisation of a probability measure $\mu$, based on the minimisation of a Maximum Mean Discrepancy (MMD). Our analysis includes kernel herding, greedy MMD minimisation and Sequential Bayesian Quadrature (SBQ). We show that the finite-sample-size approximation error, measured by the MMD, decreases as $1/n$ for SBQ and also for kernel herding and greedy MMD minimisation when using a suitable step-size sequence. The upper bound on the approximation error is slightly better for SBQ, but the other methods are significantly faster, with a computational cost that increases only linearly with the number of points selected. This is illustrated by two numerical examples, with the target measure $\mu$ being uniform (a space-filling design application) and with $\mu$ a Gaussian mixture. They suggest that the bounds derived in the paper are overly pessimistic, in particular for SBQ. The sources of this pessimism are identified but seem difficult to counter.
\end{abstract}

{\small {\bf keywords} Maximum Mean Discrepancy; quantisation; greedy algorithm; sequential Bayesian quadrature; kernel herding; space-filling design; computer experiments
}


\section{Introduction and motivation}

\paragraph{Background.} Quantisation of a probability measure $\mu$ is a basic task in many fields, such as probabilistic integration \citep{BriolOGOS2019}, MCMC computation \citep{JosephDRW2015, JosephWGLT2019} or space-filling design in computer experiments \citep{JosephGB2015, MakJ2017b, MakJ2018, PZ2020-SIAM}, and minimisation of the Maximum Mean Discrepancy (MMD) defined by a kernel $K$ is a powerful tool for this task\footnote{We do not consider quantisation methods based on Voronoi partitions, for which one can refer in particular to \citet{GrafL2000}.}. In particular, it easily allows iterative constructions that can be stopped when the discrete approximation obtained is deemed sufficient, a situation where the number of support points is not fixed in advance.

\paragraph{Claims and hint of the contents.} We derive finite-sample-size errors bounds for iterative methods to quantise a probability measure by minimising the MMD for a given kernel. The methods considered include gradient-type algorithms (kernel herding), greedy one-step-ahead minimisation, and Sequential Bayesian Quadrature (SBQ) that sets optimal weights on the current support at each iteration. Two variants of SBQ are considered, with and without the constraint that the weights sum to one (the bound for the unconstrained version is markedly pessimistic but our analysis reveals a connection with kernel herding and gives some insight for the reason of this pessimism). We consider the practical situation where the candidate set is finite; it may correspond in particular to points independently sampled with $\mu$, with the possibility to use a different set at every iteration (see Section~\ref{S:Random-candidates} and Appendix~C). The context of a finite candidate set is the most widely used in practical situations. It allows us to derive simple proofs that only use (finite-dimensional) linear algebra, but our results can be extended to the infinite-dimensional (Hilbert space) situation, where the new support point selected at each iteration is searched within a continuous set; see, e.g., \citet{ChenMGBO2018, TeymurGRO2021}. We show that the error is $\SO(n^{-1})$ for SBQ and for algorithms that use a suitable step-size sequence and construct nonuniform discrete measures (with a slightly better constant for SBQ). We show that it is also $\SO(n^{-1})$ for the construction of uniform (empirical) measures provided that the measure with total mass one minimising the MMD over the candidate set is a probability measure. We show that the complexity of gradient and greedy one-step-ahead methods grows linearly with $n$, whereas it grows quadratically for SBQ. Two variants of kernel herding are considered, with similar performance to SBQ but slightly lighter calculations. 

\paragraph{Paper organisation.} Section~\ref{S:basic+notation} recalls the background on MMD and Bayesian quadrature. It defines the notation and introduces the methods that are considered in the rest of the paper. The performance of kernel herding is analysed in Section~\ref{S:performanceKH}. The results presented in Sections~\ref{S:performance-KH-empirical} and \ref{S:KH-nonuniform} are not new, but the analysis of this basic gradient-type algorithm is central to the investigation of the convergence rate for the other methods, more sophisticated, that we consider in Sections~\ref{S:KH-OLWO-IWO} (variants of kernel herding), \ref{S:performanceMMD} (greedy MMD minimisation) and \ref{S:performanceSBQ} (SBQ). Section~\ref{S:Random-candidates} extends the results of previous sections to the case where the candidate set corresponds to points independently sampled with $\mu$. Two illustrative examples are presented in Section~\ref{S:examples}, one with $\mu$ uniform (space-filling design), the other with $\mu$ a Gaussian mixture. Section~\ref{S:conclusions} concludes briefly.

\section{Maximum Mean Discrepancy and Bayesian quadrature}\label{S:basic+notation}

\subsection{Maximum Mean Discrepancy (MMD)}

Let $\SX$ be a measurable set, equipped with a probability measure $\mu$. For instance, for application to space-filling design for computer experiments, $\SX$ is typically a compact subset of $\mathds{R}^d$ and $\mu$ is proportional to the Lebesgue measure on $\SX$.
Let $K$ be a symmetric strictly positive definite (s.p.d.) kernel defined on $\SX\times\SX$, uniformly bounded on $\SX$; that is,
\be
K(\xb,\xb)\leq \overline{K} < +\infty\,, \ \mbox{ for all } \xb\in\SX\,, \label{Kbar}
\ee
and for any $n\in\mathds{N}$ and any $\xb_1,\ldots,\xb_n$ in $\SX$, the $n\times n$ matrix $\Kb_n$ with element $i,j$ equal to $K(\xb_i,\xb_j)$ is p.d.\ and s.p.d.\ when the $\xb_i$ are all pairwise different.
Note that $K$ being s.p.d.\ implies that $K^2(\xb,\xb')\leq K(\xb,\xb)K(\xb',\xb')$ for all $\xb$ and $\xb'$ in $\SX$ with the inequality being strict when $\xb\ne\xb'$. Moreover, \eqref{Kbar} implies
$\tau_\mg(\mu) = \int_\SX K^\mg(\xb,\xb)\, \dd\mu(\xb) \leq \overline{K}^\mg < +\infty \ \mbox{ for any } \mg\geq 0$.
$K$ defines a Reproducing Kernel Hilbert Space (RKHS) $\SH_K$, and we respectively denote by $\langle\cdot,\cdot\rangle_K$ and $\|\cdot\|_{\SH_K}$ the scalar product and norm in $\SH_K$. We do not assume that $\SH_K$ is finite-dimensional. We say that $K$ is positive ($K\geq 0$) when $K(\xb,\xb')\geq 0$ for all $\xb$ and $\xb'$ in~$\SX$.

We denote by $\SM(\SX)$ the set of finite signed measures on $\SX$, by $\SM_{[1]}(\SX)$ the set of signed measures with total mass $1$, and by $\SM^+_{[1]}(\SX)$ the set of probability measures on $\SX$ (with thus $\mu\in\SM^+_{[1]}(\SX)$).
The reproducing property implies that, for any $\nu\in\SM(\SX)$, the \emph{energy} of $\nu$, defined by $\SE_K(\nu) = \int_{\SX^2} K(\xb,\xb')\,\dd\nu(\xb)\,\dd\nu(\xb')$, satisfies
\bea
\SE_K(\nu) = \int_{\SX^2} \langle K(\xb,\cdot),K(\xb',\cdot)\rangle_K \,\dd\nu(\xb)\,\dd\nu(\xb') &\leq&
\int_{\SX^2} \|K(\xb,\cdot)\|_{\SH_K}\,\|K(\xb',\cdot)\|_{\SH_K} \,\dd\nu(\xb)\,\dd\nu(\xb') \\
&=& \left[ \int_\SX K^{1/2}(\xb,\xb)\, \dd\nu(\xb)\right]^2 = \tau_{1/2}^2(\nu) < +\infty \,.
\eea
For any $\nu\in\SM(\SX)$ and any $\xb\in\SX$, we denote by
\bea
P_{K,\nu}(\xb) = \int_\SX K(\xb,\xb')\,\dd\nu(\xb')
\eea
the \emph{potential} of $\nu$ at $\xb$; $P_{K,\nu}(\cdot)$ is also called the \emph{kernel imbedding} of $\nu$ into $\SH_K$.

For $\mu$ and $\nu$ in $\SM^+_{[1]}(\SX)$, any $f\in\SH_K$ satisfies the following (Koksma-Hlawka type) inequality:
$\left| \int_\SX f(\xb)\,\dd\nu(\xb) - \int_\SX f(\xb)\,\dd\mu(\xb)\right| = \left|\langle f,P_{K,\mu}-P_{K,\nu}\rangle_K\right| \leq \|f\|_{\SH_K}\, \MMD_K(\mu,\nu)$, where
\be
\MMD_K(\mu,\nu) &=& \sup_{\|f\|_{\SH_K}=1}  \left| \int_\SX f(\xb)\,\dd\nu(\xb) - \int_\SX f(\xb)\,\dd\mu(\xb)\right| = \|P_{K,\nu}-P_{K,\mu}\|_{\SH_K} \nonumber \\
&=& \SE_K^{1/2} (\nu-\mu) = \left[\int_{\SX^2} K(\xb,\xb')\,\dd(\nu-\mu)(\xb)\,\dd(\nu-\mu)(\xb')\right]^{1/2} \nonumber \\
&=& \left[ \SE_K(\mu)+\SE_K(\nu) - 2\,\int_\SX P_{K,\mu}(\xb)\,\dd\nu(\xb) \right]^{1/2} \label{MMD4}
\ee
is called the \emph{Maximum-Mean-Discrepancy} (MMD) between $\nu$ and $\mu$; see \citet[Def.~10]{SejdinovicSGF2013}. $\MMD_K(\mu,\nu)$ defines an integral pseudometric between probability distributions and a pseudometric between kernel imbeddings. It defines a metric on $\SM^+_{[1]}(\SX)$ when $K$ is characteristic\footnote{Since $K$ is uniformly bounded, this is equivalent to the condition that $K$ be Conditionally Integrally Strictly Positive Definite (CISPD), that is, $\SE_K(\nu)>0$ for all nonzero signed measure $\nu\in\SM_{[0]}(\SX)$; see \citet[Def.~6 and Lemma~8]{SriperumbudurGFSL2010}; see also \citet{PZ2020-SIAM} for a comprehensive survey including the case of singular kernels.}, which we assume in the following. This implies in particular that $\MMD_K(\mu,\nu)>0$ for any $\nu\in\SM_{[1]}(\SX)$, $\nu\neq\mu$.

For a collection $\Xb_n=\{\xb_1,\ldots,\xb_n\}$ of $n$ points in $\SX$, called $n$-point design, we denote by $\xi_{n,e}=(1/n)\sum_{i=1}^n \delta_{\xb_i}$ the associated empirical measure, with $\delta_\xb$ the Dirac delta measure at $\xb$. For $\wb_n=(\{\wb_n\}_1,\ldots,\{\wb_n\}_n)\TT\in\mathds{R}^n$ a vector of $n$ weights, we denote by $\xi_n=\xi(\wb_n)$ the signed measure
\be\label{xin}
\xi(\wb_n) = \sum_{i=1}^n \{\wb_n\}_i\, \delta_{\xb_i}
\ee
(so that $\xi_{n,e}=\xi(\1b_n/n)$, with $\1b_n$ the $n$-dimensional vector with all components equal to 1).
An important area of application for MMD minimisation is space-filling design, where the objective is to build evenly distributed designs on a compact $\SX$; see, for example, \citet{PM_SC_2012, P-JSFdS2017}. Minimising $\MMD_K(\mu,\xi_{n,e})$ with $\mu$ uniform over $\SX$ is then an effective approach to achieve this goal. One may also minimise $\MMD_K(\mu,\xi(\wb_n))$ with respect to $\Xb_n$ and $\wb_n$, and the designs obtained differ depending on the chosen kernel $K$, the constraints set on $\wb_n$ and on the optimisation method that is used.
In this paper, we focuss our attention on the construction of extensive point sequences $\Xb_n=[\xb_1,\xb_2,\ldots,\xb_n]$, such that $\Xb_{k+1}=[\Xb_k,\xb_{k+1}]$ for all $k$, with the property that $\Xb_n$ is the support of a measure $\xi_n$ which approximates $\mu$ well in the sense of $\MMD_K(\mu,\xi_n)$.

Analytic expressions for the quantities $\SE_K(\mu)$ and $P_{K,\mu}(\cdot)$ that appear in \eqref{MMD4} are available for particular measures and particular kernels, see Table~1 of \cite{BriolOGOS2019}. This includes the case when $\mu$ is uniform on $\SX=[0,1]^d$ and $K$ is separable, see for example Table~3.1 of \citep{PZ2020-SIAM}, and separable kernels $K$ based on variants of Brownian motion covariance yield $L_2$ discrepancies (symmetric, centred, wrap-around and so on); see \cite{Hickernell1998}, \citet[Chap.~3]{FangLS2006}.
$\SE_K(\mu)$ and $P_{K,\mu}(\cdot)$ are not available when $\mu$ is a posterior distribution with unknown normalising constant; in that case, \citet{JosephDRW2015, JosephWGLT2019} suggest to construct minimum-energy designs that minimise $\SE_K(\xi_{n,e})$ for a particular kernel $K$. Another way is to minimise a kernel Stein discrepancy, that is, to minimise MMD for the image $K'$ of a kernel $K$ under a Stein operator, so that $\SE_{K'}(\mu)=0$ and $P_{K',\mu}(\xb)=0$ for any $\xb$; see \cite{ChenMGBO2018, DetommasoCSMS2018, GorhamM2017, LiuW2016, OatesGC2017}.
Throughout the paper we consider the general framework where $\SH_K$ is an infinite-dimensional RKHS and assume that $\SE_K(\mu)$ and $P_{K,\mu}(\xb)$ can be easily computed for any $\xb\in\SX$ (Monte-Carlo methods can always be used as a last resort).

\subsection{MMD and optimal weights for discrete measures}

For a given design $\Xb_n$, $\MMD_K(\mu,\xi_n)$ is quadratic in $\wb_n$, and the optimal weights are easily obtained. Indeed, \eqref{MMD4} gives
\be
\MMD_K^2(\mu,\xi_n) = \SE_K(\xi_n-\mu) &=& \sum_{i,j=1}^n \{\wb_n\}_i \{\wb_n\}_j K(\xb_i,\xb_j) - 2\, \sum_{i=1}^n \{\wb_n\}_i P_{K,\mu}(\xb_i) + \SE_K(\mu) \,, \nonumber\\
&=& \wb_n\TT \Kb_n \wb_n - 2\, \wb_n\TT \pb_n(\mu) + \SE_K(\mu) \,, \label{MMD0}
\ee
where $\pb_n(\mu)=[P_{K,\mu}(\xb_1),\ldots,P_{K,\mu}(\xb_n)]\TT$ (alternative expressions for $\MMD_K^2(\mu,\xi_n)$ are given in Appendix~A). Therefore, $\wb_n^*$ that minimises $\MMD_K^2(\mu,\xi_n)$ under the constraints $\{\wb_n\}_i\geq 0$ and $\1b_n\TT\wb_n=1$ is solution of a Quadratic Programming (QP) problem.
We assume that the $\xb_i$ in $\Xb_n$ are pairwise different, so that $\Kb_n$ has full rank. Releasing the positivity constraints, $\widehat\wb_n$ that minimises $\MMD_K^2(\mu,\xi_n)$ with $\1b_n\TT\wb_n=1$ is obtained explicitly as
\be\label{hatwn}
\widehat\wb_n = \left(\Kb_n^{-1} - \frac{\Kb_n^{-1} \1b_n\1b_n\TT \Kb_n^{-1}}{\1b_n\TT\Kb_n^{-1}\1b_n} \right) \pb_n(\mu) + \frac{\Kb_n^{-1} \1b_n}{\1b_n\TT\Kb_n^{-1}\1b_n}
\ee
(with $\wb_n^*=\widehat\wb_n$ when all components of $\widehat\wb_n$ are nonnegative). Also, the unconstrained weights that minimise $\MMD_K^2(\mu,\xi_n)$ are given by
\be\label{tildewn}
\widetilde\wb_n=\Kb_n^{-1}\pb_n(\mu)\,.
\ee
Throughout the paper, for any measure $\xi_n$ supported on $\Xb_n$, we denote by $\xi_n^*$, $\widehat\xi_n$ and $\widetilde\xi_n$ the measures with the same support and respective weights $\wb_n^*$, $\widehat\wb_n$ and $\widetilde\wb_n$, so that $\MMD_K^2(\mu,\widetilde\xi_n)\leq \MMD_K^2(\mu,\widehat\xi_n)\leq \MMD_K^2(\mu,\xi_n^*)$ (and  $\MMD_K^2(\mu,\xi_n^*)\leq \MMD_K^2(\mu,\xi_n)$ if $\xi_n\in\SM_{[1]}^+(\Xb_n)$).

\subsection{Incremental MMD minimisation}\label{S:incremental-constructions}

We consider three families of incremental constructions.

\subsubsection{Sequential Bayesian Quadrature (SBQ)}\label{S:SBQ}
The construction of a design $\Xb_n$ that minimises $\MMD_K^2(\mu,\widetilde\xi_n)$ or $\MMD_K^2(\mu,\widehat\xi_n)$ is called Bayesian quadrature (BQ); it can be Sequential (SBQ), see \citet{BriolOGO2015}, and we consider two versions of SBQ. Bounds on their finite-sample-size error are given in Section~\ref{S:performanceSBQ}. Note that $\MMD_K^2(\mu,\widetilde\xi_k)=\MMD_K^2(\mu,\widetilde\xi_{k+1})$ (respectively, $\MMD_K^2(\mu,\widehat\xi_k)=\MMD_K^2(\mu,\widehat\xi_{k+1})$) when $\widetilde\xi_k$ and $\widetilde\xi_{k+1}$ (respectively, $\widehat\xi_k$ and $\widehat\xi_{k+1}$) have the same support, so that SBQ always selects new points whenever possible (i.e., until all eligible points are exhausted). We may thus assume that the $\xb_i$ are all pairwise different, $i=1,\ldots,k$, and that $\Kb_k$ has full rank for all $k$.

\paragraph{(\textit{i}) Greedy minimisation of $\MMD_K^2(\mu,\widetilde\xi_k)$.}

The equations \eqref{MMD0} and \eqref{tildewn} give
$\MMD_K^2(\mu,\widetilde\xi_k) = \SE_K(\mu) - \pb_k\TT(\mu) \Kb_k^{-1} \pb_k(\mu)$.
We have
\bea
\Kb_{n+1}= \left[
             \begin{array}{cc}
               \Kb_n & \kb_n(\xb_{n+1}) \\
               \kb_n\TT(\xb_{n+1}) & K(\xb_{n+1},\xb_{n+1}) \\
             \end{array}
           \right]\,,
\eea
where $\kb_n(\xb)=[K(\xb_1,\xb),\ldots,K(\xb_n,\xb)]\TT$, $\xb\in\SX$. The calculation of its inverse by
\be\label{Kbnm1}
\Kb_{n+1}^{-1} = \left(
                   \begin{array}{cc}
                     \Kb_n^{-1} + \beta_{n+1}\,\ub_{n+1}\ub_{n+1}\TT & -\beta_{n+1}\,\ub_{n+1} \\
                     -\beta_{n+1}\,\ub_{n+1}\TT & \beta_{n+1} \\
                   \end{array}
                 \right)\,,
\ee
with $\ub_{n+1}=\Kb_n^{-1}\kb_n(\xb_{n+1})$ and $\beta_{n+1}=[K(\xb_{n+1},\xb_{n+1})-\kb_n\TT(\xb_{n+1})\ub_{n+1}]^{-1}$, will be used several times for incremental constructions and gives here
\bea
\MMD_K^2(\mu,\widetilde\xi_{k+1})= \MMD_K^2(\mu,\widetilde\xi_k)  - \frac{\left[\pb_k\TT(\mu)\Kb_k^{-1}\kb_k(\xb_{k+1})-P_{K,\mu}(\xb_{k+1})\right]^2} {K(\xb_{k+1},\xb_{k+1})-\kb_k\TT(\xb_{k+1})\Kb_k^{-1}\kb_k(\xb_{k+1})}  \,.
\eea
Since $\widetilde\wb_k$ satisfies \eqref{tildewn}, we get
\be\label{MMD-SBQ1}
\MMD_K^2(\mu,\widetilde\xi_{k+1})= \MMD_K^2(\mu,\widetilde\xi_k)  - \frac{\left[P_{K,\widetilde\xi_k}(\xb_{k+1})-P_{K,\mu}(\xb_{k+1})\right]^2}{\min_{\wb\in\mathds{R}^k} \|K(\xb_{k+1},\cdot)-\wb\TT\kb_k(\cdot)\|_{\SH_K}^2}  \,.
\ee
This corresponds to the ``standard'' version of SBQ, which uses general signed measures $\widetilde\xi_k$ in $\SM(\SX)$: it selects $\xb_1\in\Arg\max_{\xb\in\SX} P_{K,\mu}^2(\xb)/K(\xb,\xb)$ and then
\be\label{SBQ1}
\xb_{k+1} \in\Arg\max_{\xb\in\SX} \frac{\left[P_{K,\widetilde\xi_k}(\xb)-P_{K,\mu}(\xb)\right]^2} {K(\xb,\xb)-\kb_k\TT(\xb)\Kb_k^{-1}\kb_k(\xb)} \,, \ k\geq 1\,.
\ee

\paragraph{(\textit{ii}) Greedy minimisation of $\MMD_K^2(\mu,\widehat\xi_k)$.}

The equations \eqref{MMD0} and \eqref{hatwn} give
\bea
\MMD_K^2(\mu,\widehat\xi_k) &=& \SE_K(\mu) - \pb_k\TT(\mu) \Kb_k^{-1} \pb_k(\mu) + \frac{(1-\pb_k\TT(\mu)\Kb_k^{-1}\1b_k)^2}{\1b_k\TT\Kb_k^{-1}\1b_k} \,,
\eea
but a simpler expression can be obtained through the introduction of the reduced kernel $K_\mu$ defined by
\be\label{Kmu}
K_\mu(\xb,\xb')= K(\xb,\xb')- P_{K,\mu}(\xb)-P_{K,\mu}(\xb')+\SE_K(\mu)\,, \ \xb,\xb'\in\SX \,.
\ee
Let $\{{\Kb_\mu}_k\}_{i,j}=K_\mu(\xb_i,\xb_j)$, $i,j=1,\ldots,k$, so that ${\Kb_\mu}_k=\Kb_k-\pb_k(\mu)\1b_k\TT-\1b_k \pb_k\TT(\mu)+\SE_K(\mu)\,\1b_k\1b_k\TT$ and, for any measure $\xi_k$ with total mass one,
\be\label{MMD-with-Ktilde}
\MMD_K^2(\mu,\xi_k)=\wb_k\TT {\Kb_\mu}_k \wb_k \,.
\ee
As $K$ is characteristic and $\Kb_k$ has full rank, ${\Kb_\mu}_k$ is invertible when $\mu$ is not fully supported on $\Xb_k$. Indeed, let $\ub_k$ be an eigenvector of ${\Kb_\mu}_k$. If $a=\ub_k\TT\1b_k\neq 0$, then the measure $\xi_k$ with weights $\wb_k=\ub_k/a$ has total mass one and satisfies $\MMD_K^2(\mu,\xi_k)=\mg_k>0$ since $\xi_k\neq\mu$, so that $\ub_k\TT{\Kb_\mu}_k\ub_k=a^2 \mg_k>0$. If $a=\ub_k\TT\1b_k= 0$, then $\ub_k\TT{\Kb_\mu}_k\ub_k=\ub_k\TT\Kb_k\ub_k$, which is strictly positive since $\Kb_k$ has full rank. Direct calculation then gives
\bea
\widehat\wb_k = {\Kb_\mu}_k^{-1}\1b_k/(\1b_k\TT{\Kb_\mu}_k^{-1}\1b_k) \ \mbox{ and }  \ \MMD_K^2(\mu,\widehat\xi_k)=1/(\1b_k\TT{\Kb_\mu}_k^{-1}\1b_k) \,, 
\eea
and, using block matrix inversion for ${\Kb_\mu}_{k+1}$,
\bea
\MMD_K^2(\mu,\widehat\xi_{k+1})=\left\{ \1b_k\TT{\Kb_\mu}_k^{-1}\1b_k+\frac{\left[\1b_k\TT{\Kb_\mu}_k^{-1}{\kb_\mu}_k(\xb_{k+1})-1\right]^2} {K_\mu(\xb_{k+1},\xb_{k+1})-{\kb_\mu}_k\TT(\xb_{k+1}){\Kb_\mu}_k^{-1}{\kb_\mu}_k(\xb_{k+1})} \right\}^{-1} \,,
\eea
where ${\kb_\mu}_k(\xb)=[K_\mu(\xb_1,\xb),\ldots,K_\mu(\xb_k,\xb)]\TT$, $\xb\in\SX$.
Straightforward manipulations using \eqref{hatwn} give
\be\label{MMD-SBQ2}
\MMD_K^2(\mu,\widehat\xi_{k+1})= \MMD_K^2(\mu,\widehat\xi_k)  - \frac{\left[P_{K,\widehat\xi_k}(\xb_{k+1})-P_{K,\mu}(\xb_{k+1})+ \widehat\wb_k\TT\pb_k(\mu)-\SE_K(\widehat\xi_k)\right]^2}{\min_{\scriptsize
\begin{array}{l}
\wb\in\mathds{R}^k \\
\1b_k\TT\wb=1
\end{array}
} \|K(\xb_{k+1},\cdot)-\wb\TT\kb_k(\cdot)\|_{\SH_K}^2} \,.
\ee
This version of SBQ selects $\xb_1\in\Arg\min_{\xb\in\SX} K_\mu(\xb,\xb)$ and then
\be\label{SBQ2}
\xb_{k+1} \in\Arg\max_{\xb\in\SX} \frac{\left[P_{K,\widehat\xi_k}(\xb_)-P_{K,\mu}(\xb)+ \widehat\wb_k\TT\pb_k(\mu)-\widehat\wb_k\TT\Kb_k\widehat\wb_k)\right]^2} {K(\xb,\xb)-\kb_k\TT(\xb)\Kb_k^{-1}\kb_k(\xb)+
\frac{[1-\1b_k\TT\Kb_k^{-1}\kb_k(\xb)]^2}{\1b_k\TT\Kb_k^{-1}\1b_k}} \,, \ k\geq 1\,.
\ee

The expressions \eqref{MMD-SBQ1} and \eqref{MMD-SBQ2} are pivotal to the derivation of finite-sample-size error bounds for SBQ through the consideration of simplified versions where $\xb_{k+1}$ is chosen by kernel herding, see Section~\ref{S:off-line-KH}.
When $\xb$ is selected at each iteration within a candidate set of size $C$, see Section~\ref{S:notation}, the complexity of SBQ grows like $\SO(n^2\,C)$ for $n$ iterations (the main contribution comes from the denominators in \eqref{SBQ1} and \eqref{SBQ2} which must be calculated for the $C$ candidates, but matrix-vector multiplications $\Kb_k^{-1}\kb_k(\xb)$ can be avoided by using recursive calculation; see Remark~\ref{R:invKn}).

\subsubsection{Greedy MMD Minimisation (GM)}\label{S:greedyMMD}

To lighten the computations required by SBQ, we can consider the optimal choice of successive $\xb_k$ for a predefined sequence of weights $\wb_k$. The standard version of Greedy MMD Minimisation (GM) uses $\wb_k=\1b_k/k$ for all $k$, so that $\xi_k$ is the empirical measure $\xi_{k,e}$ supported on $\Xb_k$. It selects $\xb_1\in\Arg\min_{\xb\in\SX} K(\xb,\xb)-2\,P_{K,\mu}(\xb)=\Arg\min_{\xb\in\SX} K_\mu(\xb,\xb)$ and then minimises $\MMD_K(\mu,\xi_{k+1,e})$ incrementally: \eqref{MMD0} gives
\be\label{empirical-greedy}
\xb_{k+1} \in \Arg\min_{\xb\in\SX} \sum_{i=1}^k K(\xb_i,\xb) + \frac12\, K(\xb,\xb) - (k+1)\, P_{K,\mu}(\xb) \,, \ k\geq 1\,.
\ee
GM will be considered in Section~\ref{S:performanceMMD}. The complexity of MMD grows linearly and is $\SO(n\,C)$ for $n$ iterations when the selection is among $C$ possible candidates. In Section~\ref{S:MMD-nonuniform} we also consider versions with nonuniform weights: one must then define the weight $w_{k+1}$ to be allocated to the next point $\xb_{k+1}$, not selected yet. A convenient way to proceed, usual in the area of optimal design of experiments, is to take $\xi_{k+1}=(1-\ma_{k+1})\,\xi_k + \ma_{k+1} \delta_{\xb_{k+1}}$, for some step size $\ma_{k+1}\in[0,1]$. This construction guarantees that $\xi_{k+1}\in\SM_{[1]}^+(\SX)$ when $\xi_k\in\SM_{[1]}^+(\SX)$;
the choice of $\ma_{k+1}$ defines the sequence of weights $\wb_k$ and the point $\xb_{k+1}$ is chosen to minimise $\MMD_K(\mu,\xi_{k+1})$.

\subsubsection{The Frank-Wolfe algorithm and Kernel Herding (KH)}\label{S:FWandKH}

Another way of proceeding consists in exploiting the convexity of the functional $\phi_{K,\mu}(\cdot):\xi\to\phi_{K,\mu}(\xi)=\MMD_K^2(\mu,\xi)$ using a gradient descent algorithm. This gives a family of methods for which performance bounds can be easily established by convexity arguments, arguments that are also applicable to the derivation of performance bounds for the GM and SBQ algorithms.

For any $\xi,\nu\in\SM(\SX)$, the directional derivative of
$\phi_{K,\mu}(\cdot)$
at $\xi$ in the direction $\nu$ equals
\bea
F_{\MMD_K^2}(\xi,\nu)=2\,\left[ \int_{\SX^2} K_\mu(\xb,\xb')\,\dd\nu(\xb)\,\dd\xi(\xb') - \SE_{K_\mu}(\xi) + \int_\SX P_{K,\mu}(\xb)\, \dd(\xi-\nu)(\xb)\right] \,,
\eea
so that $F_{\MMD_K^2}(\xi,\delta_{\xb}) = 2\,[P_{K,\xi}(\xb)-P_{K,\mu}(\xb)+ \int_\SX P_{K,\mu}(\xb')\,\dd\xi(\xb')-\SE_K(\xi) ]$. Iterations of the Frank-Wolfe algorithm \citep{FrankW56} correspond to $\xi_{k+1}=(1-\ma_{k+1})\,\xi_k + \ma_{k+1} \nu_{k+1}$, where  $\nu_{k+1}\in\Arg\min_{\nu\in\SM_{[1]}^+(\SX)} F_{\MMD_K^2}(\xi_k,\nu)$ and $\ma_{k+1}\in[0,1]$. This gives $\nu_{k+1}=\delta_{\xb_{k+1}}$,
with
\be
\xb_{k+1} &\in& \Arg\min_{\xb\in\SX} F_{\MMD_K^2}(\xi_k,\delta_{\xb}) = \Arg\min_{\xb\in\SX} \left[ P_{K,\xi_k}(\xb)-P_{K,\mu}(\xb)\right] \,. \label{xk+1-KH}
\ee
When $\ma_k=1/k$ for all $k$, $\xi_k$ remains uniform on its support (unless the same $\xb$ is chosen several times); see \citet{Wynn70} for an early contribution in optimal design of experiments. The method is also called conditional-gradient and corresponds to kernel herding (KH) used in machine learning \citep{BachLJO2012}. The algorithm selects $\xb_1\in\Arg\max_{\xb\in\SX} P_{K,\mu}(\xb)$ and then
\be\label{empirical-KH}
\xb_{k+1} \in \Arg\min_{\xb\in\SX} \sum_{i=1}^k K(\xb_i,\xb) - k\,P_{K,\mu}(\xb) \,, \ k\geq 1\,.
\ee
Notice the similarity (but not full agreement) with \eqref{empirical-greedy}. In particular, \eqref{empirical-KH} can be used with singular kernels, which have an intrinsic repelling property \citep{PZ2021-JCAM} but for which $K(\xb,\xb)$ is not defined, whereas \eqref{empirical-greedy} cannot.
The complexity of KH is $\SO(n\,C)$ for $n$ iterations when the selection is among $C$ eligible candidates.
In Section~\ref{S:KH-nonuniform} we consider nonuniform weights, including the case where $\ma_{k+1}$ is chosen optimally in $\xi_{k+1}=(1-\ma_{k+1})\,\xi_k + \ma_{k+1} \delta_{\xb_{k+1}}$. In the area of optimal design of experiments, this corresponds to Fedorov's algorithm (1972)\nocite{Fedorov72}.

We shall also consider two variants of KH (Section~\ref{S:KH-OLWO-IWO}). First, in a Bayesian integration application, at iteration $k$ of the algorithm we can exploit the support of $\xi_k$ only, and use one of the optimal measures $\xi_k^*$, $\widehat\xi_k$, or $\widetilde\xi_k$, with respective weights $\wb_k^*$, $\widehat\wb_k$ and $\widetilde\wb_k$, for integration; we shall call this variant \emph{Off-Line Weight Optimisation} (OLWO)\footnote{It is called Frank-Wolfe Bayesian quadrature in \citep{BriolOGO2015}.}. Second, we can replace $\xi_k$ by $\xi_k^*$, $\widehat\xi_k$, or $\widetilde\xi_k$, \emph{in the algorithm itself} before next iteration; we shall call this variant, closely related to SBQ, \emph{Integrated Weight Optimisation} (IWO)\footnote{Depending how the `optimal' measure is constructed, this includes the minimum-norm point \citep{BachLJO2012} and fully-corrective Frank-Wolfe \citep{LacosteJJ2015} algorithms; see Remark~\ref{R:MNP}.}.

\subsection{Notation}\label{S:notation}

At each iteration, instead of searching $\xb$ in the whole set $\SX$, we shall use a finite subset $\SX_C=\{\xb^{(1)},\ldots,\xb^{(C)}\}$ of candidate points in $\SX$ (typically, $C$ points independently sampled from $\mu$; see Section~\ref{S:Random-candidates}).
We denote $\SI_C=\{1,\ldots,C\}$,
\be
\overline{K}_C &=&\max_{\xb\in\SX_C} K(\xb,\xb) \leq \overline{K} \,, \nonumber \\
\overline{K}_{\mu,C} &=& \max_{\xb\in\SX_C} K_\mu(\xb,\xb) = \max_{\xb\in\SX_C} K(\xb,\xb)-2\,P_{K,\mu}(\xb) + \SE_K(\mu) \nonumber \\
&& \leq \, \overline{K}_C + 2\, \overline{K}_C^{1/2} \tau_{1/2}(\mu) + \tau_{1/2}^2(\mu) = \left[\overline{K}_C^{1/2}+\tau_{1/2}(\mu)\right]^2  \label{KmuC}
\ee
(and $\overline{K}_{\mu,C}\leq \overline{K}_C+\tau_{1/2}^2(\mu)$ when $K\geq 0$). We also denote by $\Kb_C$ and ${\Kb_\mu}_C$ the $C\times C$ matrices with $i,j$ elements respectively equal to $K(\xb^{(i)},\xb^{(j)})$ and $K_\mu(\xb^{(i)},\xb^{(j)})$, for $\xb^{(i)}$, $\xb^{(j)}$ in $\SX_C$; $\eb_j$ is the $j$-th canonical basis vector of $\mathds{R}^C$. We assume that $\mu$ is not fully supported on $\SX_C$, so that ${\Kb_\mu}_C$ has full rank; see Section~\ref{S:SBQ} ($K_\mu$ is thus s.p.d.\ on $\SX_C$).

For any $\xi\in\SM(\SX)$, any $\xb\in\SX$ and any $\ma\in[0,1]$, we define
\be\label{xik+}
\xi^+(\xb,\ma)=(1-\ma)\,\xi+\ma\,\delta_{\xb} \,,
\ee
so that $\xi^+(\xb,\ma)\in\SM_{[1]}^+(\SX_C)$ (respectively, $\SM_{[1]}(\SX_C)$) when $\xb\in\SX_C$ and $\xi\in\SM_{[1]}^+(\SX_C)$ (respectively, $\xi\in\SM_{[1]}(\SX_C)$).

Any probability measure $\xi$ in $\SM^+_{[1]}(\SX_C)$, i.e., supported on $\SX_C$, can be represented as a vector of weights $\mob$ in the probability simplex
\bea
\SP_C=\{\mob\in\mathds{R}^C: \sum_{j=1}^C {\mob}_j =1\,, \ {\mob}_j\geq 0 \mbox{ for all } j\}\,.
\eea
Any measure $\xi_n$ with $n$ support points in $\SX_C$ can thus be represented as in \eqref{xin}, with $\wb_n$ a $n$-dimensional vector of weights attached to its support ($\wb_n\in\SP_n$ when $\xi_n\in\SM_{[1]}^+(\SX_C)$), and also as a vector $\mob_n\in\mathds{R}^C$ ($\mob_n\in\SP_C$ when $\xi_n\in\SM_{[1]}^+(\SX_C)$). For any $\xb^{(j)}\in\SX_C$ and $\xi\in\SM(\SX_C)$ with weights $\mob$, the vector of weights associated with $\xi^+(\xb^{(j)},\ma)$ equals $\mob^+(\xb^{(j)},\ma)=(1-\ma)\,\mob+\ma \eb_j$.

We shall denote
\begin{description}
  \item[$\centerdot$] $\xi^C_*$ the minimum-MMD measure in $\SM_{[1]}^+(\SX_C)$, with weights $\mob^C_*=\arg\min_{\mob\in\SP_C} \mob\TT {\Kb_\mu}_C \mob$, so that \eqref{MMD-with-Ktilde} gives
  \be\label{MC2}
    M_C^2 = \MMD_K^2(\mu,\xi^C_*) = {\mob^C_*}\TT {\Kb_\mu}_C\mob^C_* \,;
  \ee
  \item[$\centerdot$] $\widehat\xi^C$ the minimum-MMD measure in $\SM_{[1]}(\SX_C)$, with weights $\widehat\mob^C$ optimal under the total mass constraint $\1b_C\TT \widehat\mob^C$ only: $\widehat\mob^C$ is given by \eqref{hatwn} where $\1b_n$, $\Kb_n$ and $\pb_n(\mu)$ are respectively replaced by $\1b_C$, $\Kb_C$ and $\pb_C(\mu)=[P_{K,\mu}(\xb^{(1)}),\ldots,P_{K,\mu}(\xb^{(C)})]\TT$;
  \item[$\centerdot$] $\widetilde\xi^C$ the minimum-MMD unconstrained measure in $\SM(\SX_C)$, with weights $\widetilde\mob^C=\Kb_C^{-1}\pb_C(\mu)$, see~\eqref{tildewn}.
\end{description}

In the rest of the paper we derive finite-sample-size error bounds, i.e., bounds on $\MMD_K^2(\mu,\xi_k)$, for each of the constructions of Section~\ref{S:incremental-constructions} and give a numerical illustration in Section~\ref{S:examples}. Note that we are interested in situations where $k\ll C$. We start with the simplest method, the Frank-Wolfe algorithm, which has already been much studied in the literature (Section~\ref{S:performanceKH}). The derivation of the upper bound closely follows \citet{Clarkson2010}, and those arguments will be central for the derivation of the bounds for GM and SBQ in Sections~\ref{S:performanceMMD} and \ref{S:performanceSBQ}.

Table~\ref{Tb:results} summarises our results (error bounds and complexity) for the main algorithms considered. The computational complexity of KH grows like $\SO(nC)$ for $n$ iterations; the error bound decreases like $\SO((\log n)/n)$ for the standard version with step size $\ma_k=1/k$ for all $k$ (which yields uniform weights, Theorem~\ref{Th:KH-empirical}) and decreases like $\SO(1/n)$ when $\ma_k=2/(k+1)$ (Theorem~\ref{Th:KH-predefined-SZ}) or when $\ma_k$ is optimised (Theorem~\ref{Th:KH-optimal-SZ}). The error bounds for the variants OLWO and IWO also decrease like $\SO(1/n)$ (Theorem~\ref{Th:KH weighed MMD-variant2}) and their computational complexity grows like $\SO(n^2C)$. The same results apply to SBQ (Theorem~\ref{Th:SBQ12}). The numerical experiments in Section~\ref{S:examples} indicate that the error bound $\SO(1/n)$ for SBQ is pessimistic, in particular for the version where $\xb_{k+1}$ is given by \eqref{SBQ1}, some explanations for this pessimism are given in Section~\ref{S:performanceSBQ}. GM has the same complexity as KH, with a slightly better error bound for its standard version (Theorem~\ref{Th:empirical-greedy}) and similar bounds for other versions (Theorems~\ref{Th:greedy weighed MMD} and \ref{Th:greedy MMD_opt}). The case where $\SX_C$ is a random set of candidate points, possibly resampled at every iteration, is considered in Section~\ref{S:Random-candidates} and Appendix~C where we show that our results on the decrease of the error bound at finite horizon continue to apply.
In some cases, a better bound is obtained when $\widehat\xi^C=\xi^C_*$, i.e., when all weights $\widehat\omega^C_i$ are nonnegative (which occurs when $\omega_i^*>0$ for all $i$). Deriving precise sufficient conditions for this property is a difficult task (as is the question of positivity of quadrature weights in general; see \citet{KarvonenKS2019}), but it is usually satisfied when the $\xb^{(i)}$ in $\SX_C$ are independently sampled from $\mu$.

\begin{table}[h!]
\begin{center}
{\small
\caption{\small Error bound and complexity for $n$ iterations of the KH, GM and SBQ algorithms; $A_C=[\overline{K}_C^{1/2} + \tau_{1/2}(\mu)]^2$, $B_C=4\,\overline{K}_C$ ($A_C=\overline{K}_C + \tau_{1/2}^2(\mu)$ and $B_C=2\, \overline{K}_C$ when $K$ is positive), $M_C^2$ is given by \eqref{MC2}.}
\label{Tb:results}
\begin{tabular}{lcllcl}
\toprule
Method & Algorithm & & Error bound & Theorem & Complexity \\
\midrule
KH  & \ref{algo:KH-alphak}        & $\ma_k=1/k$        & $M_C^2 + B_C \, \frac{2+\log n}{n+1}$ & \ref{Th:KH-empirical}     & $\SO(nC)$ \\
    & \ref{algo:KH-alphak}        & $\ma_k=2/(k+1)$    & $M_C^2 + \frac{4\,B_C}{n+3}$          & \ref{Th:KH-predefined-SZ} & $\SO(nC)$ \\
    & \ref{algo:KH-optimal-alpha} & $\ma_k^*$          & $M_C^2 + \frac{4\,B_C}{n+3}$          & \ref{Th:KH-optimal-SZ}    & $\SO(nC)$ \\
GM  & \ref{algo:GM-alphak}        & $\ma_k=1/k$        & $M_C^2+A_C \, \frac{1+\log n}{n}$     & \ref{Th:empirical-greedy} & $\SO(nC)$ \\
    & \ref{algo:GM-alphak}        & $\ma_k=2/(k+1)$    & $M_C^2 + \frac{4\,B_C}{n+3}$          & \ref{Th:greedy weighed MMD} & $\SO(nC)$ \\
    & \ref{algo:GM-alphaopt}      & $\ma_k^*$          & $M_C^2 + \frac{4\,B_C}{n+3}$          & \ref{Th:greedy MMD_opt}     & $\SO(nC)$ \\
SBQ & Eq.\ \eqref{SBQ1}           &                    & $M_C^2+ \frac{4\,\overline{K}}{n+13/3}$ & \ref{Th:SBQ12}           & $\SO(n^2C)$ \\
    & Eq.\ \eqref{SBQ2}           &                    & $M_C^2 + \frac{4\,B_C}{n+3}$            & \ref{Th:SBQ12}           & $\SO(n^2C)$ \\

\bottomrule\\
\end{tabular}
}
\end{center}
\end{table}

\section{Performance analysis of kernel herding and its variants}\label{S:performanceKH}

\subsection{Empirical measures}\label{S:performance-KH-empirical}

Consider first the case of standard KH, corresponding to Algorithm~\ref{algo:KH-alphak} with $\ma_k=1/k$. It selects $\xi_1=\delta_{\xb_1}$, with $\xb_1\in\Arg\max_{\xb\in\SX_C} P_{K,\mu}(\xb)$, and then $\xi_{k+1}=\xi_{k}^+[\xb_{k+1},1/(k+1)]$ with $\xb_{k+1}$ given by \eqref{empirical-KH} where $\SX_C$ is substituted for $\SX$. This choice of $\ma_k$ implies that $\wb_k=\1b_k/k$ for all $k$; that is, $\xi_k=\xi_{k,e}$, the empirical measure on $\Xb_k$. The complexity only grows linearly and is $\SO(n\,C)$ for $n$ iterations: the $P_{K,\mu}(\xb^{(i)})$ are only computed once for all at the beginning, with complexity $\SO(C)$; $S_k(\xb^{(i)})=P_{K,\xi_k}(\xb^{(i)})$ is updated at each iteration for each $\xb^{(i)}$ in $\SX_C$, again with complexity $\SO(C)$. The finite-sample-size error can be bounded as indicated in Theorem~\ref{Th:KH-empirical}. The proof is given in  Appendix~B.

\begin{algorithm}[t]
{\small
\caption{Kernel herding, predefined step sizes $\ma_k$: $\xi_{k+1}=\KH(\xi_k,\ma_{k+1})$}\label{algo:KH-alphak}
\begin{algorithmic}[1]
\Require $\mu\in\SM_{[1]}^+(\SX)\cap\SM_K^{1/2}(\SX)$, $\SX_C\subset\SX$, $n\in\mathds{N}$;
\State set $S_0(\cdot)\equiv 0$ and $\xi_0=0$; compute $P_{K,\mu}(\xb)$ for all $\xb\in\SX_C$;
\State a sequence $(\ma_k)_k$ in $[0,1]$ with $\ma_1=1$;
\State $k\gets 1$
\While{$k\leq n$}
\State find $\xb_k \in \Arg\min_{\xb\in\SX_C} S_{k-1}(\xb) - \, P_{K,\mu}(\xb)$;
\State $S_k(\xb) \gets (1-\ma_k)\,S_{k-1}(\xb)+\ma_k\,K(\xb_k,\xb)$ for all $\xb\in\SX_C$;
\State $\xi_k \gets (1-\ma_k)\,\xi_{k-1}(\xb)+\ma_k\,\delta_{\xb_k}$;
\State $k\gets k+1$
\EndWhile
\State\Return $\Xb_n=[\xb_1,\ldots,\xb_n]$, $\xi_n$.
\end{algorithmic}
}
\end{algorithm}

\begin{thm}\label{Th:KH-empirical}
The empirical measure $\xi_n$ generated by Algorithm~\ref{algo:KH-alphak} with $\ma_k=1/k$ for all $k$ satisfies
\be\label{bound:KH-empirical}
\MMD_K^2(\mu,\xi_n)\leq M_C^2 + B_C \, \frac{2+\log n}{n+1} \,, \ n\geq 1\,,
\ee
where $B_C=4\,\overline{K}_C$ ($B_C=2\, \overline{K}_C$ when $K$ is positive) and $M_C^2$ is given by \eqref{MC2}.
When $\widehat\xi^C=\xi^C_*$, $\xi_n$ satisfies
\be\label{bound:KH-empirical++}
\MMD_K^2(\mu,\xi_n)\leq M_C^2 + \frac{B_C}{n} \,, \ n\geq 1\,.
\ee
\end{thm}

It may happen that the same $\xb^{(j)}$ is selected several (possibly consecutive) times at line 5 of Algorithm~\ref{Th:KH-empirical}.
One may refer to \citet{ChenMGBO2018} for the extension to the case where $\xb_{k+1}$ is searched within the whole set $\SX$ and the selection is suboptimal with some bounded error. \citet{ChenWS2010} show that the error can decrease as $\SO(n^{-2})$ when $\SH_K$ is finite-dimensional, but \citet{BachLJO2012} indicate that one can only expect the rate $\SO(n^{-1})$ in the infinite-dimensional situation; see also \citet[Appendix~A]{PZ2020-SIAM}.
In the next section, we show that a better convergence rate than \eqref{bound:KH-empirical}, without the log term, can be obtained in general (without the assumption that $\widehat\xi^C=\xi^C_*$) when we allow $\xi_k$ to be nonuniform on $\Xb_k$. The arguments are similar to those used for the proof of Theorem~\ref{Th:KH-empirical}: exploiting the convexity of $\MMD_K^2(\mu,\xi_n)$ with respect to the vector of weights $\mob_n$, we obtain a recurrence equation which imposes a particular decrease, see Lemma~\ref{L:induction} in Appendix~B. The same arguments are used for the other algorithms in the following sections.

\subsection{Nonuniform weights}\label{S:KH-nonuniform}

Next theorem shows that for a suitable predefined step-size sequence $(\ma_k)_k$ in Algorithm~\ref{algo:KH-alphak}, the squared MMD decreases as $\SO(n^{-1})$. The proof is in Appendix~B.

\begin{thm}\label{Th:KH-predefined-SZ}
The measure $\xi_n$ generated with Algorithm~\ref{algo:KH-alphak} with $\ma_k=2/(k+1)$ for all $k$ satisfies
\be\label{bound:KH-predefined-SZ}
\MMD_K^2(\mu,\xi_n)\leq M_C^2 + \frac{4\,B_C}{n+3} \,, \ n\geq 1\,.
\ee
\end{thm}

Again, it may happen that the same $\xb^{(j)}$ is selected several times at line 5 of Algorithm~\ref{Th:KH-empirical}; that is, there may exist repetitions in $\Xb_k$.
The weights $\{\wb_n\}_i$ that $\xi_n$ allocates to the $\xb_i$, $i=1,\ldots,n$, can be computed explicitly.
When $\ma_k=2/(k+1)$ for all $k$, we have $\xi_n= \sum_{i=1}^n 2i/[n(n+1)]\, \delta_{\xb_i}$.
The distribution is thus far from being uniform, contrary to the case with $\ma_k=1/k$; see the right panel of Figure~\ref{F:EX1-KH}.
When the condition $\widehat\xi^C=\xi^C_*$ is satisfied in Theorem~\ref{Th:KH-empirical}, the bound \eqref{bound:KH-empirical++} is better than \eqref{bound:KH-predefined-SZ} and there is no point in using $\ma_k=2/(k+1)$ rather than $\ma_k=1/k$.
Example~1 will illustrate that the decrease of $\MMD_K(\mu,\xi_k)$ may be worse for nonuniform weights; see Figure~\ref{F:EX1-KH}-left.

As a further attempt to improve performance, we can select $\xb_{k+1}\in\Arg\min_{\xb\in\SX_C} [ P_{K,\xi_k}(\xb) - P_{K,\mu}(\xb)]$ as previously, and then optimise $\MMD_K[\mu,\xi_k^+(\xb_{k+1},\ma)]$ with respect to $\ma$ in $[0,1]$.
This function being quadratic in $\ma$, the optimal value $\ma_{k+1}^*=\ma_{k+1}^*(\xb_{k+1})$ can be obtained explicitly; the construction is summarised in Algorithm~\ref{algo:KH-optimal-alpha} (see the proof of Theorem~\ref{Th:KH-optimal-SZ} in Appendix~B for details). Again, the complexity grows linearly with $n$. The use of the optimal $\ma$ implies that the same $\xb^{(j)}$ cannot be selected two consecutive times at line 4 of Algorithm~\ref{algo:KH-optimal-alpha}.

\begin{algorithm}[t]
{\small
\caption{Kernel herding, optimal step sizes: $\xi_{k+1}=\KH(\xi_k,\ma_{k+1}^*)$}\label{algo:KH-optimal-alpha}
\begin{algorithmic}[1]
\Require $\mu\in\SM_{[1]}^+(\SX)\cap\SM_K^{1/2}(\SX)$, $\SX_C\subset\SX$, $n\in\mathds{N}$;
\State set $S_0(\cdot)\equiv 0$ and $\xi_0=0$; compute $P_{K,\mu}(\xb)$ for all $\xb\in\SX_C$;
\State $k\gets 1$
\While{$k\leq n$}
\State find $\xb_k\in\Arg\min_{\xb\in\SX_C} S_{k-1}(\xb)-P_{K,\mu}(\xb)$;
\If{$k=1$} set $\ma_k^*=1$, $Q_1=K(\xb_1,\xb_1)$, $R_1=P_{K,\mu}(\xb_1)$;
\Else{ compute $A_k=Q_{k-1}-R_{k-1}+P_{K,\mu}(\xb_k)-S_{k-1}(\xb_k)$, \\
\qquad\qquad\qquad\qquad  $B_k=Q_{k-1}-2\,S_{k-1}(\xb_k)+K(\xb_k,\xb_k)$, \\
\qquad\qquad\qquad\qquad  and $\ma_k^*=\min\{A_k/B_k,1\}$}
\EndIf
\If{$\ma_k^* = 0$} \Return $\Xb_{k-1}=[\xb_1,\ldots,\xb_{k-1}]$, $\xi_{k-1}$ and stop; \EndIf
\State $R_k \gets (1-\ma_k^*)\,R_{k-1}+\ma_k^*\,P_{K,\mu}(\xb_k)$;
\State $Q_k \gets (1-\ma_k^*)^2\,Q_{k-1}+2\,\ma_k^*(1-\ma_k^*)S_{k-1}(\xb_k)+(\ma_k^*)^2 K(\xb_k,\xb_k)$;
\State $S_k(\xb) \gets (1-\ma_k^*)\,S_{k-1}(\xb)+\ma_k^*\,K(\xb_k,\xb)$ for all $\xb\in\SX_C$;
\State $\xi_k \gets (1-\ma_k^*)\,\xi_{k-1}+\ma_k^*\,\delta_{\xb_k}$;
\State $k\gets k+1$
\EndWhile
\State\Return $\Xb_n=[\xb_1,\ldots,\xb_n]$, $\xi_n$.
\end{algorithmic}
}
\end{algorithm}

\begin{thm}\label{Th:KH-optimal-SZ}
The measure $\xi_n$ generated with Algorithm~\ref{algo:KH-optimal-alpha} satisfies \eqref{bound:KH-predefined-SZ}; when $\widehat\xi^C=\xi^C_*$ it satisfies \eqref{bound:KH-empirical++}. The optimal $\ma$ at iteration $k$ is $\ma_{k+1}^*=\min\{\widehat\ma_{k+1},1\}$ with
\be\label{alpha_k+1^*}
\widehat\ma_{k+1}=\widehat\ma_{k+1}(\xb_{k+1})=\frac{\sum_{i=1}^k \{\wb_k\}_i \left[P_{K,\xi_k}(\xb_i)-P_{K,\mu}(\xb_i)\right] + P_{K,\mu}(\xb_{k+1})-P_{K,\xi_k}(\xb_{k+1})}{\sum_{i=1}^k \{\wb_k\}_i P_{K,\xi_k}(\xb_i) - 2\, P_{K,\xi_k}(\xb_{k+1}) + K(\xb_{k+1},\xb_{k+1})}
\ee
where $\xb_{k+1}\in\Arg\min_{\xb\in\SX_C} \left[ P_{K,\xi_n}(\xb)-P_{K,\mu}(\xb)\right]$.
If the algorithm stops at iteration $k$ with $\widehat\ma_{k+1}= 0$, then $\MMD_K(\mu,\xi_k)=\MMD_K(\mu,\xi^C_*)$.
\end{thm}

\vsp
It may seem surprising that the bound obtained with optimal step sizes is not better than when $\ma_k=2/(k+1)$ for all $k$ in Algorithm~\ref{algo:KH-alphak}, since the decrease of MMD is larger in the former case when starting from the same $\xi_k$. However, the global decrease over many iterations with the optimal $\ma$ is not necessarily better than with a predefined step-size sequence; one can refer to \citet{Dunn80} for a discussion. A numerical comparison in provided in Section~\ref{S:examples}, showing that Algorithm~\ref{algo:KH-optimal-alpha} may perform worse than Algorithm~\ref{algo:KH-alphak}; see the left panel of Figure~\ref{F:EX1-KH}.

\begin{rmk}\label{R:away-steps} \citet{DunnH78} and \citet{Dunn80} propose other choices of step-size sequences which we do not consider here.
We also do not consider Frank-Wolfe algorithm with away steps \citep{Wolfe70, Atwood73}\footnote{See also \citet{ToddY2007, AhipasaogluST2008} for a recent use in the minimum-volume ellipsoid problem.}, for which $\xi_{k+1}=\xi_k+\ma_{k+1}(\xi_k-\delta_{\xb_{j_k}})$ moves away from one of its support points $\xb_{j_k}$. Here $\xb_{j_k}$ is taken in $\Arg\max_{\xb\in\supp(\xi_k)} F_{\MMD_K^2}(\xi_k,\delta_{\xb}) =  \Arg\max_{\xb\in\SX} \left[ P_{K,\xi_k}(\xb)-P_{K,\mu}(\xb)\right]$ and $\ma_{k+1}\in[0,\xi_k(\xb_{j_k})/[1-\xi_k(\xb_{j_k})]$ to ensure that $\xi_{k+1}(\xb_{j_k})\geq 0$; the decision to use an away step instead of $\xi_{k+1}=\xi_k^+(\xb_{k+1},\ma)$ with $\xb_{k+1}$ given by
\eqref{xk+1-KH} can rely on the comparison between the criterion values obtained, or on the comparison between the absolute values of the directional derivatives $|F_{\MMD_K^2}(\xi_k,\delta_{\xb_{j_k}})|$ and $|F_{\MMD_K^2}(\xi_k,\delta_{\xb_{k+1}})|$.
Neither do we consider vertex-exchange methods, for which $\xi_{k+1}=\xi_k+\ma_{k+1}(\delta_{\xb_{k+1}}-\delta_{\xb_{j_k}})$ for $\ma_{k+1}\in[0,\xi_k(\xb_{j_k})]$; see for instance \citet[Appendix~A.3]{PZ2020-SIAM}, \citet[Chap.~9]{PP2013} and the references therein. These methods prove especially efficient for design problems for which the optimal solution is attained on the boundary of $\SP_C$, with many components equal to zero, in particular due to their ability to reduce the support of the current measure (when $\ma_{k+1}=1$). The situation is different for the type of problems we have in mind here, and we can only expect a rate of decrease of the finite-sample-size error similar to Algorithm~\ref{algo:KH-optimal-alpha}. \citet{LacosteJJ2015} give a precise analysis of the convergence of these variants of Frank-Wolfe algorithm and prove that they have a global linear convergence rate (contrary to the original Frank-Wolfe algorithm\footnote{Linear convergence is obtained for the Frank-Wolfe algorithm under the condition that $\widehat\mob^C$ is in the interior of $\SP_C$; but even in this favourable case the result has no practical interest for large $C$; see  \citet[Lemma~A4]{PZ2020-SIAM}.}). However, the pyramidal width defined in the same paper (eq.\ (9)) decreases as $C^{-1/2}$ and the constant $\rho$ in the linear convergence factor $\exp(-\rho k)$ decreases as $1/C$.
\fin
\end{rmk}

\subsection{KH with off-line and integrated weight optimisation}\label{S:KH-OLWO-IWO}

\subsubsection{Off-Line Weight Optimisation (OLWO)}\label{S:off-line-KH}
The first KH variant mentioned in Section~\ref{S:greedyMMD} (Frank-Wolfe Bayesian quadrature, \citet{BriolOGO2015}) uses the support $\Xb_k=[\xb_1,\ldots,\xb_k]$ obtained at iteration $k$ with Algorithm~\ref{algo:KH-alphak} or \ref{algo:KH-optimal-alpha}, and constructs an optimal measure $\xi_k^*$, $\widehat\xi_k$, or $\widetilde\xi_k$, respectively in $\SM_{[1]}^+(\Xb_k)$, $\SM_{[1]}(\Xb_k)$ or $\SM(\Xb_k)$, with respective weights $\wb_k^*$, obtained as solution of a QP problem, $\widehat\wb_k$ given by \eqref{hatwn}, and $\widetilde\wb_k$ given by \eqref{tildewn}. Let $\xi_k$ be the measure generated by Algorithm~\ref{algo:KH-alphak} or \ref{algo:KH-optimal-alpha}, with support $\Xb_k$; since $\xi_k$ is a probability measure,
$\MMD_K^2(\mu,\widetilde\xi_k)\leq \MMD_K^2(\mu,\widehat\xi_k)\leq \MMD_K^2(\mu,\xi_k^*) \leq \MMD_K^2(\mu,\xi_k)$ for all $k$,
and the bounds of Theorems~\ref{Th:KH-empirical}-\ref{Th:KH-optimal-SZ} remain valid.

\subsubsection{Integrated Weight Optimisation (IWO)}\label{S:integrated-KH}
The situation is more complicated for the second variant of Section~\ref{S:greedyMMD}, where we substitute $\nu_k\in\{\xi_k^*,\widehat\xi_k,\widetilde\xi_k\}$ for $\xi_k$ \emph{at every iteration} (the case $\nu_k=\xi_k^*$ corresponds to the fully-corrective Frank-Wolfe algorithm, we do not detail the minimum-norm point algorithm, see Remark~\ref{R:MNP}~below). We only consider the situation where the same choice is applied for all iterations and denote respectively by (\textit{i}), (\textit{ii}) and (\textit{iii}) the three cases $\nu_k=\xi_k^*$, $\nu_k=\widehat\xi_k$ and $\nu_k=\widetilde\xi_k$ for all $k$. The choice of $\xb_{k+1}$ is the same as for KH, but now there is no $\ma_{k+1}$ to choose. The construction is summarised in Algorithm~\ref{algo:KH-IWO}.

\begin{algorithm}[t]
{\small
\caption{Kernel herding + IWO (\textit{i}), (\textit{ii}) and (\textit{iii})}
\label{algo:KH-IWO}
\begin{algorithmic}[1]
\Require $\mu\in\SM_{[1]}^+(\SX)\cap\SM_K^{1/2}(\SX)$, $\SX_C\subset\SX$, $n\in\mathds{N}$;
\State set $S_0(\cdot)\equiv 0$ and $\xi_0=0$; compute $P_{K,\mu}(\xb)$ for all $\xb\in\SX_C$;
\State $k\gets 1$
\While{$k\leq n$}
\State find $\xb_k \in \Arg\min_{\xb\in\SX_C} S_{k-1}(\xb) - \, P_{K,\mu}(\xb)$;
\State compute (\textit{i}) $\wb_k=\wb_k^*$ (a QP problem), or (\textit{ii}) $\wb_k=\widehat\wb_k$ \eqref{hatwn}, or (\textit{iii}) $\wb_k=\widetilde\wb_k$ \eqref{tildewn},
\State compute $S_k(\xb)=\sum_{i=1}^k \{\wb_k\}_i\,K(\xb,\xb_i)$ for all $\xb\in\SX_C$;
\State $\xi_k=\sum_{i=1}^k \{\wb_k\}_i\, \delta_{\xb_i}$;
\State $k\gets k+1$
\EndWhile
\State\Return $\Xb_n=[\xb_1,\ldots,\xb_n]$, $\xi_n$.
\end{algorithmic}
}
\end{algorithm}

Note that $S_k(\cdot)=P_{K,\nu_k}(\cdot)$ can no longer be computed recursively, so that the complexity
grows faster than linearly: at iteration $k$, the complexity of the determination of $\wb_k^*$, $\widehat\wb_k$ or $\widetilde\wb_k$ is $\SO(m(k))$, independently of $C$ (with, in the last two cases, $m(k)=k^3$ in general and $m(k)=k^2$ if rank-one updating is used to compute $\Kb_k^{-1}$ in \eqref{hatwn} and \eqref{tildewn}; see Remark~\ref{R:invKn}); the complexity of the computation of all $S_k(\xb^{(i)})$ is $\SO(k\,C)$ and the complexity for $n$ iterations is thus $\SO(n^2\,C)$ for $n\ll C$.
Kernel herding with IWO satisfies the error bounds in Theorem~\ref{Th:KH weighed MMD-variant2}; the proof is in Appendix~B.

\begin{thm}\label{Th:KH weighed MMD-variant2}\mbox{}
The measure $\xi_n$ generated by Algorithm~\ref{algo:KH-IWO}-(\textit{i}) satisfies \eqref{bound:KH-predefined-SZ}; when $\widehat\xi^C=\xi^C_*$, it satisfies \eqref{bound:KH-empirical++}.
When using Algorithm~\ref{algo:KH-IWO}-(\textit{ii}), $\xi_n$ satisfies \eqref{bound:KH-predefined-SZ}; when $\widehat\xi^C=\xi^C_*$, it satisfies
\be\label{KH-ii}
\MMD_K^2(\mu,\xi_n) \leq M_C^2+ \frac{B_C}{n+2}\,, \ n\geq 2 \,,
\ee
where $B_C=4\,\overline{K}_C$ ($B_C=2\, \overline{K}_C$ when $K$ is positive) and $M_C^2$ is given by \eqref{MC2}.
When using Algorithm~\ref{algo:KH-IWO}-(\textit{iii}), $\xi_n$ satisfies
\be\label{KH-iii}
\MMD_K^2(\mu,\xi_n) \leq M_C^2+ \frac{4\,\overline{K}}{n+\frac{4\,\overline{K}}{\MMD_K^2(\mu,\xi_1)}-1} \leq M_C^2+ \frac{4\,\overline{K}}{n+13/3}\,, \ n\geq 1 \,.
\ee
\end{thm}

\begin{rmk}\label{R:stopping}
The measures used in Algorithms~\ref{algo:KH-IWO}-(\textit{ii}) and (\textit{iii}) are not constrained to belong to $\SM_{[1]}^+(\SX_C)$, so that the algorithm can still progress when $\MMD_K^2(\mu,\xi_k)\leq\MMD_K^2(\mu,\xi^C_*)=M_C^2$ (an obvious indication of the pessimism of the bounds in Theorem~\ref{Th:KH weighed MMD-variant2}). We show in Appendix~B that the following stopping condition can be added after Step 4 of IWO (\textit{ii}) and (\textit{iii}), respectively:
\begin{description}
  \item[4'-(\textit{ii}):] {\bf if} $S_{k-1}(\xb_k)-P_{K,\mu}(\xb_k)\geq S_{k-1}(\xb_{k-1})-P_{K,\mu}(\xb_{k-1})$ {\bf then return} $\Xb_{k-1}$, $\xi_{k-1}$ and stop;
    \item[4'-(\textit{iii}):] {\bf if} $S_{k-1}(\xb_k)-P_{K,\mu}(\xb_k)\geq 0$ {\bf then return} $\Xb_{k-1}$, $\xi_{k-1}$ and stop; \fin
\end{description}
\end{rmk}

\begin{rmk}\label{R:MNP}
Algorithm~\ref{algo:KH-IWO}-(\textit{i}) corresponds to the fully-corrective Frank-Wolfe algorithm analysed in \citep{LacosteJJ2015}. The Minimum-norm point variant, based on \citep{Wolfe76}, uses a sequence of affine projections based on the calculation of a sequence $\widehat\mob_{k_i}$ restricted to give nonzero weights to subsets $\SS_{k_i}$ of $\SS_{k_0}=\supp(\xi_k)$ (at most $k$ weights $\widehat\mob_{k_i}$ need to be calculated); see Algorithm~5 in \citep{LacosteJJ2015}. Since the $\widehat\mob_{k_i}$ can be obtained explicitly through \eqref{hatwn}, this construction is simpler than the fully-corrective Frank-Wolfe algorithm. The bounds in Theorem~\ref{Th:KH weighed MMD-variant2} indicated for Algorithm~\ref{algo:KH-IWO}-(\textit{i}) continue to apply, since we still have $\Delta_C(\xi_{k+1})\leq (1-\ma_{k+1})\,\Delta_C(\xi_k)+B_C\,\ma_{k+1}^2$, $k\geq 1$, for $\ma_{k+1}=2/(k+2)$, and $\Delta_C(\xi_{k+1})\leq (1-2\,\ma_{k+1})\,\Delta_C(\xi_k)+B_C\,\ma_{k+1}^2$, $k\geq 1$, for $\ma_{k+1}=1/(k+1)$ when $\widehat\xi^C=\xi^C_*$; see the proofs of Theorems~\ref{Th:KH-empirical} and \ref{Th:KH weighed MMD-variant2}.
\fin
\end{rmk}

\begin{rmk}\label{R:invKn}
OLWO and IWO require the repeated computation of optimal weights $\widehat\wb_n$ or $\widetilde\wb_n$, respectively given by \eqref{hatwn} and \eqref{tildewn},
for which it is advantageous to use the block matrix inversion \eqref{Kbnm1}. Rank-one Cholesky updates can be used too; the details are omitted. Eq.\ \eqref{Kbnm1} also gives
\bea
\kb_{n+1}\TT(\xb)\Kb_{n+1}^{-1}\kb_{n+1}(\xb) &=& \kb_n\TT(\xb)\Kb_n^{-1}\kb_n(\xb)+\beta_{n+1}\,[(\ub_{n+1}\TT,\ -1)\kb_{n+1}(\xb)]^2 \,, \\
\1b_{n+1}\TT\Kb_{n+1}^{-1}\kb_{n+1}(\xb) &=& \1b_n\TT\Kb_n^{-1}\kb_n(\xb)+\beta_{n+1}\,[(\ub_{n+1}\TT,\ -1)\1b_{n+1}][(\ub_{n+1}\TT,\ -1)\kb_{n+1}(\xb)]
\eea
for all $\xb$, so that matrix-vector multiplications can be avoided in SBQ when computing the denominators on the right-hand side of \eqref{SBQ1} and \eqref{SBQ2} by using recursive calculation.
\fin
\end{rmk}

\begin{rmk}\label{R:SBQ1}
The numerical experiments of Section~\ref{S:examples} show that the bound \eqref{KH-iii} for Algorithm~3-(\textit{iii}) becomes very loose when $k$ increases. The arguments used in the proof of Theorem~\ref{Th:KH weighed MMD-variant2} suggest two sources of pessimism. First, we substitute the inequality \eqref{Th4*2} for the convexity bound \eqref{Th4*1} (this approximation is used for all the methods considered). Second, we ignore the decrease of $K(\xb_{k+1},\xb_{k+1})-\kb_k\TT(\xb_{k+1})\Kb_k^{-1}\kb_k(\xb_{k+1})$ as $k$ increases in the denominator on the right-hand side of \eqref{MMD-SBQ1}, and simply bound it by $\overline{K}$. In the algorithm defined by \eqref{SBQ3}, the denominator is constant, so that the pessimism of the error bound is mainly due to the substitution of \eqref{Th4*2} for \eqref{Th4*1}; this effect is illustrated on Figure~\ref{F:EX1-KH-B}-left. Although this indicates that there still exists room for improvement, the derivation of better bounds seems difficult.
Similar statements can be made for Algorithm~3-(\textit{ii}), for which numerical experiments show that it performs similarly to Algorithm~\ref{algo:KH-alphak} for moderate $k$, but tends to converge faster for large $k$ (see Figure~\ref{F:EX1-KH-B}-left and Figure~\ref{F:Ex2-MMD-KH-GM}-left).
\fin
\end{rmk}
\section{Performance analysis of Greedy MMD Minimisation (GM)}\label{S:performanceMMD}

\subsection{Empirical measures}\label{S:performance-MMD-empirical}

GM with empirical measures corresponds to Algorithm~\ref{algo:GM-alphak} with $\ma_k=1/k$ for all $k$; it selects $\xb_1\in\Arg\min_{\xb\in\SX_C} K(\xb,\xb)-2\,P_{K,\mu}(\xb)$ and then chooses $\xb_{n+1}$ according to \eqref{empirical-greedy} with $\SX_C$ substituted for $\SX$.
It corresponds to Algorithm~1 in \citep{TeymurGRO2021}, where the authors derive a finite-sample-size error bound using the RKHS framework. Taking advantage of the finiteness of the candidate set $\SX_C$, we provide a simpler proof using only linear (finite-dimensional) algebra; see Appendix~B. Notice that the bound is smaller than for KH in Theorem~\ref{Th:KH-empirical}.
The complexity of Algorithm~\ref{algo:GM-alphak} is $\SO(n\,C)$ for $n$ iterations: the $K(\xb^{(i)},\xb^{(i)})$ and $P_{K,\mu}(\xb^{(i)})$ are only computed once for all at the beginning, with complexity $\SO(C)$, the $S_k(\xb^{(i)})$ are updated at each iteration for all $\xb^{(i)}\in\SX_C$, again with complexity $\SO(C)$.

\begin{algorithm}[t]
{\small
\caption{Greedy MMD minimisation, predefined step sizes $\ma_k$: $\xi_{k+1}=\GM(\xi_k,\ma_{k+1})$}\label{algo:GM-alphak}
\begin{algorithmic}[1]
\Require $\mu\in\SM_{[1]}^+(\SX)\cap\SM_K^{1/2}(\SX)$, $\SX_C\subset\SX$, $n\in\mathds{N}$;
\State set $S_0(\cdot)\equiv 0$ and $\xi_0=0$; compute $K(\xb,\xb)$ and $P_{K,\mu}(\xb)$ for all $\xb\in\SX_C$;
\State a sequence $(\ma_k)_k$ in $[0,1]$ with $\ma_1=1$; 
\State $k\gets 1$
\While{$k\leq n$}
\State find $\xb_k \in \Arg\min_{\xb\in\SX_C} 2(1-\ma_k)\, S_{k-1}(\xb) + \ma_k\, K(\xb,\xb) - 2\, P_{K,\mu}(\xb)$;
\State $S_k(\xb) \gets (1-\ma_k)\,S_{k-1}(\xb)+\ma_k\,K(\xb_k,\xb)$ for all $\xb\in\SX_C$;
\State $\xi_k \gets (1-\ma_k)\,\xi_{k-1}(\xb)+\ma_k\,\delta_{\xb_k}$;
\State $k\gets k+1$
\EndWhile
\State\Return $\Xb_n=[\xb_1,\ldots,\xb_n]$, $\xi_n$. 
\end{algorithmic}
}
\end{algorithm}

\begin{thm}\label{Th:empirical-greedy} The measure $\xi_n$ generated by Algorithm~\ref{algo:GM-alphak} with $\ma_k=1/k$ for all $k$ satisfies
\be\label{induction-1}
\MMD_K^2(\mu,\xi_n)\leq M_C^2+A_C \, \frac{1+\log n}{n} \,, \ n\geq 1\,,
\ee
where $M_C^2$ is given by \eqref{MC2} and $A_C=[\overline{K}_C^{1/2} + \tau_{1/2}(\mu)]^2$ ($A_C=\overline{K}_C + \tau_{1/2}^2(\mu)$ when $K$ is positive).
\end{thm}

\subsection{Nonuniform weights}\label{S:MMD-nonuniform}

Consider now the case of general discrete measures $\xi_n$ in $\SM_{[1]}^+(\SX_C)$, see \eqref{xin}.
We show that allowing nonuniform weights in GM yields a faster decrease of $\MMD_K(\mu,\xi_n)$.
As for KH, we consider iterations of the form
$\xi_{k+1}=\xi_k^+(\xb_{k+1},\ma_{k+1})$, $k\geq 1$, for some $\ma_{k+1}\in[0,1]$ and $\xb_{k+1}\in\SX_C$, where $\xi_k^+(\xb,\ma)$ is defined by \eqref{xik+}.
We first consider the same choice $\ma_k=2/(k+1)$ as in Section~\ref{S:KH-nonuniform}. The proof of Theorem~\ref{Th:greedy weighed MMD} is in Appendix~B.

\begin{thm}\label{Th:greedy weighed MMD}
The measure $\xi_n$ generated by Algorithm~\ref{algo:GM-alphak} with $\ma_k=2/(k+1)$ for all $k$ satisfies \eqref{bound:KH-predefined-SZ}. When $\widehat\xi^C=\xi^C_*$, Algorithm~\ref{algo:GM-alphak} with $\ma_k=1/k$ for all $k$ yields \eqref{bound:KH-empirical++}.
\end{thm}

\begin{rmk}
As for Algorithm~\ref{algo:KH-alphak}, when $\ma_k=2/(k+1)$ in Algorithm~\ref{algo:GM-alphak} the measure $\xi_n$ is not uniform on its support $\Xb_n$.
It is uniform when $\ma_k=1/k$ for all $k$, but the arguments used in the proof of Theorem~\ref{Th:greedy weighed MMD} only give \eqref{bound:KH-empirical}, which is worse than \eqref{induction-1} obtained by \citet{TeymurGRO2021}.
\fin
\end{rmk}

Consider now GM with optimal step size, which selects $\ma_{k+1}$ and $\xb_{k+1}$ optimally at each iteration:
$[\xb_{k+1},\ma_{k+1}] \in\Arg\min_{\xb\in\SX_C,\,\ma\in[0,1]} \MMD_K[\mu,\xi_k^+(\xb,\ma)]$, with $\xi_1=\delta_{\xb_1}$ and $\xb_1\in\Arg\min_{\xb\in\SX_C} K(\xb,\xb)-2\,P_{K,\mu}(\xb)$.
As in Algorithm~\ref{algo:KH-optimal-alpha}, the optimal value of $\ma_{k+1}=\ma(\xb_{k+1})$ is obtained explicitly, see the proof of Theorem~\ref{Th:greedy MMD_opt} in Appendix~B for details.
The complexity is again $\SO(n\,C)$ for $n$ iterations (it is larger than for Algorithm~\ref{algo:KH-optimal-alpha} as $\ma(\xb)$ must be calculated for all $\xb\in\SX_C$).

\begin{algorithm}[t]
{\small
\caption{Greedy MMD minimisation, optimal step sizes: $\xi_{k+1}\GM(\xi_k,\ma_{k+1}^*)$}\label{algo:GM-alphaopt}
\begin{algorithmic}[1]
\Require $\mu\in\SM_{[1]}^+(\SX)\cap\SM_K^{1/2}(\SX)$, $\SX_C\subset\SX$, $n\in\mathds{N}$;
\State compute $K(\xb,\xb)$ and $P_{K,\mu}(\xb)$ for all $\xb\in\SX_C$;
\State set $S_0(\cdot)\equiv 0$, $Q_0=R_0=0$, $\ma_1(\cdot)\equiv 1$ and $\xi_0=0$;
\State set $A_0(\xb)=P_{K,\mu}(\xb)$, $B_0(\xb)=K(\xb,\xb)$ for all $\xb\in\SX_C$;
\State $k\gets 1$
\While{$k\leq n$}
\State find $\xb_k\in\Arg\min_{\xb\in\SX_C} \ma^2(\xb)B_{k-1}(\xb)-2\,\ma(\xb) A_{k-1}(\xb)$;
\State $\ma_k \gets \ma(\xb_k)$
\State $R_k \gets (1-\ma_k)\,R_{k-1}+\ma_k\,P_{K,\mu}(\xb_k)$;
\State $Q_k \gets (1-\ma_k)^2\,Q_{k-1}+2\,\ma_k(1-\ma_k)S_{k-1}(\xb_k)+\ma_k^2 K(\xb_k,\xb_k)$;
\State $S_k(\xb) \gets (1-\ma_k)\,S_{k-1}(\xb)+\ma_k\,K(\xb_k,\xb)$ for all $\xb\in\SX_C$;
\State $\xi_k \gets (1-\ma_k)\,\xi_k+\ma_k\,\delta_{\xb_k}$;
\State compute $A_k(\xb)=Q_k-R_k+P_{K,\mu}(\xb)-S_k(\xb)$, $B_k(\xb)=Q_k-2\,S_k(\xb)+K(\xb,\xb)$ \\
\qquad\qquad and $\ma(\xb)=\max\{0,\min\{A(\xb)/B(\xb),1\}\}$ for all $\xb\in\SX_C$;
\If{all $\ma(\xb)$ equal = 0} \Return $\Xb_k=[\xb_1,\ldots,\xb_k]$, $\xi_k$ and stop; \EndIf
\State $k\gets k+1$
\EndWhile
\State\Return $\Xb_n=[\xb_1,\ldots,\xb_n]$, $\xi_n$.
\end{algorithmic}
}
\end{algorithm}

\begin{thm}\label{Th:greedy MMD_opt}
The measure $\xi_n$ generated with Algorithm~\ref{algo:GM-alphaopt} satisfies \eqref{bound:KH-predefined-SZ}. When $\widehat\xi^C=\xi^C_*$, it satisfies \eqref{bound:KH-empirical++}.
\end{thm}

Similarly to Theorem~\ref{Th:KH-optimal-SZ}, one might think that bounds obtained with predefined step sizes should be overly loose concerning an algorithm for which $\ma_k$ is optimised at every iteration. However, the observed behaviour is often similar to that of Algorithm~\ref{algo:GM-alphak}, if not worse, see the examples in Section~\ref{S:examples}, indicating that optimal but myopic steps are not necessarily preferable to myopic, non-optimised but suitably chosen steps; see, e.g., \citet{ZPB_OL-2012} for an illustration with the steepest descent algorithm.

\begin{rmk}
As for KH, one may also consider OLWO and IWO variants of GM; see Section~\ref{S:KH-OLWO-IWO}. The OLWO variant does not raise any particular difficulty as it runs in parallel and does not affect the algorithm: like in Section~\ref{S:off-line-KH} for KH, OLWO can only improve performance. The situation is different for IWO: the fact that the next point $\xb_{k+1}$ and the step size $\ma_{k+1}$ must be selected simultaneously render its use less adapted than with KH, which maximises the right-hand side of \eqref{SBQ1} and \eqref{SBQ2}, see the proof of Theorem~\ref{Th:KH weighed MMD-variant2}.
\fin
\end{rmk}

\section{Performance analysis of SBQ}\label{S:performanceSBQ}

We suppose again that the successive support points are searched within a finite candidate set $\SX_C$, and consider the two versions of SBQ presented in Section~\ref{S:SBQ} with $\SX_C$ substituted for $\SX$. We do not detail the algorithm which simply implements \eqref{SBQ1} or \eqref{SBQ2}---with $\Kb_k^{-1}$ calculated recursively as indicated in Remark~\ref{R:invKn}.
The numerical experiments of Section~\ref{S:examples} show that the two versions behave similarly to Algorithms~3-(\textit{iii}) and (\textit{ii}), respectively, but have a slightly higher computational cost.

We also consider a version of SBQ where all previous weights $\{\widetilde\wb_k\}_i$ are kept fixed, $i=1,\ldots,k$, and only the next one is optimised (without constraint) when choosing $\xb_{k+1}$. Since $\MMD_K^2(\mu,\xi_k+w\,\delta_\xb)=\MMD_K^2(\mu,\xi_k)+w^2\,K(\xb,\xb)+2w\, [P_{K,\xi_k}(\xb)-P_{K,\mu}(\xb)]$, the optimal $w$ is $w^*(\xb)=[P_{K,\mu}(\xb)-P_{K,\widetilde\xi_k}(\xb)]/K(\xb,\xb)$. This algorithm selects $\xb_1\in\Arg\max_{\xb\in\SX_C} P_{K,\mu}^2(\xb)/K(\xb,\xb)$ and then uses
\be\label{SBQ3}
\xb_{k+1} \in\Arg\max_{\xb\in\SX_C} \frac{\left[P_{K,\widetilde\xi_k}(\xb)-P_{K,\mu}(\xb)\right]^2} {K(\xb,\xb)} \,, \ \xi_{k+1}=\xi_k+w^*(\xb_{k+1})\delta_{\xb_{k+1}}\,, \ k\geq 1\,.
\ee
When $K(\xb,\xb)$ is a constant, the choice of $\xb_{k+1}$ is similar to that of KH, see \eqref{xk+1-KH}.
It corresponds to a Coordinate-Descent (CD) algorithm (see, e.g., \cite{Wright2015}) operating on the weights $\mob=(\mob_1,\ldots,\mob_C)\in\mathds{R}^C$; see Section~\ref{S:notation}. Performance bounds for these three versions of SBQ are given in Theorem~\ref{Th:SBQ12}.

\begin{thm}\label{Th:SBQ12}
Suppose that $\SX_C$ is substituted for $\SX$. Then, $\MMD_K^2(\mu,\xi_n)$ satisfies the same bounds as those indicated in Theorem~\ref{Th:KH weighed MMD-variant2} for Algorithm~\ref{algo:KH-IWO}-(\textit{ii}) when using \eqref{SBQ2}, and satisfies \eqref{KH-iii} when using \eqref{SBQ1}, or when using \eqref{SBQ3} if $K(\xb,\xb)$ is a constant.
\end{thm}

Although our numerical experiments indicate that \eqref{SBQ3} is not competitive compared to \eqref{SBQ1}, the analysis of its finite sample error helps understanding the pessimism of the error bound derived for version \eqref{SBQ1} of SBQ; see Remark~\ref{R:SBQ1} and the proof of Theorem~\ref{Th:SBQ12}.

\section{Random candidate sets}\label{S:Random-candidates}

The extension of the results in previous sections to the case where $\SX_C$ corresponds to $C$ points independently sampled from $\mu$ is fairly simple; see \citet{TeymurGRO2021}. For instance, \eqref{bound:KH-predefined-SZ} becomes
\bea
\Ex_\mu\{ \MMD_K^2(\mu,\xi_n)\} \leq \Ex_\mu\{M_C^2\} + \frac{4\, \Ex_\mu\{B_C\}}{n+3} \,, \ n\geq 1\,.
\eea
We thus only have to bound the expected values of the constants $A_C$, $B_C$ and $M_C^2$ that intervene in the various bounds that have been presented. Their values are given in the following lemma, the proof is in Appendix~B.

\begin{lemma}\label{L:XC-random} Suppose $\mu\in\SM_K^1(\SX)$ and that the $C$ points in $\SX_C$ are independently sampled from $\mu$, then
\bea
\Ex_\mu\{A_C\} &\leq& A(\mu)=[\overline{K}^{1/2} + \tau_{1/2}(\mu)]^2 \ (A(\mu)=\overline{K} + \tau_{1/2}^2(\mu) \mbox{ when } K \mbox{ is positive}), \\
\Ex_\mu\{B_C\} &\leq& B=4\,\overline{K} \ (B=2\,\overline{K} \mbox{ when } K \mbox{ is positive}), \\
\Ex_\mu\{M_C^2\} &\leq& M^2(\mu)/C=[\tau_1(\mu)-\SE_K(\mu)]/C \,,
\eea
where $M_C^2$ is given by \eqref{MC2}, $A_C=[\overline{K}_C^{1/2} + \tau_{1/2}(\mu)]^2$ and $B_C=4\,\overline{K}_C$ ($A_C=\overline{K}_C + \tau_{1/2}^2(\mu)$ and $B_C=2\, \overline{K}_C$ when $K$ is positive).
\end{lemma}

\citet{TeymurGRO2021} derive a bound similar to \eqref{induction-1} (with $A_C$ and $M_C^2$ replaced by $\Ex_\mu\{A_C\}$ and $\Ex_\mu\{M_C^2\}$) for Algorithm~\ref{algo:GM-alphak} with $\ma_k=1/k$ for all $k$ in the situation where
a different sample $\SX_C[k]$ of $C$ random points is used at each iteration; see also \citet{ChenBBGGMO2019}. The extension to this situation of the approach used in previous sections does not seem straightforward as the probability simplex $\SP_C$ and matrices $\Kb_C$ and ${\Kb_\mu}_C$ refer to a fixed set $\SX_C$. In Appendix~C we provide arguments explaining how our results extend to the case where $\SX_C=\SX_C[k]$ depends on $k$: basically, similar bounds continue to hold provided we consider the expectation of $\MMD_K^2(\mu,\xi_n)$ and bound the expected values of the constants involved as in Lemma~\ref{L:XC-random}. Note that changing the candidate set at every iteration implies that we need to calculate $K(\xb_i,\xb)$ for all $\xb_i\in\supp(\xi_k)$ and all $\xb\in\SX_C[k]$, and to recalculate $P_{K,\mu}(\xb)$ (Algorithms~\ref{algo:KH-alphak} and \ref{algo:KH-optimal-alpha}), or $P_{K,\mu}(\xb)$ and $K(\xb,\xb)$ (Algorithms~\ref{algo:GM-alphak} and \ref{algo:GM-alphaopt}), for all $\xb\in\SX_C[k]$ at every iteration, with a computational cost thus growing as $\SO(k^2\,C)$. 

We conclude this section by recalling a result on the MMD of the empirical measure $\xi_{n,e}$ of a random $n$-point sample from $\mu$; see \citet[Lemma~2]{MakJ2018}. The proof is given in Appendix~B.

\begin{thm}\label{Th:iid}
When $\xb_1,\ldots,\xb_n$ are independently sampled from $\mu$, then $n\,\MMD_K^2(\mu,\xi_{n,e})\rad Z=\sum_{i=1}^\infty \ml_i \chi_{1i}^2$, where the $\ml_i$ are the eigenvalues of the operator $T_{K_\mu}$ on $L_2(\SX,\mu)$ defined by $T_{K_\mu}f (\xb)=\int_\SX K_\mu(\xb,\xb')f(\xb')\,\dd\mu(\xb')$, $f\in L_2(\SX,\mu)$, $\xb\in\SX$, and the $\chi_{1i}^2$ are independent $\chi_{1}^2$ random variables.
\end{thm}

From Lemma~\ref{L:XC-random} and Theorem~\ref{Th:iid} we have in particular $\Ex_\mu\{\MMD_K^2(\mu,\xi_{n,e})\}=M^2(\mu)/n=[\tau_1(\mu)-\SE_K(\mu)]/n$ and $n^2\,\var_\mu\{\MMD_K^2(\mu,\xi_{n,e})\} \to 2\,\sum_{i=1}^\infty \ml_i^2$ as $n\ra\infty$. Although the bounds obtained in previous sections suggest that the measures obtained with the algorithms that have been considered do not perform necessarily better (asymptotically) than i.i.d.\ samples from $\mu$, the examples in the next section demonstrate the interest of using KH, GM or SBQ.

\section{Numerical study}\label{S:examples}

\subsection{Example 1: space-filling design}\label{S:SFD}

For illustration purpose we only consider the case $d=2$ and take $\mu$ uniform on $\SX=[0,1]^2$; $\SX_C$ corresponds to the first $2^{17}=131\,072$ points of a scrambled Sobol' sequence in $\SX$. $K$ is a separable kernel given by the product of uni-dimensional Mat\'ern 3/2 covariance functions, that is, $K(\xb,\xb')=\prod_{i=1}^d K_{3/2,\mt}(x_i,x_i')$ with
\bea
K_{3/2,\mt}(x,x')=(1+\sqrt{3}\mt\,|x-x'|)\, \exp(-\sqrt{3}\mt\,|x-x'|) \,.
\eea
We have $\SE_K(\mu)=\prod_{i=1}^d \SE_{K_{3/2,\mt}}(\mu_1)$ and $P_{K,\mu}(\xb)=\prod_{i=1}^d P_{K_{3/2,\mt},\mu_1}(x_i)$ with $\mu_1$ the uniform measure on $[0,1]$, and $\SE_{K_{3/2,\mt}}(\mu_1)$ and $P_{K_{3/2,\mt},\mu_1}(x)$ can be computed explicitly; see, e.g., \citet[Table~3.1]{PZ2020-SIAM}. Examples of space-filling design based on MMD-minimisation with $d=10$ and recommendations for the choice of $\mt$ are given in the same paper. We use $\mt=10$ throughout the example.

\begin{figure}[ht!]
\begin{center}
\includegraphics[width=.49\linewidth]{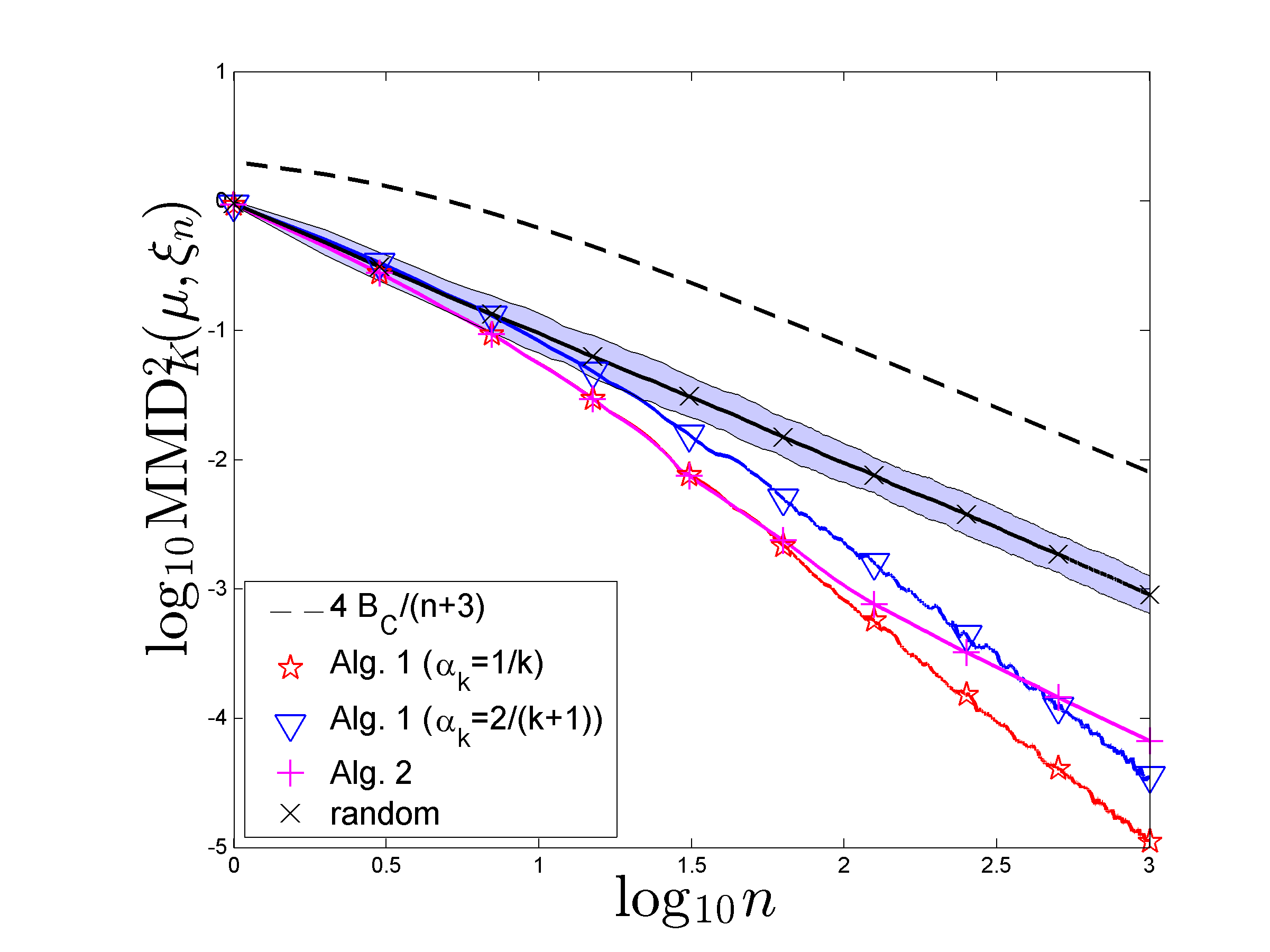} \includegraphics[width=.49\linewidth]{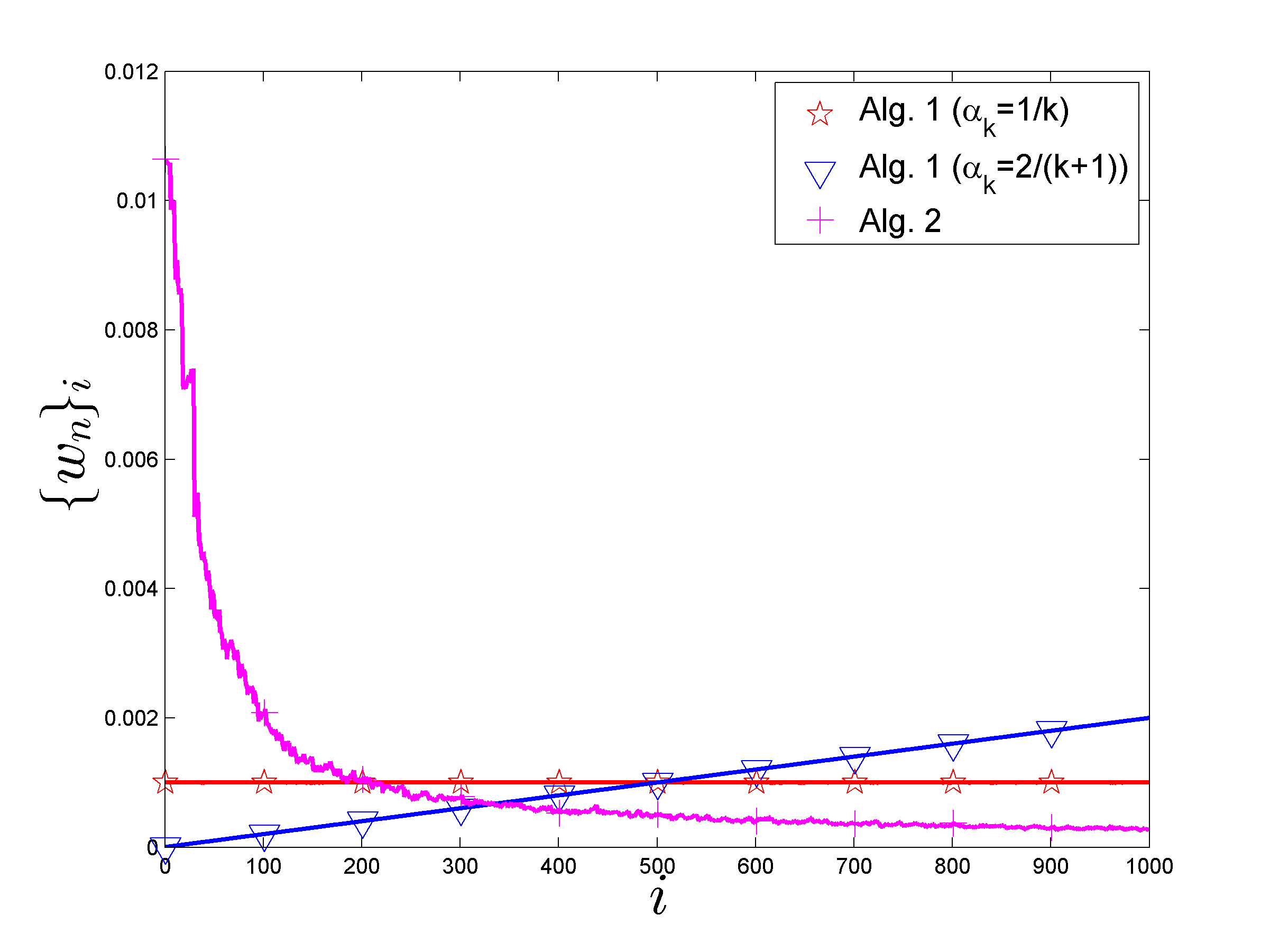}
\end{center}
\caption{\small Left: upper bound \eqref{bound:KH-predefined-SZ} and $\MMD_K(\mu,\xi_n)$ for $\xi_n$ generated with Algorithm~1 with $\ma_k=1/k$ and $\ma_k=2/(k+1)$, with Algorithm~2 and for the empirical measure $\xi_n=\xi_{n,e}$ of random $n$-point designs (mean value $\pm$ $2 \ms$ over 100 repetitions), $n=1,\ldots,1\,000$. Right: weights $\{\wb_{1\,000}\}_i$ of $\xi_{1\,000}$.}\label{F:EX1-KH}
\end{figure}

The left panel of Figure~\ref{F:EX1-KH} shows the evolution of $\MMD_K(\mu,\xi_n)$ as a function of $n$ (log-log plot) when $\xi_n$ is generated with Algorithm~1 with $\ma_k=1/k$ and $\ma_k=2/(k+1)$, or with Algorithm~2. The bound \eqref{bound:KH-predefined-SZ} is also shown ($M_C^2$ is negligible). Although Algorithm~2 uses optimal step-sizes, its performance is the worst for large $n$ and is never better than that of Algorithm~1 with $\ma_k=1/k$ (note that the rate of decrease of $\MMD_K(\mu,\xi_n)$ for Algorithm~2 closely follows $\SO(1/n)$ when $n\gtrsim 100$). Although the bound \eqref{bound:KH-predefined-SZ} of Theorem~\ref{Th:KH-predefined-SZ} is better than \eqref{bound:KH-empirical} of Theorem~\ref{Th:KH-empirical}, $\ma_k=1/k$ yields better performance than $\ma_k=2/(k+1)$ all along the sequence. This suggests that there is little interest in using more sophisticated versions of KH than Algorithm~1 with $\ma_k=1/k$.
We also computed the MMD for empirical measures associated with random designs. The average and 2$\ms$ intervals obtained for 100 repetitions are presented, showing a decrease that closely follows $\SO(1/n)$, as predicted by Theorem~\ref{Th:iid}.
The evolution of $\MMD_K(\mu,\xi_n)$ obtained for $\xi_n$ generated with Algorithm~4 with $\ma_k=1/k$ and $\ma_k=2/(k+1)$ (respectively, with Algorithm~5) is visually hardly distinguishable from that obtained with Algorithm~1 with $\ma_k=1/k$ and $\ma_k=2/(k+1)$ (respectively, with Algorithm~2) and is not presented.

The right panel of Figure~\ref{F:EX1-KH} shows the strong non-uniformity of the weights $\{\wb_{1\,000}\}_i$, $i=1,\ldots,1\,000$, associated with the measures $\xi_{1\,000}$ generated by Algorithm 1 with $\ma_k=2/(k+1)$ and by Algorithm~2 (note that most recent points are overweighed for the former and downweighed for the latter).

\begin{figure}[ht!]
\begin{center}
\includegraphics[width=.49\linewidth]{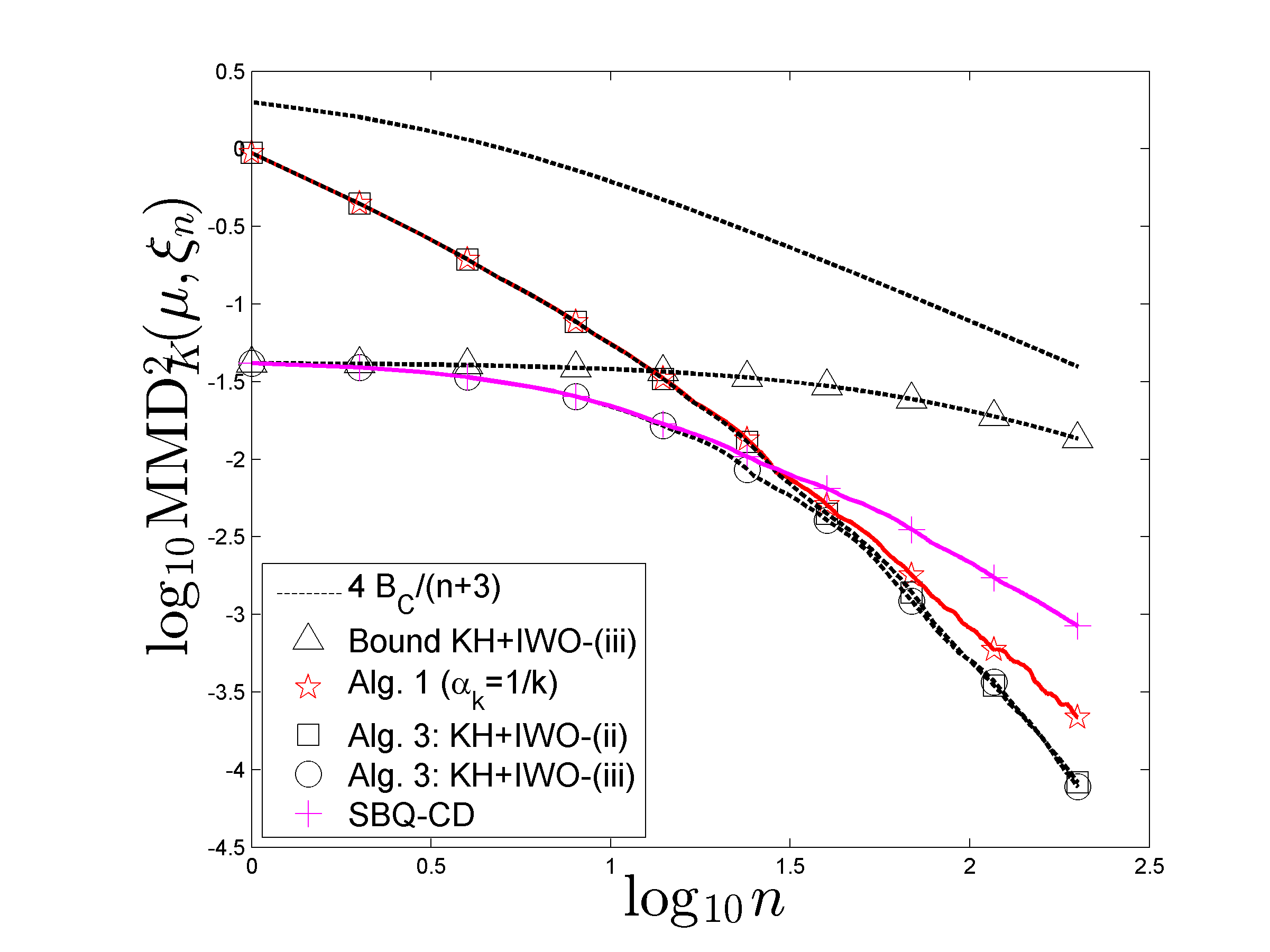} \includegraphics[width=.49\linewidth]{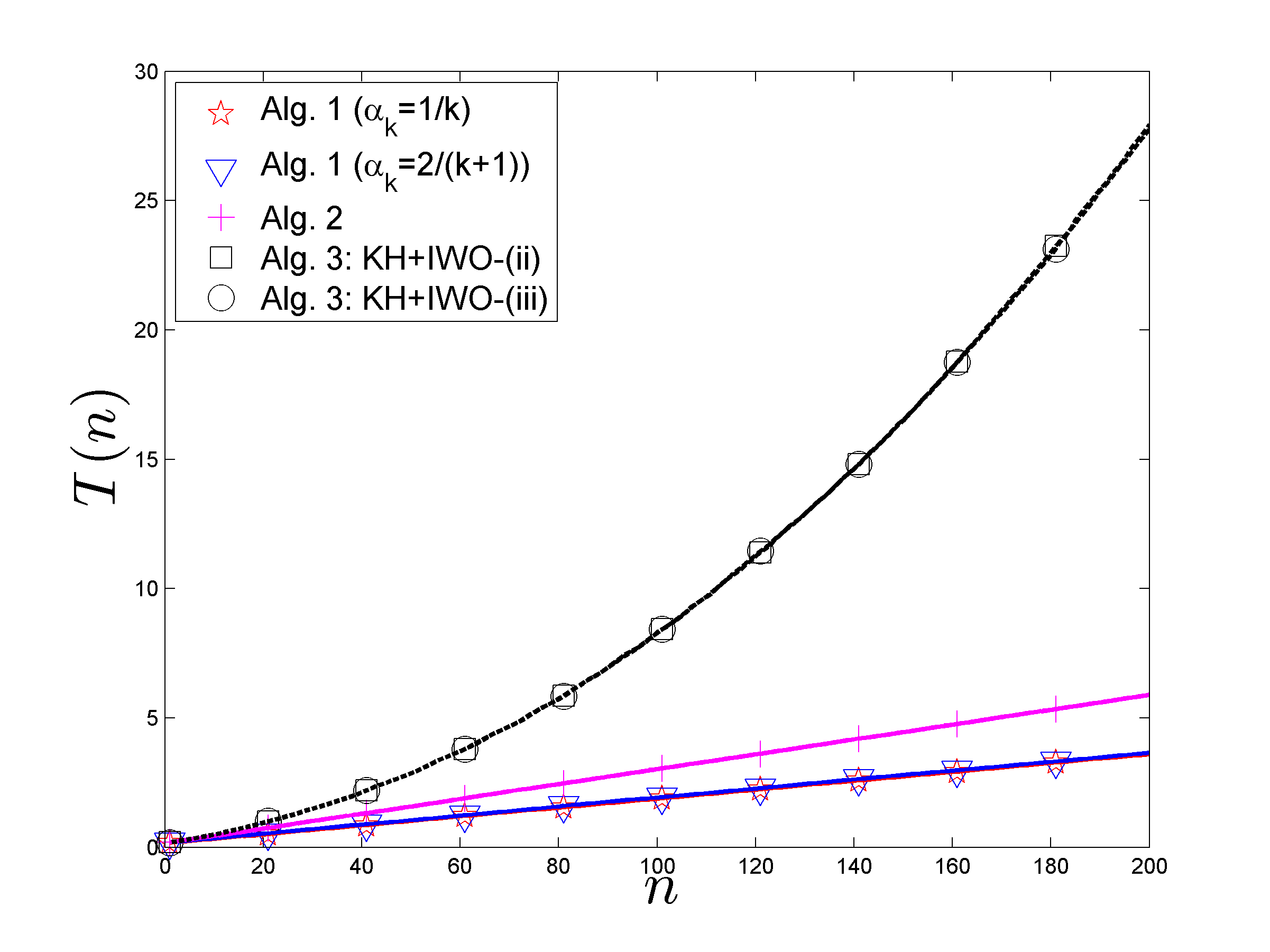}
\end{center}
\caption{\small Left: upper bounds \eqref{bound:KH-predefined-SZ} and \eqref{KH-iii}, and $\MMD_K(\mu,\xi_n)$ for $\xi_n$ generated with Algorithm~1 with $\ma_k=1/k$, with Algorithm~3-(\textit{ii}) and (\textit{iii}), and with version \eqref{SBQ3} of SBQ, $n=1,\ldots,200$. Right: computational time $T(n)$ (in s) of $\xi_n$ for Algorithm~1 with $\ma_k=1/k$ and $\ma_k=2/(k+1)$, for Algorithm~2 and for Algorithm~3-(\textit{ii}) and (\textit{iii}), $n=1,\ldots,200$.}\label{F:EX1-KH-B}
\end{figure}

We consider now the variants (\textit{ii}) and (\textit{iii}) of IWO in Algorithm~3. Unsurprisingly, $\MMD_K(\mu,\nu_n)$ is smaller for $\nu_n=\widetilde\xi_n$ of variant (\textit{iii}) than for $\nu_n=\widehat\xi_n$ of variant (\textit{ii}) since the weights are unconstrained in the former case, see the left panel of Figure~\ref{F:EX1-KH-B}. The two variants perform similarly for large $n$, however.
The bound \eqref{KH-iii} for Algorithm~3-(\textit{iii}) is accurate for small $n$ but very pessimistic for large $n$; see Remark~\ref{R:SBQ1}.
Algorithm~3-(\textit{ii}) performs as Algorithm~1 with $\ma_k=1/k$ for small $n$ ($n\lesssim 30$) but performs significantly better for larger $n$. The performances are quasi identical when using OLWO of Section~\ref{S:off-line-KH} (Frank-Wolfe Bayesian quadrature, not shown). When we stop Algorithm~1 (with $\ma_k=1/k$) at $n=200$, all weights $\{\widehat\wb_n\}_i$ and $\{\widetilde\wb_n\}_i$, $i=1,\ldots,n$, are positive. The weights are positive too for Algorithm~3-(\textit{ii}) and (\textit{iii}), so that variant (\textit{ii}) coincides with Algorithm~3-(\textit{i}), the fully-corrective Frank-Wolfe algorithm (and also with the minimum-norm point algorithm, see Remark~\ref{R:MNP}).
The evolution of $\MMD_K(\mu,\xi_n)$ for $\xi_n$ obtained with the version \eqref{SBQ2} of SBQ (with weights whose sum equals one) is indistinguishable from that obtained with Algorithm~3-(\textit{ii}). The behaviour of the version \eqref{SBQ1} of SBQ (with unconstrained weights) is similar to that of Algorithm~3-(\textit{iii}) and is rather typical, see for instance \citet{BriolOGO2015, HuszarD2012}; see also Figure~\ref{F:Ex2-MMD-KH-GM}-left. The Coordinate-Descent variant \eqref{SBQ3} of SBQ, denoted SBQ-CD, is clearly not competitive compared to the other algorithms considered.

Computational times\footnote{All calculations are made with Matlab, on a PC with a clock speed of 1.5 GHz and 16 GB RAM.} are shown on the right panel of Figure~\ref{F:EX1-KH-B} for Algorithm~1 with $\ma_k=1/k$ and $\ma_k=2/(k+1)$, for Algorithm~2, and for Algorithm~3-(\textit{ii}) and (\textit{iii})\footnote{All computational times start with a positive value at $n=0$ since we account for the calculation of $P_{K,\mu}(\xb)$ for all $\xb\in\SX_C$. The faster running time than the theoretical complexity estimates given in the paper can be explained by the internal vectorisation of operations in Matlab.}. The choice of the sequence $(\ma_k)_k$ has no influence on the computational time of Algorithm~1; Algorithm~2 is slightly more demanding, but its computational time still grows linearly with $n$; the better performance of IWO shown on the left panel comes with a significant increase of computational cost (which is similar for OLWO).

\begin{figure}[ht!]
\begin{center}
\includegraphics[width=.49\linewidth]{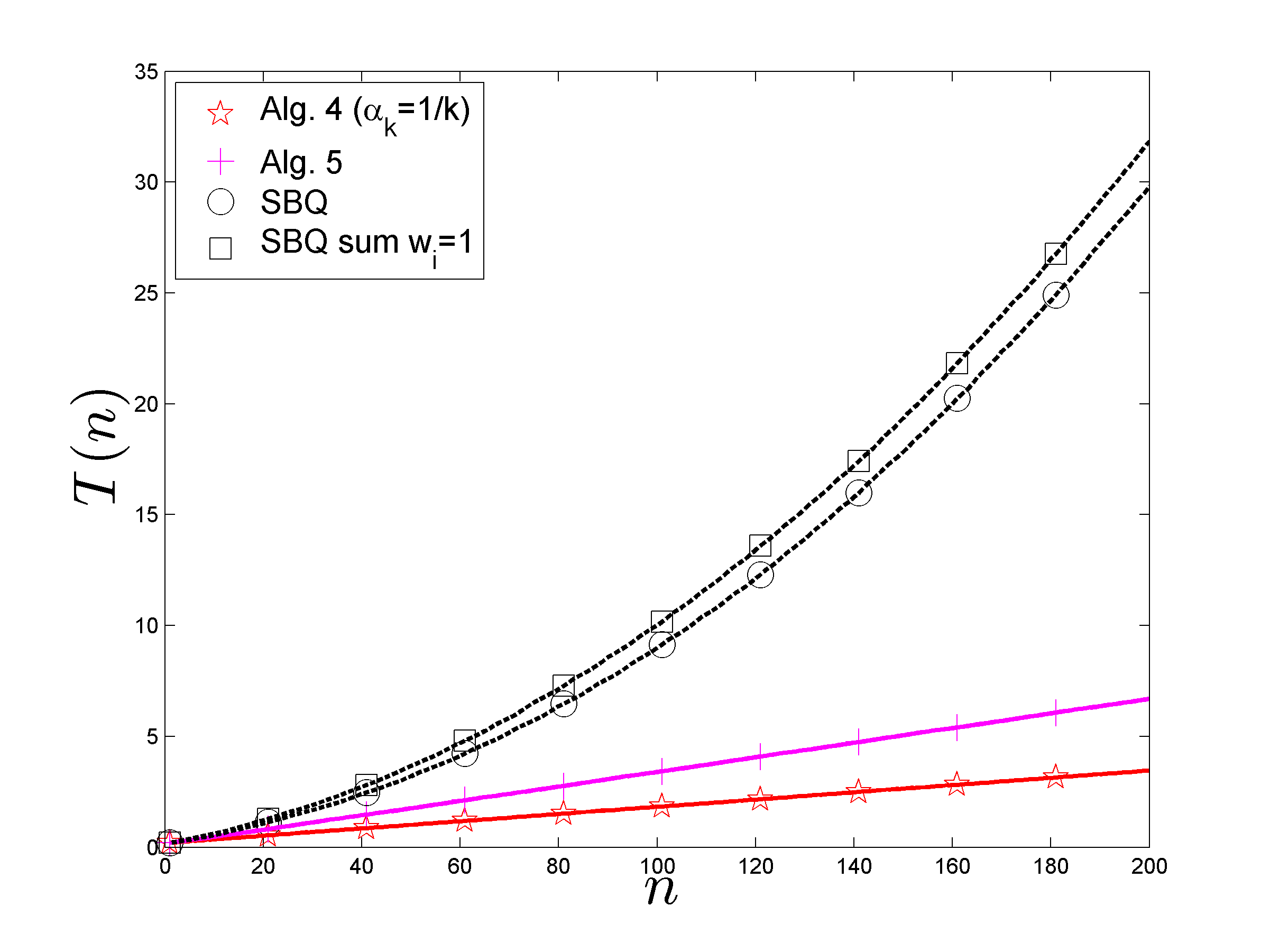} \includegraphics[width=.49\linewidth]{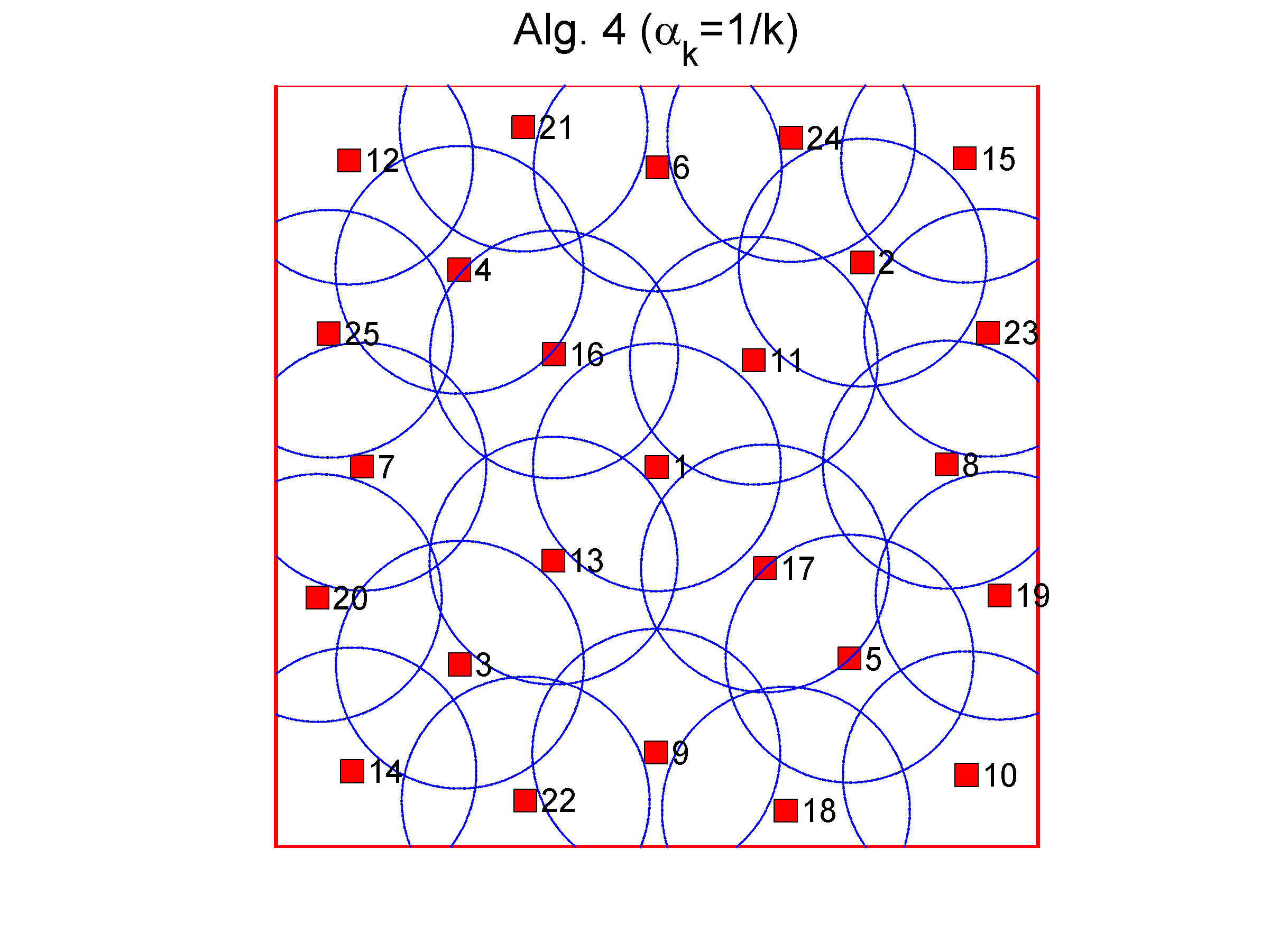}
\end{center}
\caption{\small Left: computational time $T(n)$ (in s) of $\xi_n$ for Algorithm~4 with $\ma_k=1/k$, Algorithm~5 and for the two versions \eqref{SBQ1} and \eqref{SBQ2} of SBQ, $n=1,\ldots,200$. Right: support $\Xb_n$ (ordered) of $\xi_{25}$ generated with Algorithm~4 with $\ma_k=1/k$; the radius of the circles equals the covering radius of $\Xb_n$ (the smallest value that permits to cover $\SX$).}
\label{F:EX1-GM-SBQ}
\end{figure}

Computational times for Algorithm~4 with $\ma_k=1/k$, Algorithm~5 and the two versions \eqref{SBQ1} and \eqref{SBQ2} of SBQ are shown on the left panel of Figure~\ref{F:EX1-GM-SBQ}: Algorithm~4 is as fast as Algorithm~1; Algorithm~5 is slightly slower than Algorithm~3. The two versions of SBQ are slightly slower than Algorithm~3-(\textit{ii}) and (\textit{iii}) for similar performance.

MMD minimisation with $\mu$ uniform on $\SX$ is an efficient method to construct nested space-filling designs; this is one of the main motivations in \citep{PZ2020-SIAM}. The right panel of Figure~\ref{F:EX1-GM-SBQ} shows the 25-point design corresponding to the support of the measure $\xi_n$ generated with Algorithm~4 with $\ma_k=1/k$, with a covering radius\footnote{The covering radius of a design $\Xb_n$ is defined by $\CR(\Xb_n)=\max_{\xb\in\SX}\min_{\xb_i\in\Xb_n}\|\xb-\xb_i\|$; a small value of $\CR(\Xb_n)$ indicates that for each point in $\SX$ there is a design point at proximity, hence the frequent use of $\CR(\Xb_n)$ as a space-filling characteristic to be minimised.}
$\CR(\Xb_{25})\simeq 0.1625$. Algorithm~1 with $\ma_k=1/k$ yields a very similar design, but with a different ordering of points and a slightly larger covering radius $\CR(\Xb_{25})\simeq 0.1685$. When using Algorithm~3-(\textit{ii}) (respectively, Algorithm~3-(\textit{iii})), the support of $\xi_{25}$ has a covering radius
$\CR(\Xb_{25})\simeq 0.1677$ (respectively, $\CR(\Xb_{25})\simeq 0.2024$). This illustrates the fact that a smaller MMD is not necessarily synonym to better space-filling properties: the optimal weighting of a given design improves its MMD, but space-filling performance, measured for instance by the covering radius, is unweighed. In fact, when allocating uniform weights to the support of $\xi_n$ generated with Algorithm~3-(\textit{iii}), the MMD obtained is similar to that shown on the left panel of Figure~\ref{F:EX1-KH-B} for Algorithm~1 with $\ma_k=1/k$ (red $\bigstar$), thus much worse than for the original $\xi_n$ (black $\circ$).

\subsection{Example 2: Gaussian mixture}\label{S:Ex2}

\begin{figure}[ht!]
\begin{center}
\includegraphics[width=.49\linewidth]{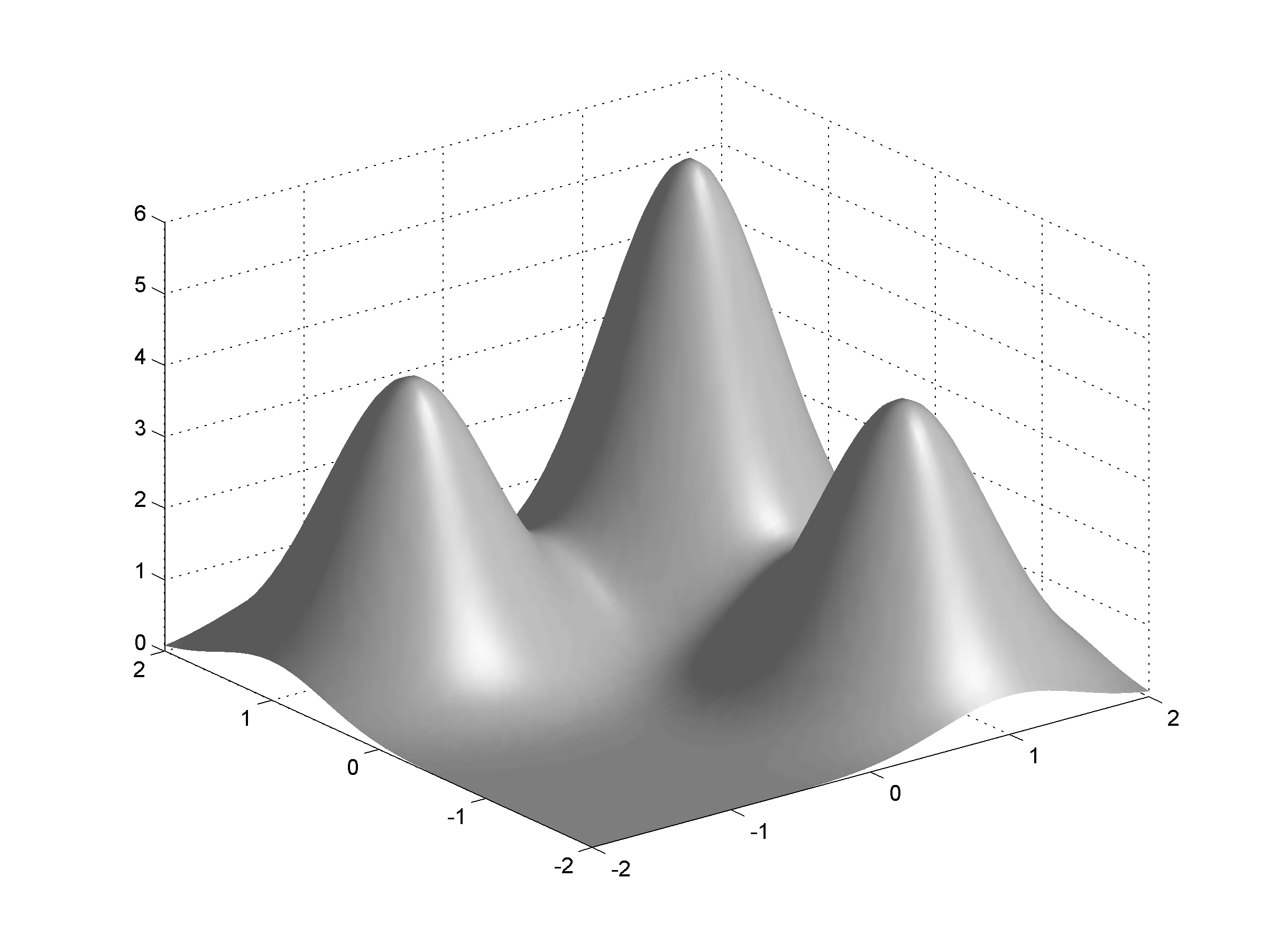} \includegraphics[width=.49\linewidth]{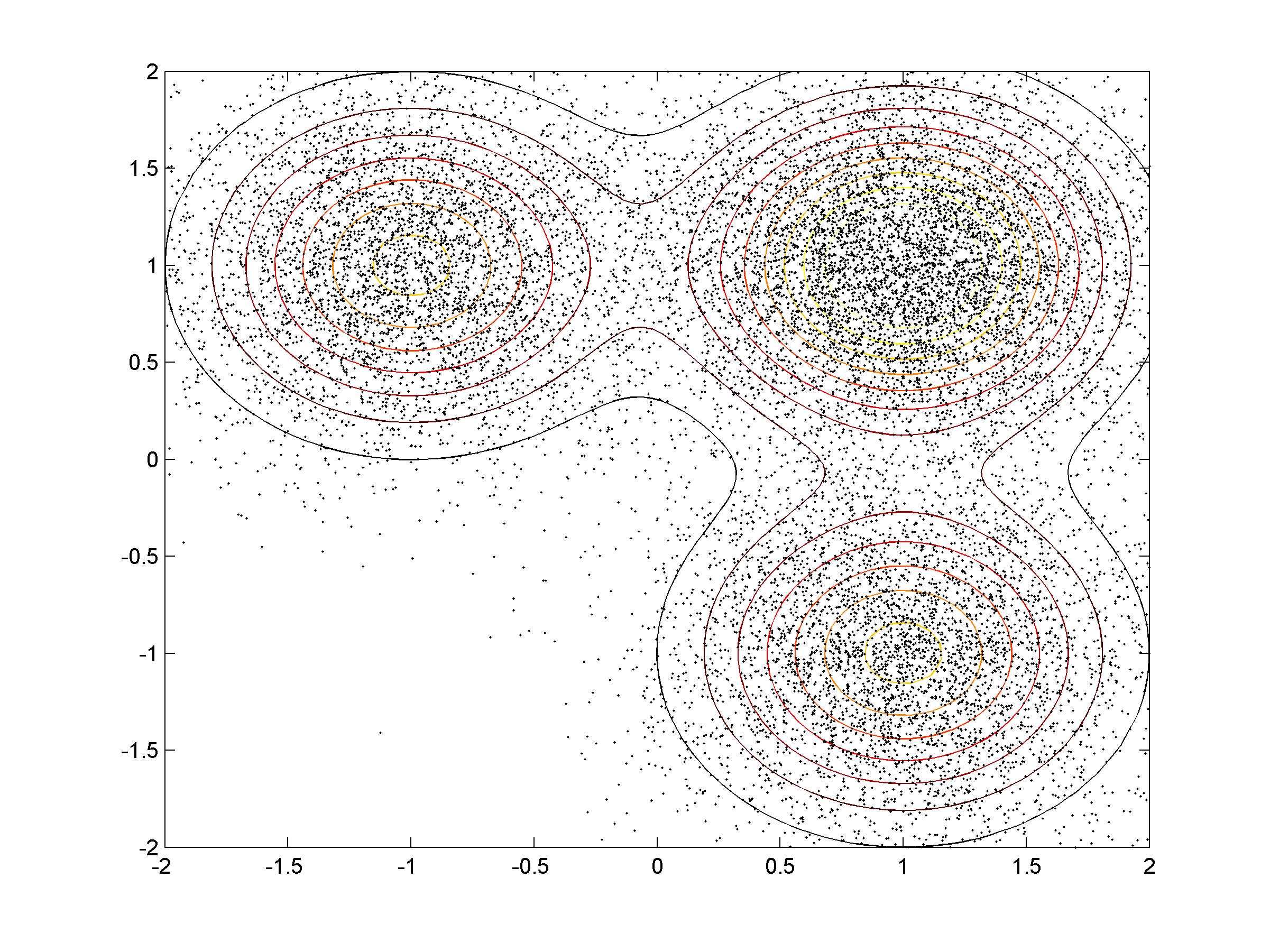}
\end{center}
\caption{\small Left: 3d-plot of the p.d.f.\ $\varphi_\mu$; right: contour lines of $\varphi_\mu$ and candidate set $\SX_C$ (dots).}\label{F:Ex2-density-mu}
\end{figure}

Here $\mu=\sum_{j=1}^m \beta_j\, \mu_\SN(\ab_j,\ms_j)$ with $\beta_j>0$, $\sum_{j=1}^m \beta_j=1$, where $\mu_\SN(\ab_j,\ms_j)$ corresponds to the normal distribution with mean $\ab_j$ and variance $\ms_j^2\, \Ib_d$, with $\Ib_d$ the identity matrix. Again, for illustration purpose, we take $d=2$. The difficulty increases with the number $m$ of components, the problem is also more difficult when the weights and/or variances of the components differ. We take $m=3$, $\ab_1=(-1,1)\TT$, $\ab_2=(1,-1)\TT$, $\ab_3=(1,1)\TT$, $\ms_j=1/2$ for all $j$, and $\beta_1=\beta_2=2/7$, $\beta_3=3/7$ (this is a slight variation of the example in Figure~1 of \citep{TeymurGRO2021} where the three components have equal weights).
Figure~\ref{F:Ex2-density-mu} presents a 3-d plot of the probability density function $\varphi_\mu$ (left) and its contour lines (right) together with the candidate set $\SX_C$ formed by $2^{14}=16\,384$ independent samples, among which we shall select a subset of $n$ representative points.

We use the Gaussian (or Radial Basis Function) kernel
\be
K_\mt(\xb,\xb')=\exp-(\mt\,\|\xb-\xb'\|^2) \,, \label{KG}
\ee
for which direct calculation gives\footnote{The analytic expression of $P_{K,\mu}(\xb)$ is also available for $K$ the product of uni-dimensional Mat\'ern 3/2 kernels as in Section~\ref{S:SFD}, though the expression is more complicated than \eqref{P-Gaussian} and involves the error function $\erf(t)= (2/\sqrt{\pi}) \int_0^t \exp(-x^2) \dd x$; the experimental results obtained are similar to those presented here for the Gaussian kernel.}
\be
\SE_{K_\mt}(\mu) &=& \sum_{j,\ell=1}^d \frac{\beta_j\beta_\ell}{(1+2\mt\ms_j^2+2\mt\ms_\ell^2)^{d/2}} \,
\exp\left( - \frac{\mt\, \|\ab_j-\ab_\ell\|^2}{1+2\mt\ms_j^2+2\mt\ms_\ell^2} \right) \nonumber\\
P_{K,\mu}(\xb) &=& \sum_{j=1}^d \frac{\beta_j}{(1+2\mt\ms_j^2)^{d/2}} \,
\exp\left( - \frac{\mt\, \|\xb-\ab_j\|^2}{1+2\mt\ms_j^2} \right), \ \ \xb\in\mathds{R}^d \,. \label{P-Gaussian}
\ee
It is important to choose a suitable order of magnitude for $\mt$, even if a precise tuning is not essential. This issue is frequently mentioned in the literature, see for example \citet{HuszarD2012}, but is often overlooked. As for the construction of space-filling designs where the target measure $\mu$ is uniform on $\SX$, see \citet{PZ2020-SIAM}, we recommend to let $\mt$ depend on the number of points to be generated. If the target size is $n_{\max}$ points, each point $\xb_i$ will ``represent" a fraction $1/n_{\max}$ of the $C$ candidate points, and having a correlation $K_\mt(\xb_i,\xb)>1/2$ with $C/n_{\max}$ points seems reasonable.
We thus choose $\mt$ such that $K_\mt(\xb_i,\xb_j)<1/2$ for $(100/n_{\max})\%$ of the pairs $(\xb^{(j)},\xb^{(k)})$ in a random sample of 1\,000 points of $\SX_C$ (that is, with obvious notation, $\mt=-\log(0.5)/Q_{1/n_{\max}}(\|\xb^{(j)}-\xb^{(k)}\|^2)$ for the example considered).

\begin{figure}[ht!]
\begin{center}
\includegraphics[width=.49\linewidth]{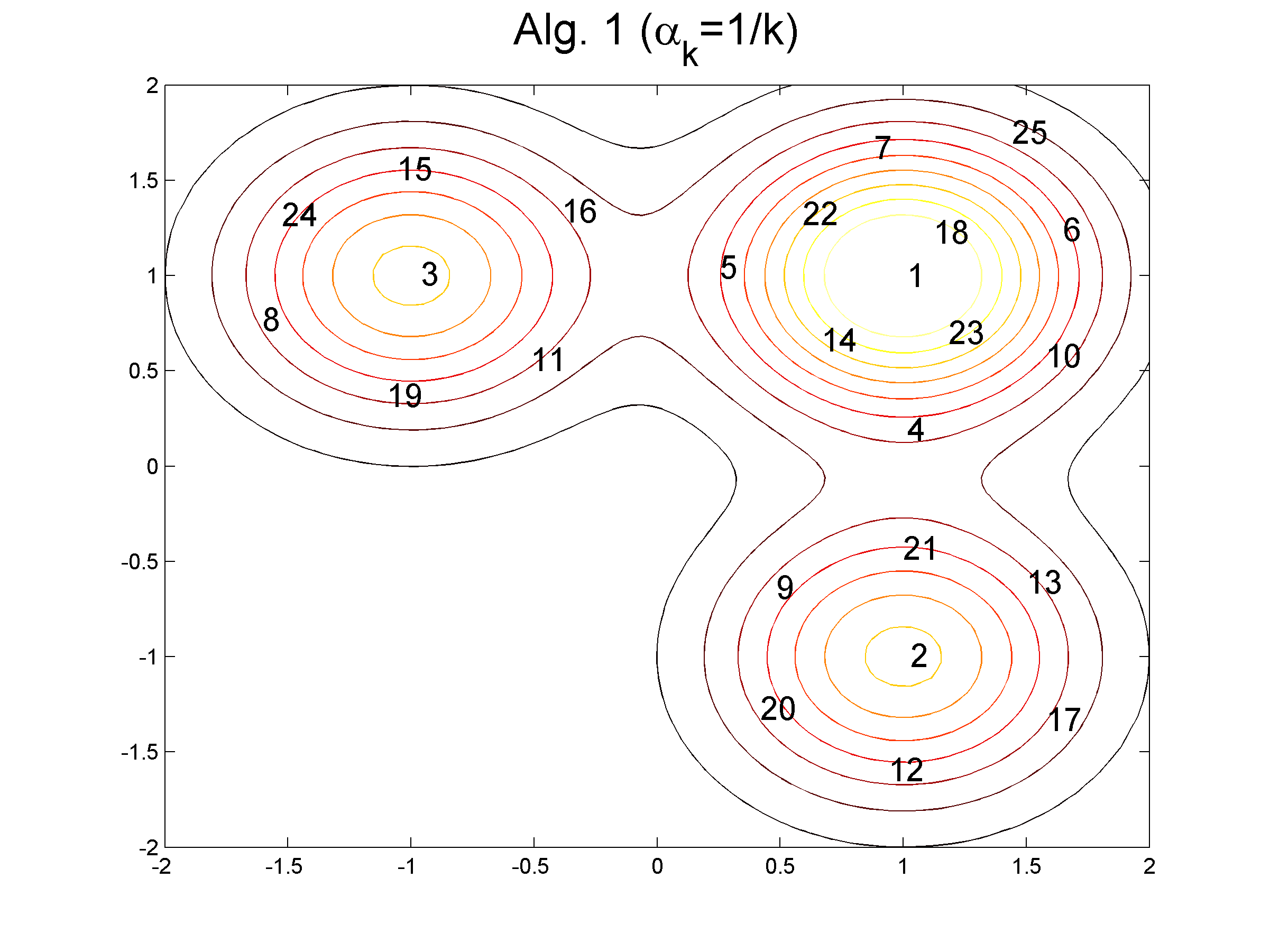} \includegraphics[width=.49\linewidth]{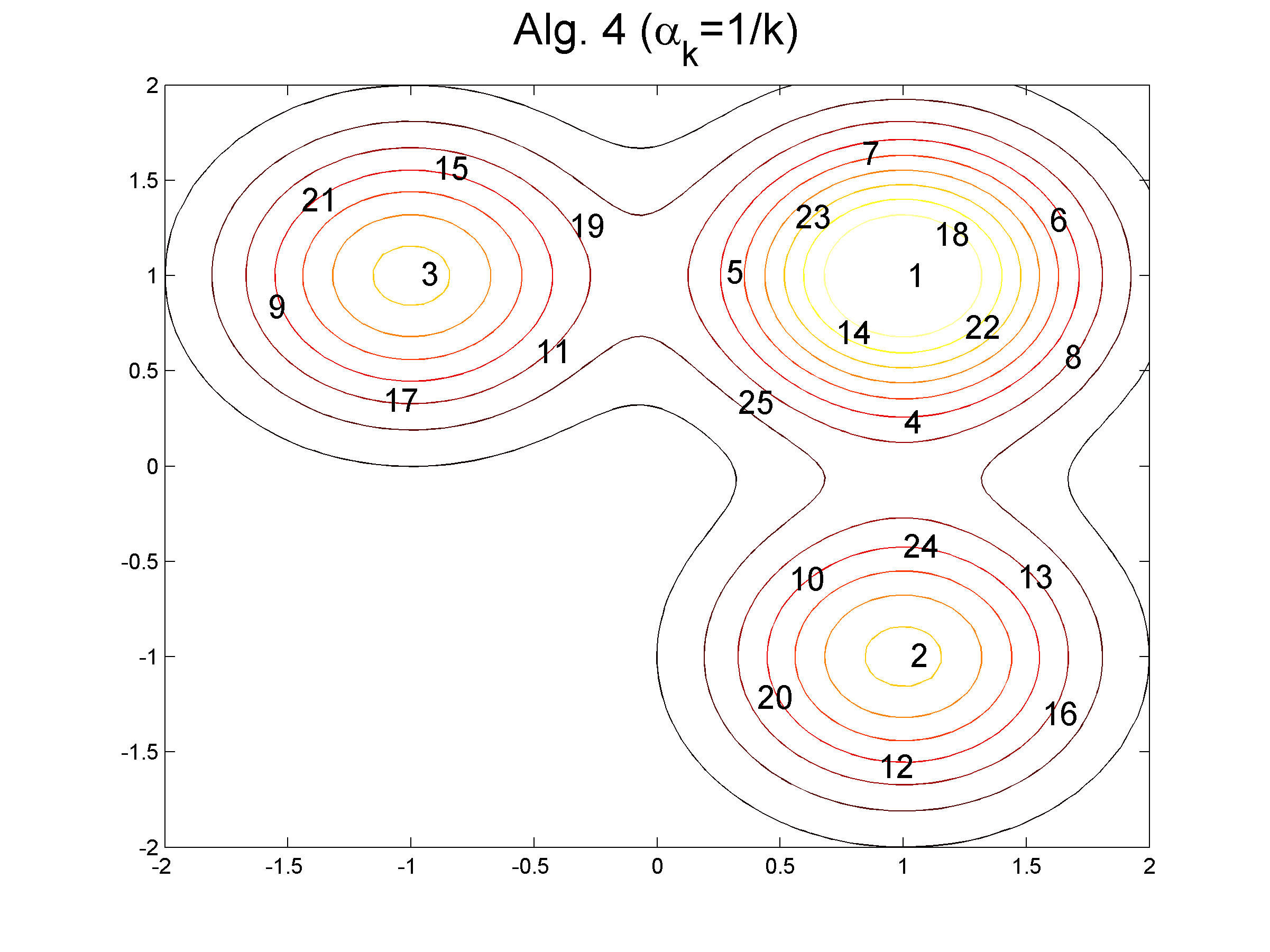}

\includegraphics[width=.49\linewidth]{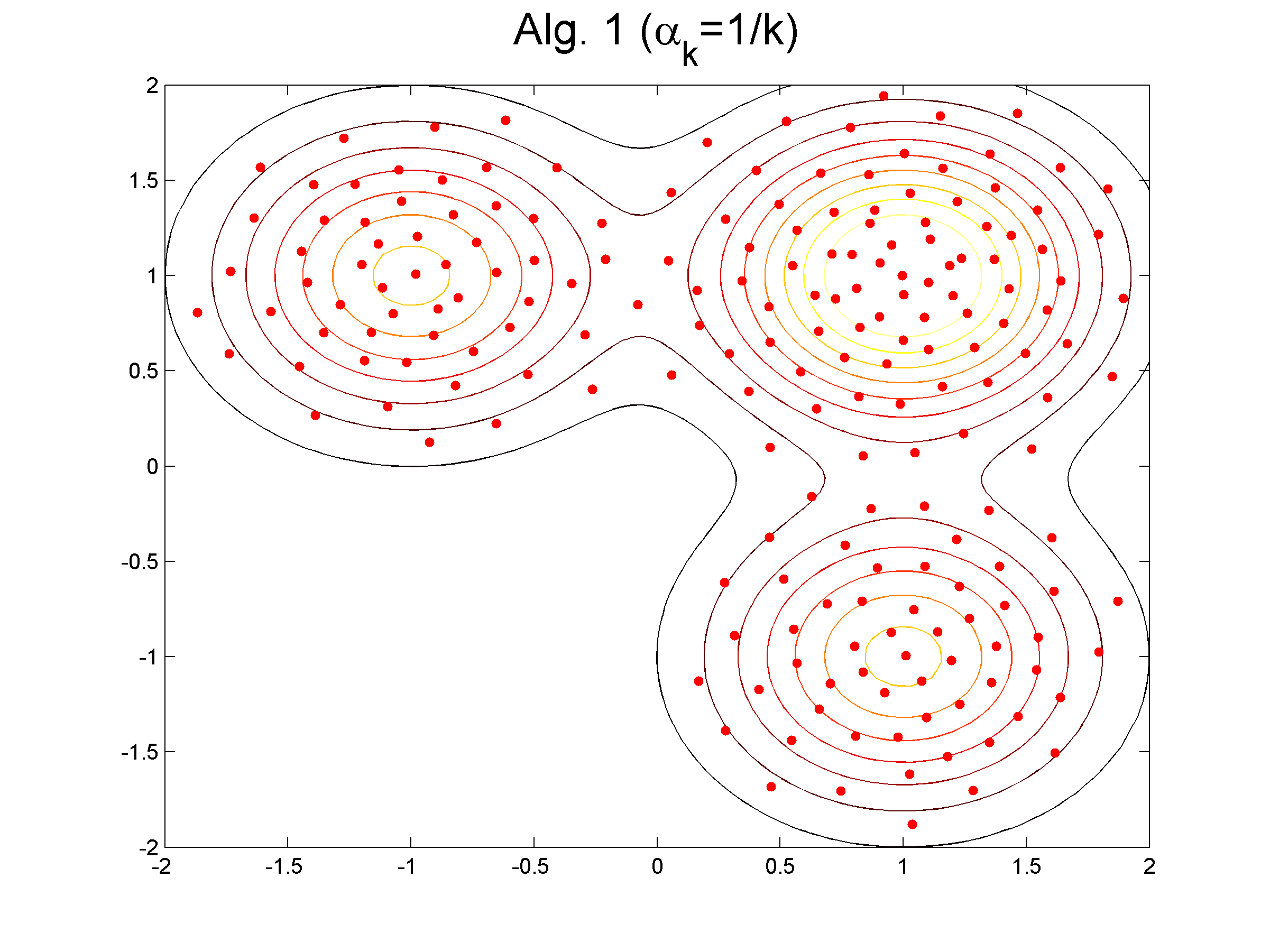} \includegraphics[width=.49\linewidth]{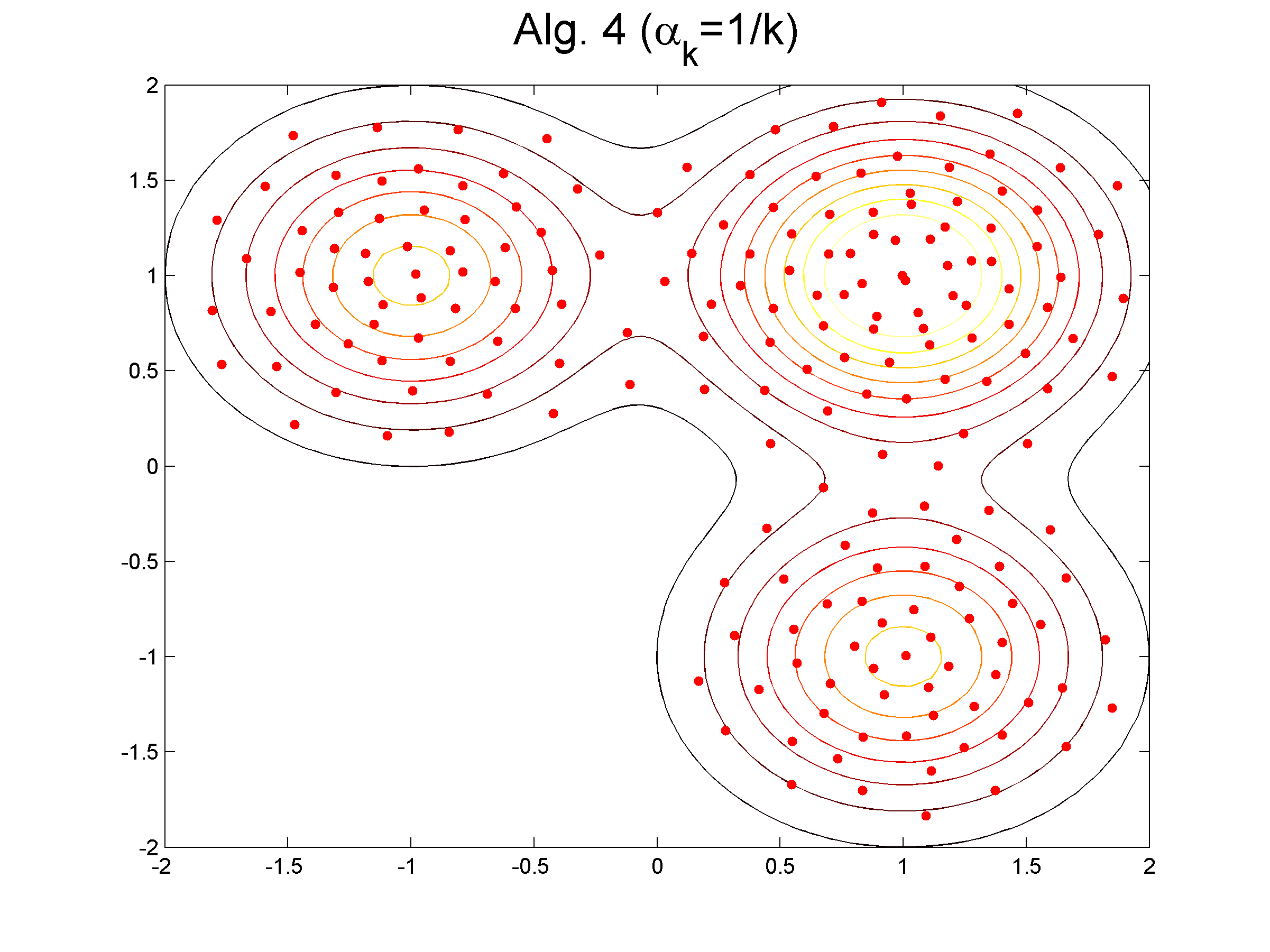}
\end{center}
\caption{\small Designs $\Xb_n$ obtained with Algorithms~1 and 4 (both with $\ma_k=1/k$); $n_{\max}=25$ ($\mt\simeq 5.7$) on the first row, $n_{\max}=200$ ($\mt\simeq 46.39$) on the second row.}\label{F:Ex2-X25_Algo1&4}
\end{figure}

Figure~\ref{F:Ex2-X25_Algo1&4} shows the first $n_{\max}$ points selected by Algorithms~1 and 4, both with $\ma_k=1/k$ for all $k$: $n_{\max}=25$ ($\mt\simeq 5.7$) on the first row, $n_{\max}=200$ ($\mt\simeq 46.4$) on the second. The points location looks roughly the same for both algorithms when $n_{\max}=25$, the ordering being however different starting at $n=12$; the designs look also similar when $n_{\max}=200$ and it is difficult to separate them.

\begin{figure}[ht!]
\begin{center}
\includegraphics[width=.49\linewidth]{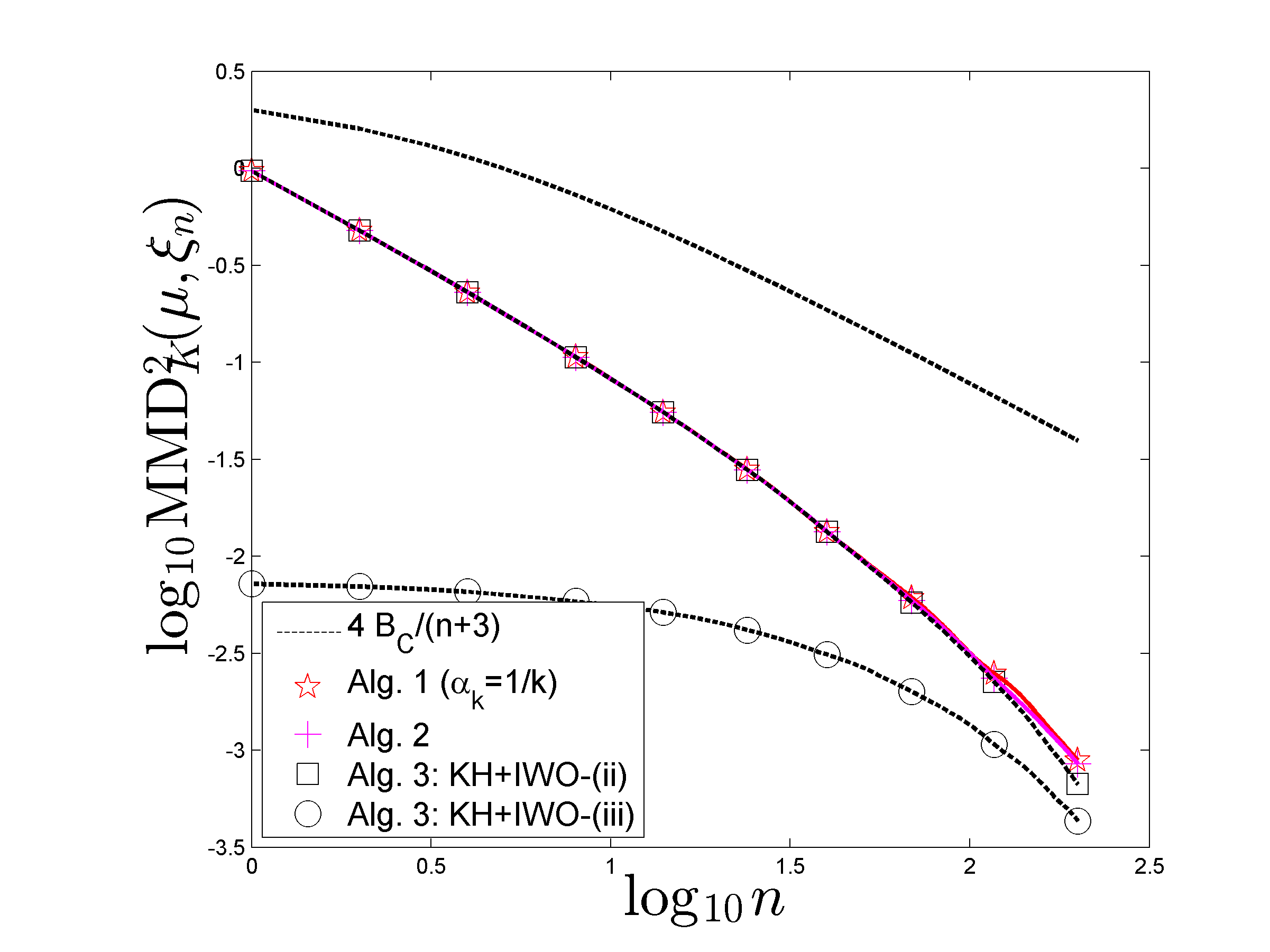} \includegraphics[width=.49\linewidth]{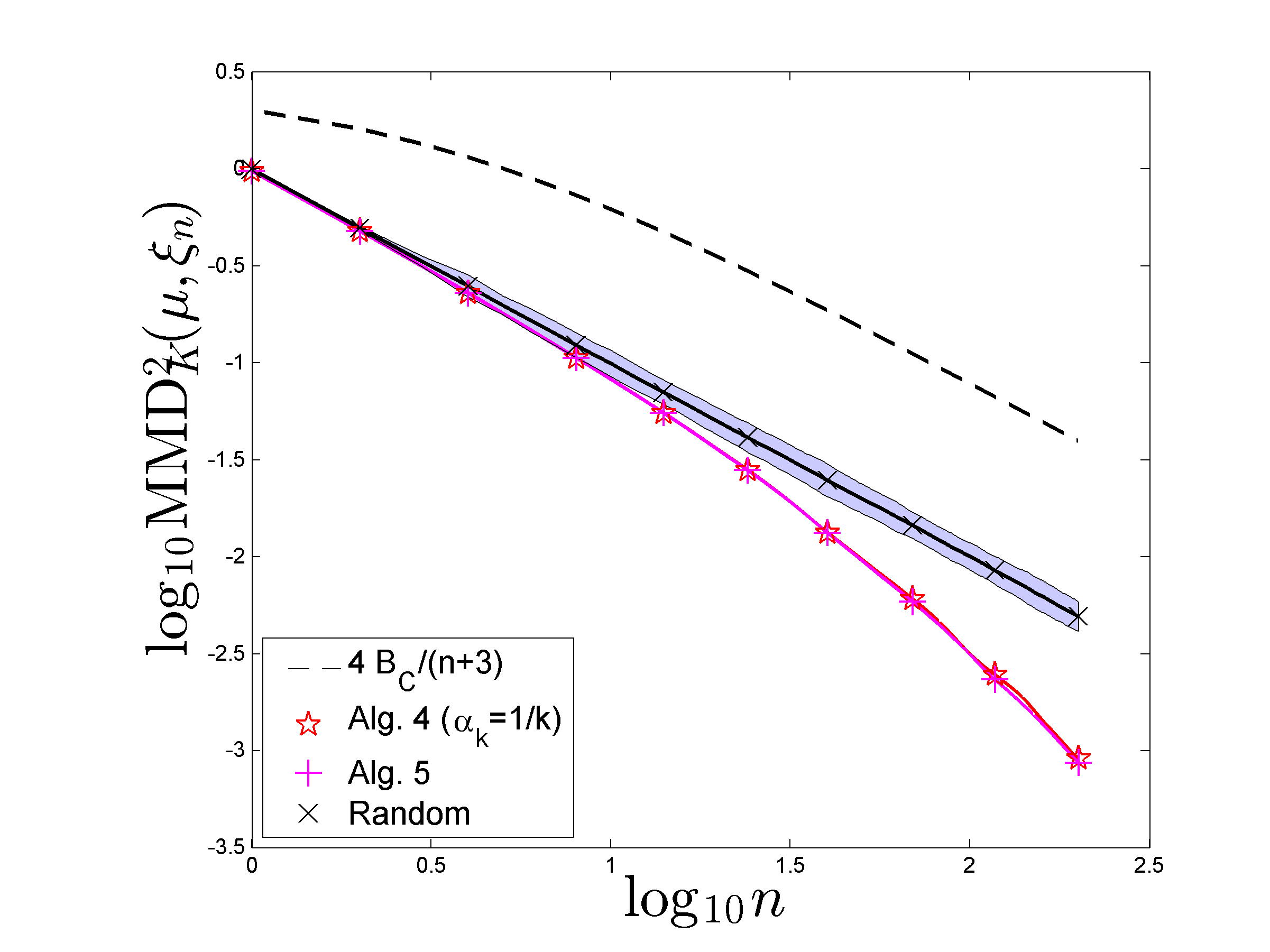}
\end{center}
\caption{\small Upper bound \eqref{bound:KH-predefined-SZ} and $\MMD_K(\mu,\xi_n)$ for $\xi_n$ generated with (left) Algorithms~1, 2, 3-(\textit{ii}), 3-(\textit{iii}); (right) Algorithms~4 and 5, and for the empirical measure of random designs (mean value $\pm$ $2\ms$ over 100 repetitions).}\label{F:Ex2-MMD-KH-GM}
\end{figure}

Figure~\ref{F:Ex2-MMD-KH-GM} shows the evolution of $\MMD_K(\mu,\xi_n)$ and its upper bound \eqref{bound:KH-predefined-SZ} for $\xi_n$ generated with Algorithms~1, 2, 3-(\textit{ii}), 3-(\textit{iii}), 4 and 5, and for empirical measures of random designs (empirical mean $\pm$ 2 standard deviations for 100 repetitions), when $n_{\max}=200$ (the designs constructed algorithmically are not shown, but they all look very similar to those on the second row of Figure~\ref{F:Ex2-X25_Algo1&4}). Algorithms~1 and 4 (both with $\ma_k=1/k$ for all $k$) and 2 and 5 perform similarly; Algorithm~3-(\textit{ii}) is only marginally superior; the MMD is significantly smaller for Algorithm~3-(\textit{iii}) which does not set constraints on $\xi_n$. The weights that $\xi_n$ allocates to its support points are positive for Algorithm~3-(\textit{ii}) and (\textit{iii}), with $\sum_{i=1}^{200} \{\wb_{200}\}_i \simeq 0.834$ for Algorithm~3-(\textit{iii}). The two versions \eqref{SBQ1} and \eqref{SBQ2} of SBQ perform similarly to Algorithm~3-(\textit{iii}) and (\textit{ii}), respectively, and their weights $\{\wb_{200}\}_i$ are positive too.

\begin{figure}[ht!]
\begin{center}
\includegraphics[width=.49\linewidth]{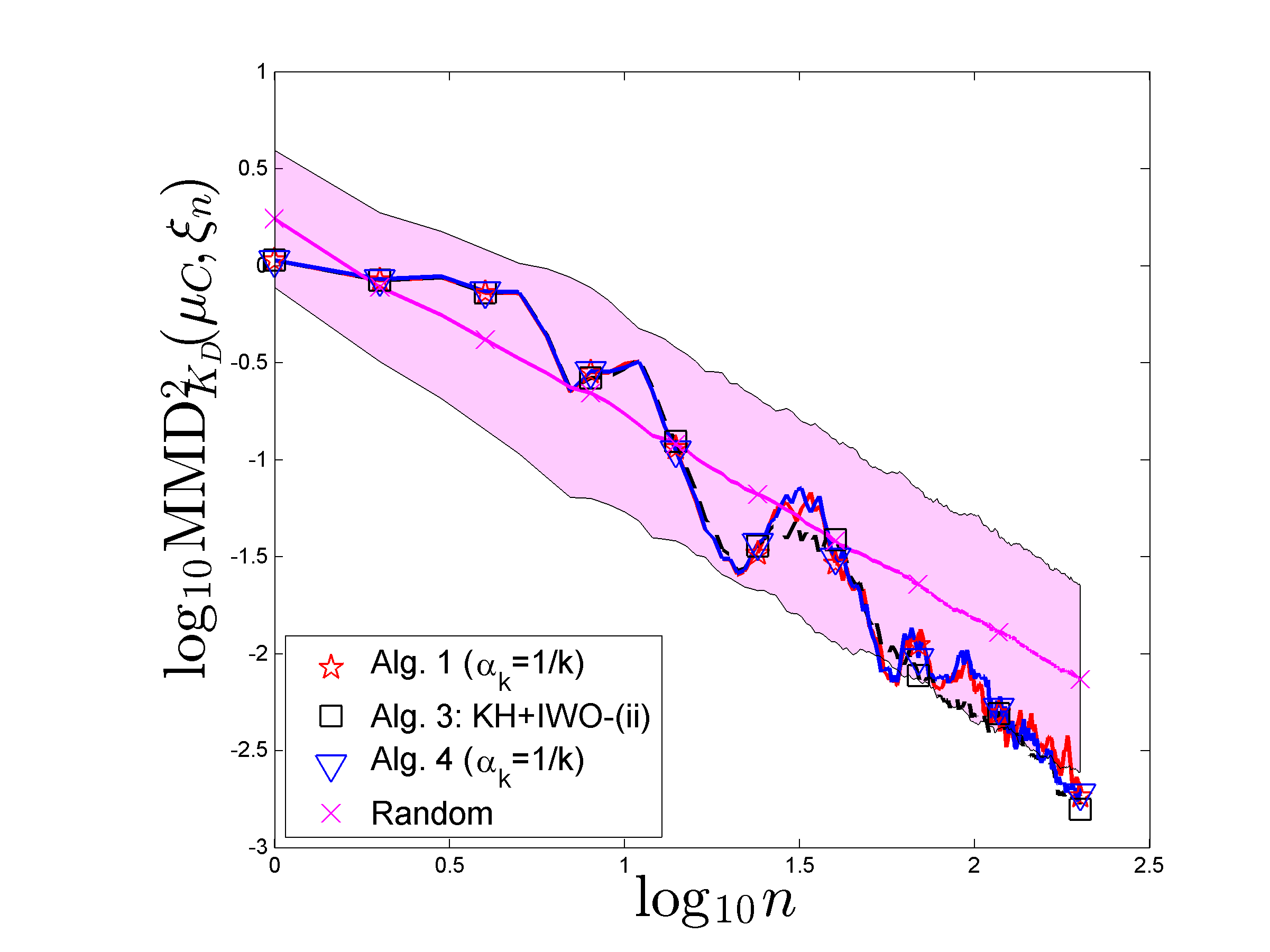}
\end{center}
\caption{\small $\MMD_{K_D}(\mu_C,\xi_n)$ for the empirical measure of random designs (mean value $\pm$ $2\ms$ over 100 repetitions) and for $\xi_n$ generated with Algorithms~1, 3-(\textit{ii}), and 4, all using the Gaussian kernel \eqref{KG}, with $K_D$ the distance kernel of \citet{SzekelyR2013}.}
\label{F:Ex3-ED}
\end{figure}

Finally, we also evaluate the approximation error by the MMD for the distance kernel $K_D(\xb,\xb')=-\|\xb-\xb'\|$ of \citet{SzekelyR2013}. Since $P_{K_D,\mu}(\cdot)$ and the energy distance $\SE_{K_D}(\mu)$ are not known explicitly, we compute $\MMD_{K_D}(\mu_C,\xi_n)$, with $\mu_C$ the empirical measure for the candidate set $\SX_C$. Figure~\ref{F:Ex3-ED} shows $\MMD_{K_D}(\mu_C,\xi_n)$ for $\xi_n$ generated with Algorithms~1, 3-(\textit{ii}) and 4, using the Gaussian kernel \eqref{KG}---$\xi_n$ generated with Algorithm~3-(\textit{iii}) cannot be tested since its weights do not sum to one\footnote{$K_D$ is Conditionally Integrally Strictly Positive Definite and defines a metric between probability measures, but $\SE_{K_D}(\mu_C-\xi)$ can be negative for $\xi\not\in\SM_{[1]}(\SX)$.}. The three algorithms appear to perform similarly and tend to provide better approximations of $\mu$ than random sampling in terms of $\MMD_{K_D}(\mu_C,\cdot)$ (the empirical mean $\pm$ two standard deviations for 100 random designs is presented). Due to their much smaller computational costs, Algorithms~1 and 4 are preferable to Algorithm~3-(\textit{ii}) (and version \eqref{SBQ2} of SBQ) in this example. We also tried to use a Stein kernel, based on the inverse multiquadric kernel $K_{s,\mt}(\xb,\xb')=1/(1+\mt\,\|\xb-\xb'\|^2)^s$, $\mt>0$, $s\in(0,1)$, instead of \eqref{KG} to generate $\xi_n$, but the values obtained for $\MMD_{K_D}(\mu_C,\xi_n)$ were significantly larger than with \eqref{KG}\footnote{We used $s=1/2$ and $\mt$ given by the median heuristic of \citet{GarreauJK2017}; see also \cite{TeymurGRO2021}.}.

\section{Conclusions}\label{S:conclusions}

Bounds on the finite-sample-size approximation error of iterative methods for the minimisation of an MMD discrepancy have been derived and illustrated by numerical experiments. These experiments indicate that the bounds give
a fair picture of the decrease rate of the true MMD for some of the methods considered (Algorithms~\ref{algo:KH-optimal-alpha} and \ref{algo:GM-alphaopt}), but are pessimistic for most of them. This is particularly true for SBQ with unconstrained weights \eqref{SBQ1}, for which the link with kernel herding used for the derivation of the error bound gives a plausible explanation for its marked pessimism; see Remark~\ref{R:SBQ1}.
These numerical results also indicate that the performances of kernel herding and greedy MMD minimisation do not improve by considering other step-size sequences than $1/k$ (which generate empirical measures), and that a variant of kernel herding with optimised weights, Algorithm~\ref{algo:KH-IWO}-(\textit{iii}), yields performance similar to standard SBQ for a slightly lower computational cost.
Therefore, on the whole, Algorithm~\ref{algo:KH-IWO}-(\textit{iii}) appears to be the best option when the budget $n$ is very limited (its complexity is quadratic in $n$), and standard KH or MMD, with uniform weights, seem generally preferable to more sophisticated methods for large $n$ (their complexity is linear in $n$).

We have restricted our attention to finite candidate sets. This situation is at the same time easier and computationally more efficient in terms of practical implementation, and simpler in terms of analysis since only finite-dimensional linear algebra is used, but the extension to the Hilbert-space situation remains possible.

Finally, we have only considered the case where one adds one-point-at-a-time to the construction. Less myopic methods that select several (say $m>1$) points at each iterations could also be considered. The extension of our results to this context, and the development of computationally efficient methods that avoid the combinatorial explosion due to considering all possible $C \choose m$ subset selections, deserve further studies. One can refer to \cite{TeymurGRO2021} for an exciting contribution in this direction.

{\small
\appendix

\section*{Appendix A: alternative expressions for $\MMD_K^2(\mu,\xi)$}

The quadratic form \eqref{MMD0} of $\MMD_K^2(\mu,\xi_n)$ and the fact that $\widetilde\xi_n$ has unconstrained optimal weights implies that, for any $\xi_n \in \SM(\Xb_n)$, \quad
\bea
\MMD_K^2(\mu,\xi_n) = (\wb_n-\widetilde\wb_n)\TT\Kb_n(\wb_n-\widetilde\wb_n) + \MMD_K^2(\mu,\widetilde\xi_n) \,. 
\eea
Therefore, any measure $\xi$ in $\SM(\SX_C)$ with associated weights $\mob$ satisfies
\be\label{MMD2-tg}
\MMD_K^2(\mu,\xi) = \tg_C(\mob)+ \MMD_K^2(\mu,\widetilde\xi^C) \,,
\ee
where, for any $\mob\in\mathds{R}^C$, we denote
\be \label{gtilde}
\tg_C(\mob)= \|\mob-\widetilde\mob^C\|_{\Kb_C}^2=(\mob-\widetilde\mob^C)\TT\Kb_C(\mob-\widetilde\mob^C) \quad (=\MMD_K^2(\xi,\widetilde\xi^C)) \,.
\ee
When $\xi_n\in\SM_{[1]}(\Xb_n)$ (i.e., $\wb_n\TT\1b_n=1$), as $(\wb_n-\widehat\wb_n)\TT\1b_n=0$ and $\Kb_n(\widehat\wb_n-\widetilde\wb_n)\propto\1b_n$, see \eqref{hatwn} and \eqref{tildewn},
\bea
\MMD_K^2(\mu,\xi_n) &=& (\wb_n-\widehat\wb_n+\widehat\wb_n-\widetilde\wb_n)\TT\Kb_n(\wb_n-\widehat\wb_n+\widehat\wb_n-\widetilde\wb_n) + \MMD_K^2(\mu,\widetilde\xi_n) \,, \\ 
&=& (\wb_n-\widehat\wb_n)\TT\Kb_n(\wb_n-\widehat\wb_n) + \MMD_K^2(\mu,\widehat\xi_n) \,. 
\eea
Therefore, any measure $\xi$ in $\SM_{[1]}(\SX_C)$ with associated weights $\mob$ satisfies
\be\label{MMD2-1}
\MMD_K^2(\mu,\xi) = \hg_C(\mob)+ \MMD_K^2(\mu,\widehat\xi^C) \,,
\ee
where
\be\label{MMD2-hg}
\hg_C(\mob)= \|\mob-\widehat\mob^C\|_{\Kb_C}^2=(\mob-\widehat\mob^C)\TT\Kb_C(\mob-\widehat\mob^C)  \quad (=\MMD_K^2(\xi,\widehat\xi^C)) \,.
\ee

For any measure $\xi\in\SM(\SX_C)$ with associated weights $\mob\in\SP_C$, we define
\be\label{Delta_C}
\Delta_C(\xi)=\MMD_K^2(\mu,\xi)-M_C^2 \,,
\ee
so that \eqref{MMD2-tg} implies $\Delta_C(\xi)= \tg_C(\mob)-\tg_C(\mob^C_*)$ and, when $\xi\in\SM_{[1]}(\SX_C)$, \eqref{MMD2-1} implies $\Delta_C(\xi)=\hg_C(\mob)-\hg_C(\mob^C_*)$.

\section*{Appendix B: proofs}

Our derivations of bounds on $\Delta_C(\xi_k)$ given by \eqref{Delta_C} (i.e., on $\MMD_K^2(\mu,\xi_k)$) rely on the convexity of $\hg_C(\cdot)$ and $\tg_C(\cdot)$ and on the following lemma.

\begin{lemma}\label{L:induction} Let $(t_k)_k$ and $(\ma_k)_k$ be two real positive sequences and $A$ be a strictly positive real. \\
(\textit{i}) If $t_k$ satisfies
\be\label{seq1}
t_1 \leq A \ \mbox{ and } \ t_{k+1} \leq (1-\ma_{k+1})\,t_k+ A\,\ma_{k+1}^2\,, \ k \geq 1\,,
\ee
with $\ma_k=1/k$ for all $k$, then $t_k\leq A\,(2+\log k)/(k+1)$ for all $k\geq 1$. \\
(\textit{ii}) If $t_k$ satisfies \eqref{seq1} with $\ma_k=2/(k+1)$ for all $k$, then $t_k\leq 4\,A/(k+3)$ for all $k\geq 1$. \\
(\textit{iii}) If $t_k$ satisfies
\be\label{seq2}
t_1 \leq A \ \mbox{ and } \ t_{k+1} \leq (1-2\,\ma_{k+1})\,t_k+ A\,\ma_{k+1}^2\,, \ k \geq 1\,,
\ee
with $\ma_k=1/k$ for all $k$, then $t_k\leq A/k$ for all $k\geq 1$.\\
(\textit{iv}) If $t_k$ satisfies
\be\label{seq3}
t_1 \leq A \ \mbox{ and } \ t_{k+1} \leq t_k - \frac{t_k^2}{A}\,, \ k \geq 1 \,,
\ee
then, $t_k\leq A/(k+p_2)$ for all $k \geq 2$, with $p_2=A/t_2-2\geq 2$; moreover, when $t_1\leq A/2$,
$t_k\leq A/(k+p_1)$ for all $k \geq 1$, with $p_1=A/t_1-1\geq 1$.
\end{lemma}

\noindent{\em Proof.}

(\textit{i}) Suppose that $t_k$ satisfies \eqref{seq1} with $\ma_k=1/k$ for all $k$. We show that $t_k\leq A \, (2+\log k)/(k+1)$ by induction on $k$. The inequality is satisfied for $k=1$, assume that it is satisfied for $k\geq 1$. We get $A\, [ 2+\log(k+1)]/(k+2) - t_{k+1} \geq A\, a(k)/[(k+2)(k+1)^2]$, with $a(k)=(k+1)^2\,\log(1+1/k)+\log k -k \geq 0$, implying that the inequality is satisfied for all $k$.

(\textit{ii}) Suppose now that $t_k$ satisfies \eqref{seq1} with $\ma_{k+1}=b/(k+1+q)$ for all $k$, for some $0<b<q+2$. We prove that $t_k\leq A\, a/(k+p)$ for some $a,p>0$ by induction on $k$. Not all values of $a,b,p,q$ are legitimate, and a natural objective is to have $a$ and $p$ respectively as small and large as possible. We show that the best choice is that indicated in Lemma~\ref{L:induction}.
For $k=1$, since $t_1\leq A$, to ensure that $t_1\leq A\, a/(p+1)$ we need to have $p\leq a-1$.
Assume that $t_k\leq A\, a/(k+p)$, and denote $\delta_k= A\,a/(k+1+p) - t_{k+1}$. It satisfies
\bea
\delta_k &\geq& A\,\frac{a}{k+1+p} - \left[(1-\ma_{k+1})\,t_k + B_C\, \ma_{k+1}^2\right] \geq A\, \frac{a(k)}{(k+p)(k+1+p)(k+1+q)^2} \,,
\eea
where $a(k)$ is a second-degree polynomial in $k$, with leading term $(ab-a-b^2)\,k^2$. We thus need to choose a pair $(a,b)$ of positive numbers such that $ab-a-b^2\geq 0$; the pair with the smallest value of $a$ is $(4,2)$. For this choice of $a,b$, we get $a(k)=4\,[k+1+p-(p-q)^2]$, which increases with $k$. We only need to guarantee that $a(1) \geq 0$, which corresponds to
$q+1/2-(1/2)\,\sqrt{4q+9} \leq p \leq q+1/2+(1/2)\,\sqrt{4q+9}$. Since $a=4$, the largest $p$ allowed is $p=3$, which is admissible for $q=1$. In that case, $a(k)=4k$ and $\delta_k\geq 0$, showing that $t_k\leq 4\,A/(k+3)$ for all $k$ when \eqref{seq1} is satisfied with $\ma_k=2/(k+1)$ for all $k$.

(\textit{iii}) Suppose that $t_k$ satisfies \eqref{seq2} with $\ma_k=1/k$ for all $k$. We have $t_1\leq A$ by hypothesis; the induction hypothesis $t_k\leq A/k$ and \eqref{seq2} give $A/(k+1)-t_{k+1}\geq A/[k(k+1)^2]>0$, and thus imply that $t_{k+1}\leq A/(k+1)$.

(\textit{iv}) The function $t\to f(t)=t-t^2/A$ is increasing on $[0,A/2)$ with a maximum on $[0,A]$ equal to $A/4$ attained for $t=A/2$.
We have $t_2 \leq f(A/2)=A/4$, and thus $t_k \leq A/4$ for all $k\geq 2$. Take $p=A/t_2-2$, so that $p\geq 2$ and $t_2= A/(p+2)\leq A/4$. Suppose that $t_k\leq A/(p+k)$; we have $t_{k+1}\leq A\,[1/(k+p)-1/(k+p)^2]=A\,(k+p-1)/(k+p)^2=A/(k+p+1)-A/[(k+p)^2(k+p+1)]<A/(k+p+1)$, showing that $t_k\leq A/(p+k)$ for all $k\geq 2$.

When $t_1\leq A/2$, we take $p=A/t_1-1$, which gives $p\geq 1$, $t_1= A/(p+1)\leq A/2$ and $t_k< A/2$ for all $k> 1$. Assuming that $t_k\leq A/(p+k)$, we get $t_{k+1}<A/(k+p+1)$, showing that $t_k\leq A/(p+k)$ for all $k\geq 1$.
\carre
\vsp


\noindent{\em Proof of Theorem~\ref{Th:KH-empirical}.} The proof is based on \citep{Clarkson2010}.
For any $\xb^{(j)}\in\SX_C$, the definition \eqref{Delta_C} of $\Delta_C(\xi)$ gives
\be
\hspace{-0.7cm} \Delta_C[\xi_k^+(\xb^{(j)},\ma_{k+1})] &=& \hg_C[\mob_k+\ma_{k+1}(\eb_j-\mob_k)] - \hg_C(\mob^C_*) \nonumber \\
&=& \hg_C(\mob_k)-\hg_C(\mob^C_*)+2\,\ma_{k+1}\,(\eb_j-\mob_k)\TT\Kb_C(\mob_k-\widehat\mob^C)  +\ma_{k+1}^2\|\eb_j-\mob_k\|_{\Kb_C}^2 \,, \label{T1}
\ee
where $\ma_{k+1}=1/(k+1)$. The definition of $\widehat\mob^C$ implies that $\ub\TT\Kb_c\widehat\mob^C=\ub\TT\pb_C(\mu)$ for any $\ub\in\mathds{R}^C$ orthogonal to $\1b_C$, see \eqref{hatwn}. Therefore, $(\eb_j-\mob_k)\TT\Kb_C\widehat\mob^C=P_{K,\mu}(\xb^{(j)})-\sum_{i=1}^k \{\wb_k\}_i P_{K,\mu}(\xb_i)$,
and $\xb_{k+1}=\xb^{(j_{k+1})}$ with
\bea
j_{k+1} \in \Arg\min_{j\in\SI_C} P_{K,\xi_k}(\xb^{(j)})-P_{K,\mu}(\xb^{(j)}) = \Arg\min_{j\in\SI_C}(\eb_j-\mob_k)\TT\Kb_C(\mob_k-\widehat\mob^C)\,,
\eea
in agreement with \eqref{empirical-KH}. The convexity of $\hg_C(\cdot)$ implies that
\be
\hg_C(\mob^C_*) \geq \hg_C(\mob_k)+2\,(\mob^C_*-\mob_k)\TT\Kb_C(\mob_k-\widehat\mob^C) \geq \hg_C(\mob_k)+2\, \min_{j\in\SI_C} (\eb_j-\mob_k)\TT\Kb_C(\mob_k-\widehat\mob^C) \,, \label{convexity}
\ee
where the second inequality follows from $\mob^C_*\in\SP_C$, implying that
\bea
\Delta_C(\xi_{k+1}) = \Delta_C[\xi_k^+(\xb_{k+1},\ma_{k+1})] &\leq& (1-\ma_{k+1})\, [\hg_C(\mob_k)-\hg_C(\mob^C_*)] +\ma_{k+1}^2\|\eb_{j_{k+1}}-\mob_k\|_{\Kb_C}^2 \,.
\eea
The last term can be bounded as follows: as the weights $\mob_k$ belong to $\SP_C$ for all $k$, for all $j\in\SI_C$, we have
\be
\|\eb_j-\mob_k\|_{\Kb_C}^2 \leq \max_{\mob,\mob'\in\SP_C} (\mob-\mob')\TT\Kb_C(\mob-\mob') &=& \max_{i,j\in\SI_C} (\eb_i-\eb_j)\TT\Kb_C(\eb_i-\eb_j) \nonumber \\
&& \hspace{-5cm} = \max_{i,j\in\SI_C} K(\xb^{(i)},\xb^{(i)})+K(\xb^{(j)},\xb^{(j)})-2\,K(\xb^{(i)},\xb^{(j)}) \leq B_C \,, \label{Th1*}
\ee
with $B_C=4\, \overline{K}_C$ ($B_C=2\, \overline{K}_C$ when $K\geq 0$). It implies that
\be\label{recurrence-Th:KH-empirical}
\Delta_C(\xi_{k+1}) \leq (1-\ma_{k+1})\,\Delta_C(\xi_k) + B_C\, \ma_{k+1}^2 \,.
\ee
Since
\be\label{DeltaC1-max}
\Delta_C(\xi_1)\leq\MMD_K^2(\mu,\xi_1)=K(\xb_1,\xb_1)-2\,P_{K,\mu}(\xb_1)+\SE_K(\mu) \leq B_C \,,
\ee
Lemma~\ref{L:induction}-(\textit{i}) gives \eqref{bound:KH-empirical} when $\ma_k=1/k$ for all $k$.

When $\widehat\xi^C=\xi^C_*$, $\widehat\mob^C\in\SP_C$, and $\Delta_C(\xi_k)=\hg_C(\mob_k)$. Therefore,
\be\label{dum6}
\min_{j\in\SI_C} (\eb_j-\mob_k)\TT\Kb_C(\mob_k-\widehat\mob^C) \leq (\widehat\mob^C-\mob_k)\TT\Kb_C(\mob_k-\widehat\mob^C) = -\hg_C(\mob_k) \,,
\ee
and we get $\Delta_C(\xi_{k+1}) \leq (1-2\,\ma_{k+1})\,\Delta_C(\xi_k) + B_C\, \ma_{k+1}^2$ instead of \eqref{recurrence-Th:KH-empirical}. Lemma~\ref{L:induction}-(\textit{iii}) gives \eqref{bound:KH-empirical++}.
\carre
\vsp


\noindent{\em Proof of Theorem~\ref{Th:KH-predefined-SZ}.} $\Delta_C(\xi_k)$ satisfies \eqref{recurrence-Th:KH-empirical} with $\ma_{k+1}=2/(k+2)$. Lemma~\ref{L:induction}-(\textit{ii}) gives \eqref{bound:KH-predefined-SZ}.
\carre
\vsp


\noindent{\em Proof of Theorem~\ref{Th:KH-optimal-SZ}.} Consider again the proof of Theorem~\ref{Th:KH-empirical}. We have
\be
\Delta_C(\xi_{k+1}) = \min_{\ma\in[0,1]} \Delta_C[\xi_k^+(\xb^{(j_{k+1})},\ma)] \leq \Delta_C[\xi_k^+(\xb^{(j_{k+1})},\ma_{k+1})] \leq (1-\ma_{k+1})\,\Delta_C(\xi_k) + B_C\, \ma_{k+1}^2 \,, \label{Pth3}
\ee
for any arbitrary choice of $\ma_{k+1}\in[0,1]$, and $\ma_{k+1}=2/(k+2)$ for all $k$ has been shown to imply \eqref{bound:KH-predefined-SZ} in Theorem~\ref{Th:KH-predefined-SZ}. When $\widehat\xi^C=\xi^C_*$, we get $\Delta_C(\xi_{k+1}) \leq (1-2\,\ma_{k+1})\,\Delta_C(\xi_k) + B_C\, \ma_{k+1}^2$, see the proof of Theorem~\ref{Th:KH-empirical}. When we use $\ma_k=1/k$ for all $k$, Lemma~\ref{L:induction}-(\textit{iii}) implies \eqref{bound:KH-empirical++}.

The value of $\ma$ minimising $\Delta_C[\xi_k^+(\xb^{(j_{k+1})},\ma)]$ is
\be\label{alpha_k+1}
\widehat\ma_{k+1}=\frac{(\mob_k-\eb_{j_{k+1}})\TT\Kb_C(\mob_k-\widehat\mob^C)}{\|\eb_{j_{k+1}}-\mob_k\|_{\Kb_C}^2} \,,
\ee
so that \eqref{convexity} implies $\widehat\ma_{k+1}>0$ when $\MMD_K(\mu,\xi_k)>\MMD_K(\mu,\xi^C_*)$. The algorithm can thus be stopped if $\widehat\ma_{k+1}=0$.
Since $(\mob_k-\eb_{j_{k+1}})\TT\Kb_C(\mob_k-\widehat\mob^C)= \|\eb_{j_{k+1}}-\mob_k\|_{\Kb_C}^2+(\eb_{j_{k+1}}-\widehat\mob^C)\TT\Kb_C(\mob_k-\widehat\mob^C)-\|\eb_{j_{k+1}}-\widehat\mob^C\|_{\Kb_C}^2$, we have
$\widehat\ma_{k+1}\leq 1 + (\eb_{j_{k+1}}-\widehat\mob^C)\TT\Kb_C(\mob_k-\widehat\mob^C)/\|\eb_{j_{k+1}}-\mob_k\|_{\Kb_C}^2$.
When $\widehat\mob^C\in\SP_C$ (that is, when $\widehat\mob^C=\mob^C_*$), the definition of $j_{k+1}$ implies that $\widehat\ma_{k+1}\leq 1$ (since $\sum_{j=1}^C \{\widehat\omega^C\}_j (\eb_j-\widehat\mob^C)\TT\Kb_C(\mob_k-\widehat\mob^C)=0$ with all $\{\widehat\omega^C\}_j\geq 0$). However, nothing guarantees that $\widehat\ma_{k+1}\leq 1$ in general.
Direct calculation gives
\be
&& (\mob_k-\eb_j)\TT\Kb_C(\mob_k-\widehat\mob^C) = \sum_{i,\ell=1}^k \{\wb_k\}_i \{\wb_k\}_\ell K(\xb_i,\xb_\ell)-\sum_{i=1}^k \{\wb_k\}_i K(\xb_i,\xb^{(j)}) \nonumber \\
&& \hspace{5cm} + P_{K,\mu}(\xb^{(j)}) - \sum_{i=1}^k \{\wb_k\}_i P_{K,\mu}(\xb_i) \,, \label{alpha-opt1} \\
&& \|\eb_j-\mob_k\|_{\Kb_C}^2 = \sum_{i,\ell=1}^k \{\wb_k\}_i \{\wb_k\}_\ell K(\xb_i,\xb_\ell) - 2\, \sum_{i=1}^k \{\wb_k\}_i K(\xb_i,\xb^{(j)}) + K(\xb^{(j)},\xb^{(j)})\,, \label{alpha-opt2}
\ee
which together with \eqref{alpha_k+1} gives \eqref{alpha_k+1^*}.
The recursive updating of $Q_k=\sum_{i,\ell=1}^k \{\wb_k\}_i \{\wb_k\}_\ell K(\xb_i,\xb_\ell)$, $R_k=\sum_{i=1}^k \{\wb_k\}_i P_{K,\mu}(\xb_i)$ and $S_k(\xb)=P_{K,\xi_k}(\xb)$ gives Algorithm~\ref{algo:KH-optimal-alpha}.
\carre
\vsp


\noindent{\em Proof of Theorem~\ref{Th:KH weighed MMD-variant2}.} \mbox{}

(\textit{i}) When $\nu_k=\xi_k^*$ is substituted for $\xi_k$, we have
$\xb_{k+1}=\xb^{(j_{k+1})}$ with $j_{k+1}\in\Arg\min_{j\in\SI_C} (\eb_j-\mob_k^*)\TT\Kb_C(\mob_k^*-\widehat\mob^C)$ and, by construction, $\Delta_C(\xi_{k+1}) \leq \Delta_C[\nu_k^+(\xb_{k+1},\ma)]$ for any $\ma\in[0,1]$, where $\Delta_C(\xi)$ is given by \eqref{Delta_C} and $\nu_k^+(\xb,\ma)=(1-\ma)\,\xi_k^* + \ma\,\delta_{\xb}$.
Consider \eqref{T1} in the proof of Theorem~\ref{Th:KH-empirical}: we have
\bea
\Delta_C[\nu_k^+(\xb_{k+1},\ma)] = \hg_C(\mob_k^*)-\hg_C(\mob^C_*)+2\,\ma\,(\eb_{j_{k+1}}-\mob_k^*)\TT\Kb_C(\mob_k^*-\widehat\mob^C)+\ma^2\|\eb_{j_{k+1}}-\mob_k^*\|_{\Kb_C}^2 \,,
\eea
and the convexity of $\hg_C(\cdot)$ implies
\bea
\hg_C(\mob^C_*) \geq \hg_C(\mob_k^*)+2\,(\mob^C_*-\mob_k^*)\TT\Kb_C(\mob_k^*-\widehat\mob^C) \geq \hg_C(\mob_k^*)+2\, (\eb_{j_{k+1}}-\mob_k^*)\TT\Kb_C(\mob_k^*-\widehat\mob^C)\,,
\eea
where the second inequality follows from $\mob^C_*\in\SP_C$ and the choice of $j_{k+1}$.
Using $\|\eb_{j_{k+1}}-\mob_k^*\|_{\Kb_C}^2\leq B_C$, see the proof of Theorem~\ref{Th:KH-empirical}, we thus obtain
\be\label{dum1}
\Delta_C(\xi_{k+1}) \leq \Delta_C[\nu_k^+(\xb_{k+1},\ma_{k+1})] \leq (1-\ma_{k+1})\,\Delta_C(\nu_k)+B_C\,\ma_{k+1}^2\,, \quad k\geq 1\,,
\ee
for any predefined $\ma_{k+1}$. When $\ma_{k+1}=2/(k+2)$, the induction used in the proof of Theorem~\ref{Th:KH-predefined-SZ} gives \eqref{bound:KH-predefined-SZ}. When $\widehat\xi^C=\xi^C_*$, the right-hand side of \eqref{dum1} becomes $(1-2\,\ma_{k+1})\,\Delta_C(\nu_k)+B_C\,\ma_{k+1}^2$, see the proof of Theorem~\ref{Th:KH-empirical}, and if we take $\ma_k=1/k$ for all $k$, Lemma~\ref{L:induction}-(\textit{iii}) implies \eqref{bound:KH-empirical++}.

\vsp
(\textit{ii}) Suppose now that $\nu_k=\widehat\xi_k$ is substituted for $\xi_k$ at iteration $k$. Equation \eqref{alpha-opt1} gives
\bea
P_{K,\widehat\xi_k}(\xb_{k+1})-P_{K,\mu}(\xb_{k+1})+ \widehat\wb_k\TT\pb_k(\mu)-\SE_K(\widehat\xi_k) = (\eb_{j_{k+1}}-\widehat\mob_k)\TT\Kb_C(\widehat\mob_k-\widehat\mob^C)\,,
\eea
so that
\eqref{MMD-SBQ2} gives
\bea
\Delta_C(\xi_{k+1}) = \Delta_C(\xi_k) - \frac{\left[(\eb_{j_{k+1}}-\widehat\mob_k)\TT\Kb_C(\widehat\mob_k-\widehat\mob^C)\right]^2}{\min_{\scriptsize
\begin{array}{l}
\wb\in\mathds{R}^k \\
\1b_k\TT\wb=1
\end{array}
} \|K(\xb_{k+1},\cdot)-\wb\TT\kb_k(\cdot)\|_{\SH_K}^2} \,.
\eea
As long as $\hg_C(\mob^C_*) \leq \hg_C(\widehat\mob_k)$, that is, $\Delta_C(\widehat\xi_k)\geq 0$, \eqref{convexity} with $\widehat\mob_k$ substituted for $\mob_k$ gives
\be
\Delta_C(\xi_{k+1}) &\leq& \Delta_C(\xi_k) - \frac{\left[\hg_C(\widehat\mob_k)-\hg_C(\mob^C_*)\right]^2}{4\,\|K(\xb_{k+1},\cdot)-(\1b_k\TT/k)\kb_k(\cdot)\|_{\SH_K}^2} \nonumber \\
&&  = \Delta_C(\xi_k) - \frac{\Delta_C^2(\xi_k)}{4\, \left[K(\xb_{k+1},\xb_{k+1})+\1b_k\TT\Kb_k\1b_k/k^2-2\,\1b_k\TT\kb_k(\xb_{k+1})/k\right]} \nonumber \\
&&  \leq \Delta_C(\xi_k) - \frac{\Delta_C^2(\xi_k)}{4\,B_C} \,, \label{dum7}
\ee
with $\Delta_C(\xi_1)\leq B_C$, see \eqref{DeltaC1-max}. Lemma~\ref{L:induction}-(\textit{iv}) with $A=4\,B_C$ and $p_1=3$ gives \eqref{bound:KH-predefined-SZ}. When $\widehat\xi^C=\xi^C_*$, \eqref{dum6} gives $\Delta_C(\xi_{k+1}) \leq \Delta_C(\xi_k) - \Delta_C^2(\xi_k)/B_C$, and Lemma~\ref{L:induction}-(\textit{iv}) with $A=B_C$ and $p_2=2$ gives \eqref{KH-ii}.

\vsp
(\textit{iii}) Suppose finally that $\nu_k=\widetilde\xi_k$ is substituted for $\xi_k$ at iteration $k$. Since $\widetilde\xi_k$ is not necessarily in $\SM_{[1]}(\SX_C$), we shall use \eqref{gtilde} instead of \eqref{MMD2-hg}.
The definition of $\widetilde\mob^C$ implies $\Kb_C\widetilde\mob^C=\pb_C(\mu)$, and thus
$\xb_{k+1}=\xb^{(j_{k+1})}$ with
\bea
j_{k+1} \in \Arg\min_{j\in\SI_C} P_{K,\nu_k}(\xb^{(j)})-P_{K,\mu}(\xb^{(j)}) = \Arg\min_{j\in\SI_C} \eb_j\TT\Kb_C(\widetilde\mob_k-\widetilde\mob^C)\,.
\eea
Let $\widetilde\gb_k=2\,\Kb_C(\widetilde\mob_k-\widetilde\mob^C)$ denote the gradient of $\tg(\cdot)$ at $\widetilde\mob_k$. By construction, $\eb_j\TT\widetilde\gb_k=0$ for all $j$ with $\xb^{(j)}\in\supp(\widehat\xi^k)$, so that $\widetilde\mob_k\TT\Kb_C(\widetilde\mob_k-\widetilde\mob^C)=0$, and the convexity of $\tg(\cdot)$ implies
\be\label{Th4*1}
\tg_C(\mob^C_*) \geq \tg_C(\widetilde\mob_k)+2\,(\mob^C_*-\widetilde\mob_k)\TT\Kb_C(\widetilde\mob_k-\widetilde\mob^C) = \tg_C(\widetilde\mob_k)+2\,{\mob^C_*}\TT\Kb_C(\widetilde\mob_k-\widetilde\mob^C) \,.
\ee
Since $\mob^C_*\in\SP_C$, the definition of $j_{k+1}$ implies
\be\label{Th4*2}
\tg_C(\mob^C_*) \geq \tg_C(\widetilde\mob_k)+2\,\eb_{j_{k+1}}\TT\Kb_C(\widetilde\mob_k-\widetilde\mob^C) \,.
\ee
Therefore, as long as $\Delta_C(\widetilde\xi_k)=\tg_C(\widetilde\mob_k)-\tg_C(\mob^C_*)\geq 0$, \eqref{MMD-SBQ1} gives
\be
\Delta_C(\xi_{k+1}) &=& \Delta_C(\xi_k) - \frac{\left[\eb_{j_{k+1}}\TT\Kb_C(\widetilde\mob_k-\widetilde\mob^C) \right]^2}{\left[K(\xb_{k+1},\xb_{k+1})-\kb_k\TT(\xb_{k+1})\Kb_k^{-1}\kb_k(\xb_{k+1})\right]} \nonumber \\
&\leq& \Delta_C(\xi_k) - \frac{\Delta_C^2(\xi_k)}{4\, \left[K(\xb_{k+1},\xb_{k+1})-\kb_k\TT(\xb_{k+1})\Kb_k^{-1}\kb_k(\xb_{k+1})\right]}  \nonumber \\
&& \leq \Delta_C(\xi_k) - \frac{\Delta_C^2(\xi_k)}{4\,\overline{K}_C} \leq \Delta_C(\xi_k) - \frac{\Delta_C^2(\xi_k)}{4\,\overline{K}}\,. \label{dum5}
\ee
Since any signed measure supported on $\SX_C$ can be used, we may start with $\xi_0=0$, with weights $\mob_0=\0b_C$, so that \eqref{MMD-SBQ1} and the inequality above apply from $k=0$. We have $\Delta_C(\xi_0)\leq \MMD_K^2(\mu,\xi_0)=\SE_K(\mu)\leq \overline{K}$; therefore, $\Delta_C(\xi_k)\leq 4\,\overline{K}/(k+p_1)$, $k\geq 1$, for $p_1=4\,\overline{K}/\MMD_K^2(\mu,\xi_1)-1 \leq 4\,\overline{K}/\Delta_C(\xi_1)-1$; see the proof of Lemma~\ref{L:induction}-(\textit{iv}).
$\Delta_C(\xi_0)\leq \overline{K}$ implies $\Delta_C(\xi_1)\leq 3\,\overline{K}/4$, and we can also take $p_1=13/3$ in Lemma~\ref{L:induction}-(\textit{iv}), which gives $\Delta_C(\xi_k)\leq 4\,\overline{K}/(k+13/3)$, $k\geq 1$. This completes the proof of \eqref{KH-iii}.
\carre
\vsp

\noindent{\em Stopping conditions for Algorithms~\ref{algo:KH-IWO}-(\textit{ii}) and (\textit{iii}).}
Let $\widehat\gb_k=2\,\Kb_C(\widehat\mob_k-\widehat\mob^C)$ denote the gradient of $\hg(\cdot)$ at $\widehat\mob_k$. By construction, $(\eb_j-\widehat\mob_k)\TT\widehat\gb_k=0$ for all $j$ such that $\xb^{(j)}\in\supp(\widehat\xi^k)$, with
\bea
(\eb_j-\widehat\mob_k)\TT\widehat\gb_k = 2\left\{ P_{K,\widehat\xi^k}(\xb^{(j)})-P_{K,\mu}(\xb^{(j)})
+ \sum_{i=1}^k \{\widehat\wb_k\}_i \left[P_{K,\mu}(\xb_i)-P_{K,\widehat\xi^k}(\xb_i)\right] \right\} \,.
\eea
Therefore, $P_{K,\widehat\xi^k}(\xb^{(j)})-P_{K,\mu}(\xb^{(j)})$ equals a constant $c_k$ for all $\xb^{(j)}\in\supp(\widehat\xi^k)$, and the existence of an $\xb\in\SX_C$ such that $P_{K,\widehat\xi^k}(\xb)-P_{K,\mu}(\xb)<c_k$ implies that Algorithm~\ref{algo:KH-IWO}-(\textit{ii}) can still progress. Conversely, if $P_{K,\widehat\xi^k}(\xb)-P_{K,\mu}(\xb)\geq c_k$ for all $\xb\in\SX_C$, the algorithm can be stopped. We can thus add the following line to Algorithm~\ref{algo:KH-IWO}-(\textit{ii}):
\begin{description}
  \item[4'-(\textit{ii}):] {\bf if} $S_{k-1}(\xb_k)-P_{K,\mu}(\xb_k)\geq S_{k-1}(\xb_{k-1})-P_{K,\mu}(\xb_{k-1})$ {\bf then return} $\Xb_{k-1}$, $\xi_{k-1}$ and stop;
\end{description}

Similarly, $\eb_j\TT\widetilde\gb_k = 2\left[P_{K,\widehat\xi^k}(\xb^{(j)})-P_{K,\mu}(\xb^{(j)})\right]=0$ for all $j$ with $\xb^{(j)}\in\supp(\widehat\xi^k)$, with $\widetilde\gb_k$ the gradient of $\tg(\cdot)$ at $\widetilde\mob_k$; see the proof of Theorem~\ref{Th:KH weighed MMD-variant2}-(\textit{iii}). Therefore, $P_{K,\widehat\xi^k}(\xb^{(j)})=P_{K,\mu}(\xb^{(j)})$ for all $\xb^{(j)}\in\supp(\widetilde\xi^k)$; Algorithm~\ref{algo:KH-IWO}-(\textit{iii}) can be stopped when
$P_{K,\widehat\xi^k}(\xb)\geq P_{K,\mu}(\xb)$ for all $\xb\in\SX_C$ and we can add the following line to Algorithm~\ref{algo:KH-IWO}-(\textit{iii}):
\begin{description}
  \item[4'-(\textit{iii}):] {\bf if} $S_{k-1}(\xb_k)-P_{K,\mu}(\xb_k)\geq 0$ {\bf then return} $\Xb_{k-1}$, $\xi_{k-1}$ and stop;
\end{description}


\noindent{\em Proof of Theorem~\ref{Th:empirical-greedy}.}
We have $\SE_K(\nu-\mu)=\SE_{K_\mu}(\nu)$ for any $\nu\in\SM_{[1]}(\SX_C)$ with $K_\mu$ the reduced kernel defined by \eqref{Kmu}; see \citet{DamelinHRZ2010}, \citet[Th.~3.5]{PZ2020-SIAM}. Let $\mob_k$ be the vector of weights associated with $\xi_k$ at step $k$. Since $\MMD_K^2(\mu,\xi_k) = \mob_k\TT {\Kb_\mu}_C \mob_k$,
see \eqref{MMD-with-Ktilde}, we have
\bea
(k+1)^2\,\MMD_K^2(\mu,\xi_{k+1}) = k^2\,\MMD_K^2(\mu,\xi_k)+2k \, \mob_k\TT {\Kb_\mu}_C\eb_{j_{k+1}} + K_\mu(\xb^{(j_{k+1})},\xb^{(j_{k+1})}) \,,
\eea
where $j_{k+1}\in\Arg\min_{j\in\SI_C} 2k \, \mob_k\TT {\Kb_\mu}_C\eb_j + K_\mu(\xb^{(j)},\xb^{(j)})$, and $\xb^{(j_{k+1})}$ coincides with \eqref{empirical-greedy} when $\SX_C$ is substituted for $\SX$. Therefore, for any $\mob\in\SP_C$,
we have
\be
(k+1)^2\,\MMD_K^2(\mu,\xi_{k+1}) &\leq& k^2\,\MMD_K^2(\mu,\xi_k)+2k \, \mob_k\TT {\Kb_\mu}_C \mob + \overline{K}_{\mu,C}  \nonumber \\
&& \hspace{-2cm} \leq k^2\,\MMD_K^2(\mu,\xi_k)+ 2k \, (\mob_k\TT {\Kb_\mu}_C \mob_k)^{1/2}(\mob\TT {\Kb_\mu}_C \mob)^{1/2}  + \overline{K}_{\mu,C} \,, \label{dum2}
\ee
where $\overline{K}_{\mu,C} = \max_{\xb\in\SX_C} K_\mu(\xb,\xb)\leq A_C$, see \eqref{KmuC}. The inequality \eqref{dum2} is satisfied in particular for $\mob^C_*\in\Arg\min_{\mob\in\SP_C} \mob\TT {\Kb_\mu}_C \mob$, with ${\mob^C_*}\TT {\Kb_\mu}_C\mob^C_*=M_C^2$, therefore
\be\label{ineq_MMD1}
(k+1)^2\,\MMD_K^2(\mu,\xi_{k+1}) &\leq& k^2\,\MMD_K^2(\mu,\xi_k)+ 2k\, M_C\, \MMD_K(\mu,\xi_k) + A_C \,.
\ee
We prove \eqref{induction-1} by induction on $n$.
For $n=1$, $\MMD_K^2(\mu,\delta_{\xb_1})=K_\mu(\xb_1,\xb_1) \leq \overline{K}_{\mu,C} \leq A_C$. Suppose that \eqref{induction-1} is true for $n$. Then, \eqref{ineq_MMD1} implies
\bea
\MMD_K^2(\mu,\xi_{n+1}) &\leq& \frac{1}{(n+1)^2} \, \left\{ n^2 \left(M_C^2+A_C \, \frac{1+\log n}{n}\right)  + 2n\,M_C\,\left(M_C^2+A_C \, \frac{1+\log n}{n}\right)^{1/2} + A_C \right\} \\
&& \hspace{-3cm} \leq \frac{1}{(n+1)^2} \, \left\{ n^2 \left(M_C^2+A_C \, \frac{1+\log n}{n}\right)  + n\,\left(2\,M_C^2+A_C \, \frac{1+\log n}{n}\right) + A_C \right\} \\
&& \hspace{-3cm} = M_C^2+A_C \, \frac{1+\log (n+1)}{n+1} - \frac{M_C^2+A_C\left[(n+1)\log(1+1/n)-1\right]}{(n+1)^2} \\
&& \hspace{-3cm} = M_C^2+A_C \, \frac{1+\log (n+1)}{n+1} - \frac{M_C^2/(n+1)+A_C\left[-\log(1-1/(n+1))-1/(n+1)\right]}{n+1} \\
&& \hspace{-3cm} \leq M_C^2+A_C \, \frac{1+\log (n+1)}{n+1}
\eea
since $\log(1+x)\leq x$ for $x>-1$,
which concludes the proof of \eqref{induction-1}.
\carre
\vsp


\noindent{\em Proof of Theorem~\ref{Th:greedy weighed MMD}.}
Using \eqref{T1} and \eqref{Th1*}, we get
\bea
\min_{\xb\in\SX_C} \Delta_C[\xi_k^+(\xb,\ma_{k+1})] \leq \hg_C(\mob_k)-\hg_C(\mob^C_*) + \min_{j\in\SI_C} 2\,\ma_{k+1}\,(\eb_j-\mob_k)\TT\Kb_C(\mob_k-\widehat\mob^C) + B_C\, \ma_{k+1}^2 \,,
\eea
and, from the same arguments as in the proof of Theorem~\ref{Th:KH-empirical},
\be\label{Main-th2}
\Delta_C(\xi_{k+1}) = \min_{\xb\in\SX_C} \Delta_C[\xi_k^+(\xb,\ma_{k+1})] \leq (1-\ma_{k+1})\,\Delta_C(\xi_k) + B_C\, \ma_{k+1}^2 \,.
\ee
When $\ma_k=2/(k+1)$ for all $k$, Lemma~\ref{L:induction}-(\textit{ii}) gives \eqref{bound:KH-predefined-SZ}.

Similarly, when $\widehat\xi^C=\xi^C_*$, we have
\bea
\Delta_C(\xi_{k+1}) = \min_{\xb\in\SX_C} \Delta_C[\xi_k^+(\xb,\ma_{k+1})] \leq (1-2\,\ma_{k+1})\,\Delta_C(\xi_k) + B_C\, \ma_{k+1}^2 \,,
\eea
see the proof of Theorem~\ref{Th:KH-empirical}, and Lemma~\ref{L:induction}-(\textit{iii}) gives \eqref{bound:KH-empirical++}.
\carre
\vsp

\noindent{\em Proof of Theorem~\ref{Th:greedy MMD_opt}.}
For $\ma\in[0,1]$, any $\xb^{(j)}\in\SX_C$ satisfies
\be\label{dum3}
\Delta_C[\xi_k^+(\xb^{(j)},\ma)] = \hg_C(\mob_k)-\hg(\mob^C_*) + 2\,\ma\,(\eb_j-\mob_k)\TT\Kb_C(\mob_k-\widehat\mob^C)+\ma^2\, \|\eb_j-\mob_k\|_{\Kb_C}^2 \,,
\ee
see \eqref{T1}, the right-hand side of which is minimum when
\bea
\ma=\widehat\ma_{k+1,j} = \frac{(\eb_j-\mob_k)\TT\Kb_C(\widehat\mob^C-\mob_k)}{\|\eb_j-\mob_k\|_{\Kb_C}^2} \,.
\eea
By restricting $\ma$ to $[0,1]$, we obtain $[\xb_{k+1},\ma_{k+1}]=[\xb^{(j_{k+1}^*)},\ma_{k+1,j}^*]$ with
\bea
\ma_{k+1,j}^* &=& \max\{0,\min\{\widehat\ma_{k+1,j},1\}\} \nonumber \\
j_{k+1}^* &=& \arg\min_{j\in\SI_C} 2\,\ma_{k+1,j}^*\,(\eb_j-\mob_k)\TT\Kb_C(\mob_k-\widehat\mob^C)+(\ma_{k+1,j}^*)^2 \, \|\eb_j-\mob_k\|_{\Kb_C}^2 \,. 
\eea
Using \eqref{alpha-opt1} and \eqref{alpha-opt2}, we obtain that $\widehat\ma_{k+1,j}$ is given by \eqref{alpha_k+1^*} with $\xb^{(j)}$ substituted for $\xb_{k+1}$. The recursive updating of $Q_k=\sum_{i,\ell=1}^k \{\wb_k\}_i \{\wb_k\}_\ell K(\xb_i,\xb_\ell)$, $R_k=\sum_{i=1}^k \{\wb_k\}_i P_{K,\mu}(\xb_i)$ and $S_k(\xb)=P_{K,\xi_k}(\xb)$ gives Algorithm~\ref{algo:GM-alphaopt}. As in the proof of Theorem~\ref{Th:KH-optimal-SZ},  $\MMD_K(\mu,\xi_k)>\MMD_K(\mu,\xi^C_*)$ implies that there exists $j\in\SI_C$ such that $(\eb_j-\mob_k)\TT\Kb_C(\mob_k-\widehat\mob^C)<0$, and therefore $\widehat\ma_{k+1,j}=\ma(\xb^{(j)})>0$. Conversely, $\ma(\xb) =0$ for all $\xb\in\SX_C$ implies that $\MMD_K(\mu,\xi_k)=\MMD_K(\mu,\xi^C_*)$ and Algorithm~\ref{algo:GM-alphaopt} can be stopped.

As $\|\eb_j-\mob_k\|_{\Kb_C}^2 \leq B_C$ in \eqref{dum3}, see \eqref{Th1*}, we have
\bea
\min_{\xb\in\SX_C,\ma\in[0,1]} \Delta_C[\xi_k^+(\xb,\ma)] \leq \hg_C(\mob_k)-\hg_C(\mob^C_*) + \min_{j\in\SI_C} 2\,\ma_{k+1}\,(\eb_j-\mob_k)\TT\Kb_C(\mob_k-\widehat\mob^C) + B_C\, \ma_{k+1}^2
\eea
for any predefined choice of $\ma_{k+1}$ in $[0,1]$. The rest of the proof is similar to that of Theorem~\ref{Th:KH-optimal-SZ}.
\carre
\vsp

\noindent{\em Proof of Theorem~\ref{Th:SBQ12}.}
For any $\xi_k$, the value of $\Delta_C(\xi_{k+1})$ obtained by SBQ cannot exceed that obtained by IWO applied to KH, which yields the bounds given in Theorem~\ref{Th:SBQ12} for \eqref{SBQ1} and \eqref{SBQ2}.

Denote $\xi^{++}(\xb,\ma)=\xi+w\delta_\xb$ for any $\xi\in\SM(\SX_C)$, $\xb\in\SX_C$ and $w\in\mathds{R}$,
so that \eqref{SBQ3} corresponds to $\xi_{k+1}=\xi_k^{++}(\xb_{k+1},w_{k+1})$ with $[\xb_{k+1},w_{k+1}] \in \Arg\min_{\xb\in\SX_C,\,w} \MMD_K^2[\xi_k^{++}(\xb,w)]$. We get
\bea
\Delta_C(\xi_{k+1})= \Delta_C(\xi_k) - \frac{[\eb_{j_{k+1}}\TT\Kb_C(\mob_k-\widetilde\mob^C)]^2}{K(\xb_{k+1},\xb_{k+1})} \,,
\eea
where $j_{k+1}\in\Arg\max_{j\in\SI_C} [\eb_j\TT\Kb_C(\mob_k-\widetilde\mob^C)]^2/K(\xb^{(j)},\xb^{(j)})$, see the proof of Theorem~\ref{Th:KH weighed MMD-variant2}-(\textit{iii}). The convexity of $\tg(\cdot)$ implies
$\tg_C(\mob^C_*) \geq \tg_C(\mob_k)+2\,{\mob^C_*}\TT\Kb_C(\mob_k-\widetilde\mob^C)$, see \eqref{Th4*1}. Therefore, as long as $\Delta_C(\xi_k)=\tg_C(\mob_k)-\tg_C(\mob^C_*)\geq 0$, $[{\mob^C_*}\TT\Kb_C(\mob_k-\widetilde\mob^C)]^2\geq [\tg_C(\mob_k)-\tg_C(\mob^C_*)]^2/4$. The maximum of $[\mob\TT\Kb_C(\mob_k-\widetilde\mob^C)]^2$ with respect to $\mob\in\SP_C$ is attained on a vertex of $\SP_C$, which implies that $[\eb_{j_{k+1}}\TT\Kb_C(\mob_k-\widetilde\mob^C)]^2 \geq [\tg_C(\mob_k)-\tg_C(\mob^C_*)]^2/4$ when $K(\xb,\xb)=\overline{K}_C$ for all $\xb$, and $\Delta_C(\xi_k)$ satisfies \eqref{dum5}. The conclusion is the same as for Theorem~\ref{Th:KH weighed MMD-variant2}-(\textit{iii}).
\carre
\vsp


\noindent{\em Proof of Lemma~\ref{L:XC-random}.} $A_C$ depends on $\SX_C$, but $A_C\leq A(\mu)$ since $\overline{K}_C \leq \overline{K}$; similarly, $B_C\leq B$. Since $M_C^2 \leq \1b_C\TT{\Kb_\mu}_C\1b_C/C^2$, we get
\bea
\Ex\{M_C^2\} &\leq& \frac{1}{C^2}\, \Ex\left\{\sum_{i,j=1}^C K_\mu(\xb^{(i)},\xb^{(j)}) \right\} \\
&& = \frac{1}{C^2}\, \Ex\left\{\sum_{i,j=1}^C K(\xb^{(i)},\xb^{(j)}) \right\} - \frac{2}{C} \, \Ex\left\{\sum_{i=1}^C P_{K,\mu}(\xb^{(i)}) \right\} + \SE_K(\mu) \\
&& = \frac{1}{C^2}\, \Ex\left\{\sum_{i=1}^C K(\xb^{(i)},\xb^{(i)}) \right\} + \frac{1}{C^2}\, \Ex\left\{\sum_{\scriptsize\begin{array}{l}i,j=1 \\
i\neq j \end{array}}^C K(\xb^{(i)},\xb^{(j)}) \right\} -  \SE_K(\mu) \\
&& = \frac{\tau_1(\mu)}{C} + \frac{C(C-1)}{C^2} \, \SE_K(\mu) -  \SE_K(\mu) = \frac{\tau_1(\mu)-\SE_K(\mu)}{C} \,. \hspace{4cm} \carre
\eea

\vsp
\noindent{\em Proof of Theorem~\ref{Th:iid}.}
We have $\MMD_K^2(\mu,\xi_{n,e})=\1b_n\TT{\Kb_\mu}_n\1b_n/n^2$, with $\Ex_\mu\{K_\mu(\cdot,X)\}\equiv 0$ on $\SX$ and $\int_{\SX^2} K_\mu(\xb,\xb')\,\dd\mu(\xb)\dd\mu(\xb')=\SE_{K_\mu}(\mu)=0$. From \citet[p.~194]{Serfling80}, the U-statistic $U_n=2/[n(n-1)]\,\sum_{i<j} K_\mu(\xb_i,\xb_j)$ satisfies $n\,U_n \rad Y=\sum_{i=1}^\infty \ml_i (\chi_{1i}^2-1)$.
The V-statistic $V_n=(1/n^2)\,\sum_{i,j} K_\mu(\xb_i,\xb_j)=\MMD_K^2(\mu,\xi_{n,e})$ satisfies
$V_n=(1-1/n)\,U_n+ (1/n^2) \,\sum_i K_\mu(\xb_i,\xb_i)$, with $U_n \raas 0$ and $(1/n) \,\sum_i K_\mu(\xb_i,\xb_i) \raas \sum_{i=1}^\infty \ml_i$. Therefore, $n\,V_n \rad Z=\sum_{i=1}^\infty \ml_i \chi_{1i}^2$.
\carre

\section*{Appendix C: multiple random candidate sets}

Suppose that $\xb_{k+1}$ is selected within $\SX_C[k+1]$ at iteration $k$. Denote $\SX_{C_{k+1}}=\cup_{i=1}^{k+1} \SX_C[i]$ and let $\SP_{C_{k+1}}$ be the corresponding probability simplex in $\mathds{R}^{C_{k+1}}$ (with $C_{k+1}=(k+1)\,C$ when the $\SX_C[i]$ do not intersect). To a measure $\xi_k$ in $\SM(\SX_{C_k})$ (respectively, in $\SM_{[1]}^+(\SX_{C_k})$) corresponds a vector of weights $\mob_k$ in $\mathds{R}^{C_k}$ (respectively, in $\SP_{C_k}$), and we denote by $\mob_k'$ the same vector plunged into $\mathds{R}^{C_{k+1}}$ (respectively, into $\SP_{C_{k+1}}$); $\xi_*^{C[k+1]}$ is the measure in $\SM_{[1]}^+(\SX_C[k+1])$ that minimises $\MMD(\mu,\xi)$ and $\mob_*^{C[k+1]'}$ is the vector of associated weights in $\SP_{C_{k+1}}$. Similarly, $\widehat\mob^{C_{k+1}}$ denotes the vector of weights for the optimal measure in $\SM_{[1]}^+(\SX_{C_{k+1}})$.

Consider one-step-ahead algorithms (Algorithms~\ref{algo:KH-alphak}, \ref{algo:KH-optimal-alpha}, \ref{algo:GM-alphak} and \ref{algo:GM-alphaopt}) and $\xi_k^+[\xb_{k+1},\ma_{k+1}]=(1-\ma_{k+1})\,\xi_k+\ma_{k+1}\,\delta_{\xb_{k+1}}$ constructed at iteration $k$. We have
\bea
\MMD_K^2(\mu,\xi_k^+[\xb_{k+1},\ma_{k+1}]) - \MMD_K^2(\mu,\xi_*^{C[k+1]}) &=& \hg_{C_{k+1}}(\mob_k')-\hg_{C_{k+1}}(\mob_*^{C[k+1]'}) \\
&& \hspace{-5cm} + 2\,\ma_{k+1}\,(\eb_{j_{k+1}}-\mob_k')\TT\Kb_{C_{k+1}}(\mob_k'-\widehat\mob^{C_{k+1}}) + \ma_{k+1}^2 \|\eb_{j_{k+1}}-\mob_k'\|^2_{\Kb_{C_{k+1}}}\,,
\eea
with $\eb_{j_{k+1}}$ the basis vector in $\mathds{R}^{C_{k+1}}$ corresponding to $\xb_{k+1}\in\SX_C[k+1]$, and $\|\eb_{j_{k+1}}-\mob_k'\|^2_{\Kb_{C_{k+1}}} \leq B$, see \eqref{T1}, \eqref{Th1*} and Lemma~\ref{L:XC-random}. The convexity of $\hg_{C_{k+1}}(\cdot)$ implies
\bea
\hg_{C_{k+1}}(\mob_*^{C[k+1]'}) \geq \hg_{C_{k+1}}(\mob_k') +2\, (\mob_*^{C[k+1]'}-\mob_k')\TT\Kb_{C_{k+1}}(\mob_k'-\widehat\mob^{C_{k+1}})
\eea
so that $2\,(\eb_{j_{k+1}}-\mob_k')\TT\Kb_{C_{k+1}}(\mob_k'-\widehat\mob^{C_{k+1}}) \leq \hg_{C_{k+1}}(\mob_*^{C[k+1]'}) - \hg_{C_{k+1}}(\mob_k')$ when $\xb_{k+1}$ is chosen by kernel herding. This gives
\bea
\MMD_K^2(\mu,\xi_k^+[\xb_{k+1},\ma_{k+1}]) - \MMD_K^2(\mu,\xi_*^{C[k+1]}) &\leq& \\
&& \hspace{-5cm} (1-\ma_{k+1})\,\left[\MMD_K^2(\mu,\xi_k) - \MMD_K^2(\mu,\xi_*^{C[k+1]})\right] + B\, \ma_{k+1}^2 \,.
\eea
When each $\SX_C[i]$ is made of independent samples from $\mu$, we have $\Ex_\mu\{\MMD_K^2(\mu,\xi_*^{C[i]})\}=\Ex_\mu\{M_C^2\}$ for all $i$, and therefore, when $\ma_{k+1}$ is predefined (deterministic),
\bea
\Ex_\mu\{\Delta(\xi_{k+1})\} \leq (1-\ma_{k+1})\,\Ex_\mu\{\Delta(\xi_k)\} + B\, \ma_{k+1}^2 \,,
\eea
where we have denoted $\Delta(\xi)=\Delta_{C[1]}(\xi)=\MMD_K^2(\mu,\xi) - \MMD_K^2(\mu,\xi_*^{C[1]})$.
From the same arguments as those used in the proof of Theorem~\ref{Th:KH-empirical}, we thus obtain that the measure generated by Algorithm~\ref{algo:KH-alphak} with $\ma_k=1/k$ and using a set $\SX_C[k]$ composed of $C$ independent samples from $\mu$ for all $k$ satisfies $\Ex_\mu\{\MMD_K^2(\mu,\xi_k)\} \leq \Ex_\mu\{M_C^2\}+ B\,(2+\log n)/(n+1)$, $n\geq 1$, compare with \eqref{bound:KH-empirical}. When $\ma_k=2/(k+1)$ in Algorithm~\ref{algo:KH-alphak}, then $\Ex_\mu\{\MMD_K^2(\mu,\xi_k)\} \leq \Ex_\mu\{M_C^2\}+ 4\,B/(n+3)$, $n\geq 1$, compare with \eqref{bound:KH-predefined-SZ}.
Following the arguments used in the proof of Theorem~\ref{Th:KH-optimal-SZ}, see \eqref{Pth3}, we get the same bound for Algorithm~\ref{algo:KH-optimal-alpha}. Also, following the arguments in the proofs of Theorems~\ref{Th:greedy weighed MMD} and \ref{Th:greedy MMD_opt}, similar bounds are obtained for Algorithms~\ref{algo:GM-alphak} and \ref{algo:GM-alphaopt}, which extends the results in those theorems to this situation of multiple random candidate sets. A similar extension applies to Algorithm~\ref{algo:KH-IWO}-(\textit{i}); see the proof of Theorem~\ref{Th:KH weighed MMD-variant2}-(\textit{i}).

Consider now Algorithm~\ref{algo:KH-IWO}-(\textit{ii}) and (\textit{iii}) and SBQ. For Algorithm~\ref{algo:KH-IWO}-(\textit{ii}), we have
\bea
\MMD_K^2(\mu,\xi_{k+1}) \leq \MMD_K^2(\mu,\xi_k) - \frac{\left[(\eb_{j_{k+1}}-\mob_k')\TT\Kb_{C_{k+1}}(\mob_k'-\widehat\mob^{C_{k+1}}) \right]^2}{B}
\eea
and the convexity of $\hg_{C_{k+1}}(\cdot)$ gives
\bea
\MMD_K^2(\mu,\xi_{k+1}) \leq \MMD_K^2(\mu,\xi_k) - \frac{\left[\hg_{C_{k+1}}(\mob_k')-\hg_{C_{k+1}}(\mob_*^{C[k+1]'})\right]^2}{4\,B}\,.
\eea
Since $\Ex_\mu\{\MMD_K^2(\mu,\xi_*^{C[i]})\}=\Ex_\mu\{M_C^2\}$ for all $i$, Jensen's inequality gives
\bea
\Ex_\mu\left\{\left[\hg_{C_{k+1}}(\mob_k')-\hg_{C_{k+1}}(\mob_*^{C[k+1]'})\right]^2\right\} &\geq &
\left[\Ex_\mu\left\{\hg_{C_{k+1}}(\mob_k')-\hg_{C_{k+1}}(\mob_*^{C[k+1]'})\right\}\right]^2 = \Ex_\mu^2\{\Delta(\xi_k)\}\,.
\eea
We thus obtain
\bea
\Ex_\mu\{\Delta(\xi_{k+1})\} \leq \Ex_\mu\{\Delta(\xi_k)\} - \frac{\Ex_\mu^2\{\Delta(\xi_k)\}}{4\,B} \,,
\eea
an inequality similar to \eqref{dum7} in the proof of Theorem~\ref{Th:KH weighed MMD-variant2}-(\textit{ii}), and $\xi_n$ satisfies an inequality of the form \eqref{bound:KH-predefined-SZ} where each term is replaced by its expected value.

Similar developments yield an inequality similar to \eqref{dum5}, with expected values everywhere, and thus an extension of \eqref{KH-iii} for Algorithm~\ref{algo:KH-IWO}-(\textit{iii}). Theorem~\ref{Th:SBQ12}, that indicates that the performance of SBQ cannot be worse that that of Algorithm~\ref{algo:KH-IWO}, continues to apply; the details are omitted.

\section*{Acknowledgments}
This work was partly supported by project INDEX (INcremental Design of EXperiments) ANR-18-CE91-0007 of the French National Research Agency (ANR). The author would like to thank Chris Oates for sending him a preprint of the paper \citep{TeymurGRO2021} which was deeply inspiring.

The author is grateful to the two anonymous referees for their careful reading and their comments and suggestions which helped to improve the paper.

\bibliographystyle{apalike}

}

\end{document}